\documentclass[10pt,twocolumn,letterpaper]{article}

\usepackage[pagenumbers]{cvpr} 

\newcommand{\rr}[1]{{\color{black}#1}}

\definecolor{cvprblue}{rgb}{0.21,0.49,0.74}
\usepackage[pagebackref,breaklinks,colorlinks,allcolors=cvprblue]{hyperref}

\usepackage[accsupp]{axessibility}  
\usepackage{colortbl} 
\usepackage{amssymb} 
\usepackage{booktabs} 
\usepackage{multirow}
\usepackage{array}
\usepackage{fancyvrb} 
\usepackage{fvextra} 
\usepackage{tcolorbox} 
\tcbuselibrary{breakable} 
\tcbuselibrary{listings}
\usepackage{tabularx}

\newcommand{\sysname}{GUIDE}

\def\paperID{35968} 
\def\confName{CVPR}
\def\confYear{2026}

\title{\sysname{}: A Benchmark for Understanding and Assisting Users in Open-Ended GUI Tasks}

\author{
Saelyne Yang$^{1,2}$
\quad Jaesang Yu$^1$
\quad Yi-Hao Peng$^2$
\quad Kevin Qinghong Lin$^3$
\quad Jae Won Cho$^4$
\\
Yale Song$^5$
\quad 
Juho Kim$^{1,6}$
\\
{\small $^1$KAIST \quad
$^2$Carnegie Mellon University \quad
$^3$University of Oxford \quad
$^4$Konkuk University \quad
$^5$Google Inc. \quad
$^6$SkillBench}
}

\begin{document}
\maketitle
\begin{abstract}
Graphical User Interface (GUI) agents have the potential to assist users in interacting with complex software (e.g., PowerPoint, Photoshop). While prior research has primarily focused on automating user actions through clicks and keystrokes, this paradigm overlooks human intention, where users value the ability to explore, iterate, and refine their ideas while maintaining agency.
To move beyond automation and toward collaboration, GUI agents must understand what users are doing and why. We introduce \rr{\textbf{\sysname{}} (\textbf{G}UI \textbf{U}ser \textbf{I}ntent \textbf{D}etection \textbf{E}valuation)}, a benchmark that evaluates AI models on their ability to perceive user behavior, infer intent, and provide assistance in open-ended GUI tasks. \sysname{} consists of 67.5 hours of screen recordings from 120 novice user demonstrations with think-aloud narrations, across 10 software. \sysname{} defines three tasks—(i) Behavior State Detection, (ii) Intent Prediction, and (iii) Help Prediction that test a model’s ability to recognize behavior state, reason about goals, and decide when and how to help. Evaluations across eight state-of-the-art multimodal models reveal that all models struggled, achieving only 44.6\% and 55.0\% accuracy on behavior state and help prediction. However, providing user context significantly improved the performance, raising help prediction by up to 50.2\rr{pp}, highlighting the critical role of structured user understanding in effective assistance.
\rr{Our dataset is available at} \url{https://guide-bench.github.io}.
\end{abstract}    
\section{Introduction}
\begin{table*}[t]
\centering
\small
\setlength{\tabcolsep}{1pt} 
\begin{tabular}{lcccccccc}
\toprule
\textbf{Dataset} & \textbf{Domain \#} & \textbf{Video \#} & \textbf{Video Duration} & \textbf{Video Source} & \textbf{Primary Goal} & \multicolumn{3}{c}{\textbf{Evaluation Focus}} \\
\cmidrule(lr){7-9}
 & & & & & & \textbf{Behavior} & \textbf{Intent} & \textbf{Help} \\
\midrule
PsTuts~\cite{li2020pstuts} & 1 & - & 71.4 h & Instructional Videos & Action Understanding &  &  &  \\
VideoWebArena~\cite{jang2025videowebarena} & 6 & 74 & 3.8 h & Human-Recorded Tutorials & Task Automation &  &  &  \\
VideoGUI~\cite{lin2024videogui} & 11 & 178 & 7.1 h & Instructional Videos & Task Automation &  & \checkmark &  \\
UI-Vision~\cite{pmlr-v267-nayak25a} & 83 & 450 & 4.8 h & Experts Performing Tasks & Task Automation &  &  &  \\
AssistGUI~\cite{gao2024assistgui} & 9 & 100 & \textless 8.3 h & Instructional Videos & Task Automation &  & \checkmark &  \\
WorldGUI~\cite{zhao2025worldgui} & 10 & 611 & \textless 30.5 h & Instructional Videos & Task Automation &  & \checkmark &  \\
\midrule
\sysname{} (Ours) & 10 & 120 & 67.5 h & Novice Users’ Demonstrations & Behavior Understanding & \checkmark & \checkmark & \checkmark \\
\bottomrule
\end{tabular}
\caption{\textbf{Comparison of \sysname{} with existing GUI video understanding datasets.} \sysname{} differs from existing benchmarks by $(i)$ collecting screen recordings from novice users, $(ii)$ capturing how they naturally behave in open-ended tasks with a focus on behavior understanding, and $(iii)$ evaluating systems based on human user needs rather than task automation. }
\label{tab:datasets}
\end{table*}
\begin{figure}[t]
  \centering
  \includegraphics[width=\linewidth]{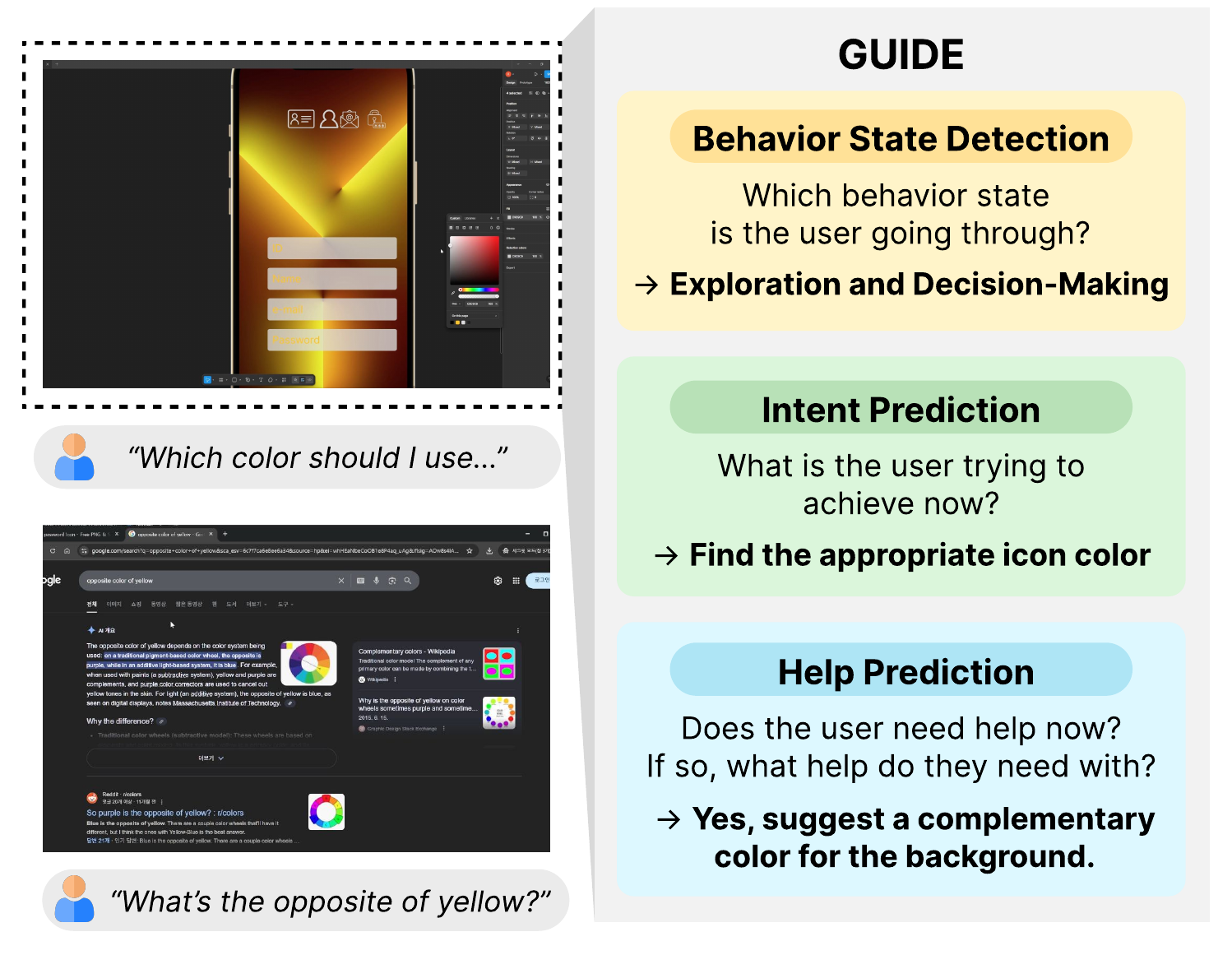}
  \caption{
  An example of the \sysname{} benchmark, which jointly models three tasks: Behavior State Detection, Intent Prediction, and Help Prediction, to interpret what the user is doing, aiming to achieve, and whether and what they may need assistance with during open-ended software tasks.
  }
  \label{fig:taxonomy}
\end{figure}

Graphical User Interface (GUI) agents hold great promise for supporting users in complex workflows, in mobile~\cite{liu2025learnact, jang2025_monday, zhang24android}, web~\cite{song2025bearcubsbenchmarkcomputerusingweb, hong2023cogagent, ye2025realwebassist, deng2023mind2web, zhou2023webarena}, and software application tasks~\cite{zhang2025tongui, watchandlearn, feizi2025groundingcomputeruseagents}.
In creative and analytical tools such as Photoshop or PowerPoint, these agents can automate repetitive subtasks or provide guidance to help users achieve their goals more efficiently. 
Most existing GUI agents, both in academic research~\cite{showui, gao2024assistgui, lin2024videogui, zhao2025worldgui} and in commercial services like Microsoft Office Copilot~\cite{copilot} or Figma Make~\cite{figma_make}, focus on full automation: given a goal, they either execute a sequence of clicks and keystrokes to complete the task or directly generate the desired output.

While this approach offers convenience, it overlooks how people actually work. 
In real-world open-ended workflows, 
\rr{success is not driven solely by efficiency—user satisfaction plays an equally critical role.
Automated agents assume fixed goals, yet users frequently revise their intentions mid-task. 
For example, a user may reposition an element multiple times before reverting to the original---behavior an automated agent would treat as redundant, but which is essential for forming a preference.
Rather than replacing user agency, effective assistance should accelerate exploration while keeping the user in control~\cite{Khurana2025DoItWithMe}.}

Recent work on proactive task assistance takes a more balanced approach~\cite{collabllm2025, lu2025proactive, yang2025contextagentcontextawareproactivellm, zhang2024proagentbuildingproactivecooperative, yang2025fingertip20kbenchmarkproactive}. Rather than automate tasks for users, proactive assistants infer a user's context and intent and deliver timely help. Studies in programming and productivity tools show higher efficiency and satisfaction when a system detects a need and intervenes at the right moment~\cite{Pu2025Assistance,Chen2025NeedHelp,peng2025morae,wu2025collabllm}.
Yet, the ability to model and track users’ evolving context remains underexplored in current multimodal systems that power GUI agents.

To achieve a truly human-assisting GUI agent, a key ability is to comprehend users' cognitive context and intentions to provide appropriate support~\cite{horvitz2013lumiereprojectbayesianuser}. In real-world scenarios, users rarely articulate their goals or needs explicitly, making it natural for systems to rely primarily on visual cues from the screen. 
These user actions often carry semantic structure, such as hovering, undoing, or repeatedly opening menus, that signal intent. 
However, interpretation remains challenging: similar actions may stem from entirely different intents. For example, repeated undo actions might indicate confusion or deliberate refinement. Without deeper reasoning, assistance based solely on surface-level actions can lead to shallow or misaligned responses.

To address this challenge, we present \rr{\textbf{\sysname{}} (\textbf{G}UI \textbf{U}ser \textbf{I}ntent \textbf{D}etection \textbf{E}valuation)}, a benchmark designed to evaluate multimodal LLMs (MLLMs) on their ability to understand and assist users in complex software workflows. \sysname{} introduces a three-stage evaluation framework: (1) \textit{Understanding} the user’s behavioral state to identify their current workflow phase; (2) \textit{Reasoning} about their underlying intentions and goals; and (3) \textit{Assisting} by delivering the appropriate form of help at the right moment.

We collected 67.5 hours of screen recordings from 120 human demonstrations across 10 widely used applications---including Photoshop, Figma, PowerPoint, Premiere Pro, and Excel---covering 40 open-ended tasks designed to elicit natural user behavior. Unlike prior work that primarily targets video understanding from expert-recorded instructional videos on closed-ended tasks~\cite{li2020pstuts, lin2024videogui, pmlr-v267-nayak25a, gao2024assistgui, zhao2025worldgui}, our focus is on novice users working on open-ended tasks, with the goal of building collaborative AI systems that assist users during exploration, trial-and-error, and learning. Observing novice workflows allows us to capture authentic moments of confusion, decision-making, and discovery, offering rich opportunities for AI to provide timely, context-aware support. Each session includes both screen recordings and think-aloud narrations that surface the user’s underlying intentions and cognitive states.

Building on this dataset, we define three-staged benchmark tasks: First, \textbf{(i) Behavior State Detection} evaluates whether a model can identify the user’s behavioral state, such as exploration or confusion, based solely on visual cues. To support this, we developed a taxonomy of nine user states reflecting diverse cognitive and behavioral phases in open-ended GUI workflows, grouped into four high-level categories: Planning, Execution, Problem-Solving, and Evaluation (Figure~\ref{fig:taxonomy}). This structure aligns with human cognition and interaction theories~\cite{bloom1956taxonomy, norman1988design}, while introducing finer distinctions tailored to GUI-based task behavior.
Next, \textbf{(ii) Intent Prediction} targets inference of the user’s immediate goal---what they are trying to accomplish in the given moment. The final task, \textbf{(iii) Help Prediction}, assesses whether a model can determine 1) whether the user needs assistance or not, and if so, 2) what type of help would be most appropriate, such as explaining a feature, suggesting an alternative, or addressing an error.
By leveraging both visual screen recordings and accompanying think-aloud narrations, we automatically generated data for each task, which was subsequently verified through human review for accuracy and consistency. 

Evaluation across eight state-of-the-art MLLMs reveals that while current models struggle to interpret user behavior and predict underlying intent and help needed---achieving only 44.6\% accuracy on behavior state detection and 55.0\% on help prediction, performance improves significantly when structured user context is provided. For example, supplying behavioral state and intent information boosted help prediction accuracy by up to 50.2 percent points for the lowest-performing model.

\rr{Our results suggest a promising path forward: providing different layers of human-grounded context, such as behavioral cues, inferred goals, and temporal history, can lead to more accurate assistance decisions. Our benchmark provides a foundation for training and evaluating the next generation of context-aware, collaborative GUI agents.}


\section{Related Work}
\subsection{Video Understanding meets GUI}
Several benchmarks evaluate video understanding in the context of GUI and software workflows. Early work by Li et al.~\cite{li2020pstuts} collected Photoshop tutorial videos to understand screencast videos. More recent datasets span multiple applications and tasks. For example, AssistGUI~\cite{gao2024assistgui} focuses on automating GUI tasks using an actor-critic agent, serving as a benchmark for task-oriented GUI automation. VideoWebArena~\cite{jang2025videowebarena} evaluates long-horizon multimodal agents on web browsing tasks, emphasizing extended video context and web UI interactions. VideoGUI~\cite{lin2024videogui} compiles high-quality instructional screen recordings and introduces a hierarchical model for mapping visual observations to GUI actions. UI-Vision~\cite{pmlr-v267-nayak25a} provides a fine-grained desktop UI video benchmark with dense annotations for perception and interaction. Lastly, WorldGUI~\cite{zhao2025worldgui} increases task diversity by allowing arbitrary initial interface states for each task, challenging agents to handle varied starting conditions. 

These prior benchmarks primarily focus on close-ended tasks with predetermined goals, aiming to replicating expert demonstrations. In contrast, our work targets open-ended GUI workflows with novice users, emphasizing understanding of user intent and context rather than step-by-step replication of actions. This shift toward user-centric evaluation fills a gap not covered by existing GUI video datasets that evaluate task completion or action prediction.

\subsection{Collaborative and Proactive Agents}
While GUI agents that automate interface operations based on a given goal or instruction can be effective, this fully autonomous approach can conflict with the needs of users in creative or analytical environments, where retaining control and exploring alternatives are essential.
To address this, recent research has shifted toward assistive GUI agents that collaborate with users by understanding context and offering timely support. Several works have explored inferring user goals and intent in both web~\cite{pawar2025earl} and software environments~\cite{usergoal25, kiran21intent, zhao2025proactivevaproactivevisualanalytics} to better align assistance with user needs. For example, Zhao et al.~\cite{zhao2025proactivevaproactivevisualanalytics} introduce ProactiveVA, a visual analytics agent that monitors user interactions and leverages LLMs to detect when users may be stuck, providing context-sensitive suggestions or guidance.

Several recent works in the Human-Computer Interaction (HCI) community explore this shift toward collaboration and contextual support. CowPilot~\cite{Huq2025CowPilot} proposes a mixed-initiative framework that enables users to share control with an autonomous web navigation agent, improving efficiency while preserving agency. In programming, proactive assistants like Codellaborator~\cite{Pu2025Assistance} and NeedHelp~\cite{Chen2025NeedHelp} demonstrate how real-time intervention can aid users when well-timed. Studies on software applications~\cite{Khurana2025DoItWithMe} show users prefer AI agents that guide them rather than take over entirely, reinforcing the need for transparency and shared control. ProMemAssist~\cite{Pu2025ProMemAssist} further highlights the benefits of modeling user cognition to deliver timely, non-intrusive support. These findings echo broader discussions on autonomy levels~\cite{Feng2025Autonomy} and the importance of aligning agent behavior with human preferences~\cite{Li2020GUI, Khurana2024LLMHelp}. Our work builds on these insights, evaluating how well current multimodal models can perceive a user's state and intentions in GUI workflow recordings and decide if and how to assist. By situating the evaluation in real user workflows, we aim to push GUI agents toward true user-aware collaboration.

\section{\sysname{} Benchmark}

\begin{figure*}[ht]
  \centering
  \includegraphics[width=\linewidth]{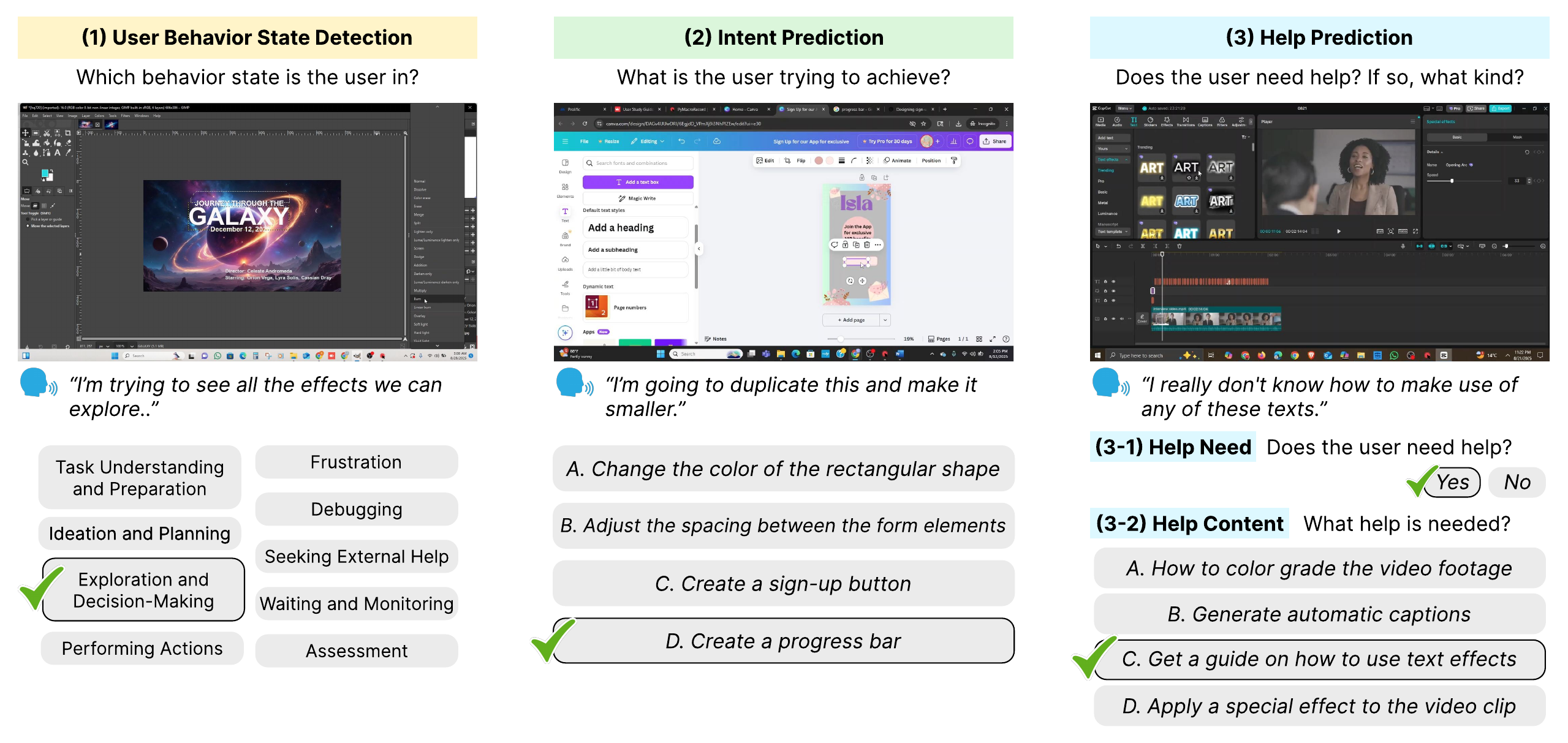}
  \caption{
  Overview of the three core tasks in the \sysname{} benchmark.
  (1) \textbf{User Behavior State Detection} identifies the user's current behavioral mode (e.g., \textit{Exploration and Decision-Making}).
  (2) \textbf{Intent Prediction} infers what the user is trying to achieve (e.g., \textit{Create a progress bar}).
  (3) \textbf{Help Prediction} determines whether the user needs assistance and, if so, what kind of help is relevant (e.g., \textit{Get a guide on how to use text effects}).
  Together, these tasks enable a comprehensive understanding of user behavior and assistance needs in software GUI environments. 
  We evaluate MLLMs on their ability to infer these solely from the visual input, without access to the demonstrator’s narration --- a setting that closely reflects real-world use.
  }
  \label{fig:main}
\end{figure*}

To develop a benchmark that focuses on understanding and assisting users, we collected demonstrations from novice users. Unlike existing datasets that focus primarily on expert demonstrations or polished instructional videos~\cite{li2020pstuts, lin2024videogui, pmlr-v267-nayak25a, gao2024assistgui, zhao2025worldgui, lu2025videoagenttrekcomputerusepretraining, wu25guinarrator}, our dataset captures the authentic challenges and exploratory behaviors that novices exhibit during task completion, serving a crucial role in building collaborative agents.
Building on these demonstrations, we propose a suite of tasks designed to evaluate models’ capabilities to understand users and provide effective assistance.

\subsection{Video Collection}
We collected 120 demonstrations from novice users across 10 applications spanning five categories: Photo Editing (Photoshop, GIMP), Graphic Design (Figma, Canva), Presentation Design (PowerPoint, Google Slides), Video Editing (Premiere Pro, CapCut), and Data Analysis (Google Sheets, Microsoft Excel). For each application, we designed four open-ended tasks 
aimed at eliciting natural and diverse user behaviors and approaches 
(Table~\ref{tab:task_descriptions} in supp.).

We chose creative and analytical tools to surface exploratory workflows and variation in problem-solving strategies. 
Each task was completed by three different users to capture diverse strategies and behaviors. We ensured that each task was flexible enough, while still incorporating elements of challenge. Participants were asked to spend at least 20 minutes per task and meet a few minimal requirements (e.g., inserting a relevant image) to mark it as complete.

We recruited 54 novice users of software from Prolific and our institution. 
\rr{Participants were screened based on self-reported expertise to ensure novice-level familiarity with the features in the target application (Mean: 2.8, SD: 1.1, Range: 1–5).}
During the study, participants worked on the assigned task while recording their screen and keyboard/mouse input events. They were also asked to think aloud and record their voice as they carried out the task, verbalizing what they were doing and their thought process. 

\subsection{Benchmark Tasks}
To evaluate a model’s ability to understand user context and deliver appropriate assistance, we design our benchmark as a unified three-stage framework: 
\textit{Understanding} $\rightarrow$ \textit{Reasoning} $\rightarrow$ \textit{Assisting}. 
These stages progress from interpreting user behavior to inferring intentions and ultimately providing helpful assistance. 
Each task corresponds to a distinct level of cognitive inference required for a human-assisting GUI agent to effectively support users in open-ended software workflows.

To construct a dataset for task evaluation, we used the Human-AI collaborative method. We first transcribed the think-aloud narration using WhisperX~\cite{bain2022whisperx}, and used the narration as a main source for extracting initial annotations in addition to the video. 
We employed \textit{Gemini-2.5-Pro}~\cite{Gemini2025} to first create annotations needed for each task, which were then refined by human annotators. 
Note that we use narration only as an annotation source to capture users' intentions and mental states. The benchmark evaluates vision-only understanding, testing whether models can infer these states solely from visual cues, as in real-world settings \rr{with limited access to user speech}.

\begin{figure*}[ht]
  \centering
  \includegraphics[width=0.95\textwidth]{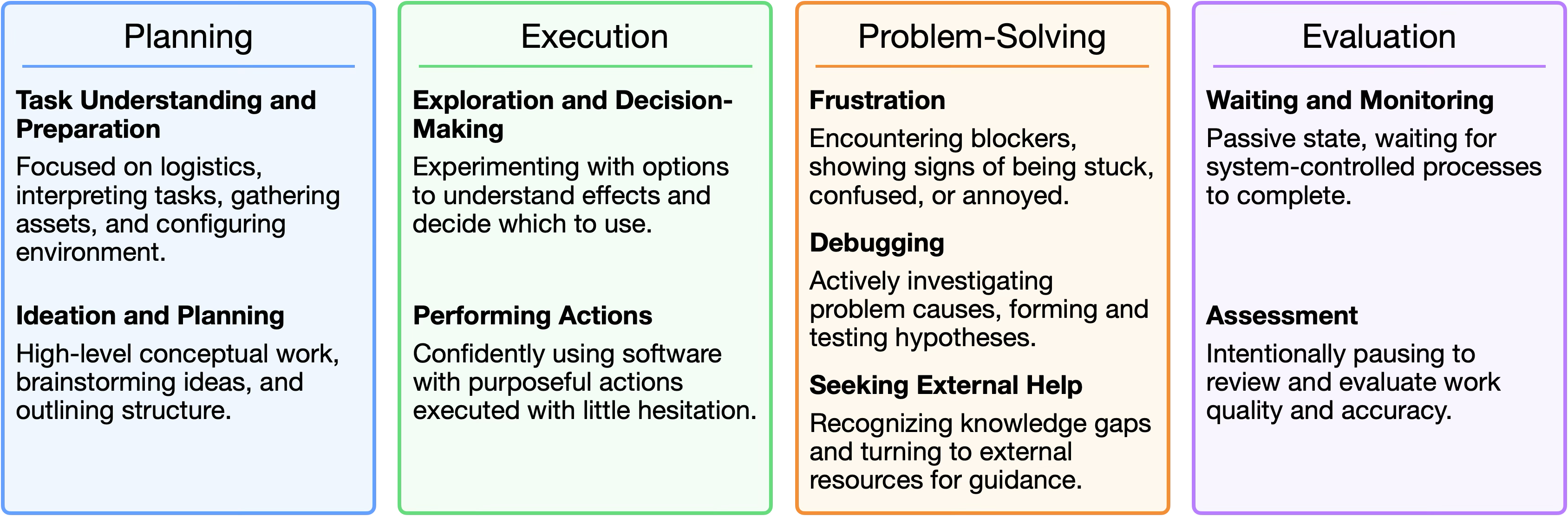}
  \caption{
    Our proposed taxonomy of user behavior states in GUI-based software tasks, 
    organized into four main phases: \textbf{Planning}, \textbf{Execution}, 
    \textbf{Problem-Solving}, and \textbf{Evaluation}. 
    Each phase captures distinct patterns of user cognition and interaction, 
    from initial goal formulation to iterative action, troubleshooting, and reflection.
    }
  \label{fig:taxonomy}
\end{figure*}

\subsubsection{User Behavior State Detection}

\noindent\textbf{{Description.}} 
This task evaluates whether a model can interpret the user's behavioral context directly from visual cues. Models are asked to classify a video segment into one of nine behavior states in our taxonomy (Figure~\ref{fig:taxonomy}), 
which spans the full range of cognitive and behavioral processes observed in creative and analytical workflows.

We developed the taxonomy through a multi-stage, human–AI collaborative process ~\cite{humanAICollabTaxonomy}. First, three authors iteratively created and consolidated an initial taxonomy over five sessions based on observations of online software task videos. Separately, we prompted \textit{Gemini-2.5-Pro} to generate a taxonomy from scratch using our collected video dataset, without providing our initial version. We then augmented the human-generated taxonomy by integrating novel categories identified by the LLM. Finally, the combined taxonomy was validated against the entire video dataset to ensure comprehensive coverage and reorganized into the final set of nine distinct states.
\rr{Our taxonomy aligns with Norman's Seven Stages of Action~\cite{norman1988design}, mapping Planning, Execution, and Evaluation to goal formation, action, and outcome assessment, and draws on Bloom's cognitive hierarchy~\cite{bloom1956taxonomy} that captures the shifts between operational (Execution) and critical work (Evaluation).}


\noindent\textbf{{Dataset Curation.}} After constructing the taxonomy, we aligned each video with its corresponding narration segments. For every segment, we annotated the user’s behavior state using \textit{Gemini-2.5-Pro} according to the taxonomy, prompting the model to produce both a predicted label and its reasoning. Two human annotators recruited from Prolific then verified and refined these annotations, achieving a 96.1\% agreement rate. Finally, we uniformly sampled 200 instances from each of the nine classes, resulting in a balanced dataset of 1.8K annotated segments.

\subsubsection{Intent Prediction}
\noindent\textbf{{Description.}} 
This task evaluates whether a model can reason about the user's short-term, immediate goal in context. It focuses on identifying what the user aims to achieve within open-ended workflows.

\noindent\textbf{{Dataset Curation.}}
Using the narration-aligned video segments, we prompted \textit{Gemini-2.5-Pro} to infer users' intention in each segment. The think-aloud narrations often revealed users’ goals (e.g., \textit{``I’m going to align these objects'', ``I’ll try another color''}). Leveraging this signal, we prompted the model to infer the underlying user intention. After collecting and deduplicating the inferred intents, we further instructed the model to generate three plausible but incorrect alternatives to serve as distractors for the multiple-choice evaluation. The resulting intent annotations and distractors were then validated by the authors, with 88.68\% of the data retained, yielding a final set of 1.3K instances.

\subsubsection{Help Prediction}
\noindent\textbf{{Description.}} 
The final task evaluates whether a model can progress from understanding and reasoning to deciding how to assist. 
Help Prediction consists of two subtasks: 
(1) \textbf{Help Need Detection}, a binary classification task that determines whether the user needs help, and 
(2) \textbf{Help Content Prediction}, which identifies the specific type of help needed, such as explaining a feature or suggesting an alternative.
Together, these subtasks assess a model's ability to anticipate user needs and recommend appropriate assistance, bridging the gap between perception and actionable support.

\noindent\textbf{{Dataset Curation.}}
We identified potential help-seeking moments using two complementary signals.
First, \textit{explicit help-seeking} behaviors, such as switching to external resources (e.g., Google, YouTube, ChatGPT), indicated direct attempts to seek guidance.
Second, \textit{implicit help-seeking} cues were extracted from user narration, where they expressed uncertainty or confusion (e.g., \textit{``How do I align this?''}, \textit{``I can't find Layer Mask.''}).
Additionally, we included clear \textit{no-help-needed} moments, where users demonstrated confidence through their narration.
Using these signals, \textit{Gemini-2.5-Pro} was prompted to generate initial annotations for help-need and help-content labels. After deduplication, the model was additionally prompted to generate three plausible but incorrect options for each instance for multiple-choice question evaluation.
All annotations and distractors were then reviewed by the authors, resulting in 1K validated instances, with 78.89\% of the original data retained. For 12.5\% of the retained instances, the segment's start or end time was adjusted to exclude explicit visual help signals (e.g., user turning to Google Search) to ensure fair evaluation. Overall, 66\% of the instances were labeled as help-needed, while the remaining 34\% required no help.

\section{Experiments}

\begin{table*}[ht]
\centering
\footnotesize
\setlength{\tabcolsep}{2pt}
\renewcommand{\arraystretch}{1.2}
\begin{tabular}{l cc cc @{\hspace{8pt}} ccc @{\hspace{8pt}} ccc}
\toprule
\multirow{3}{*}{\textbf{Model}} &
\multicolumn{2}{c}{\textbf{(1) Behavior Detection}} &
\multicolumn{2}{c}{\textbf{(2) Intent Prediction}} &
\multicolumn{6}{c}{\textbf{(3) Help Prediction}} \\
\cmidrule(lr){2-3}\cmidrule(lr){4-5}\cmidrule(lr){6-11}
 & \textit{--} & \textit{+ Prev.} &
 \textit{--} & \textit{+ Behavior} &
 \multicolumn{3}{c}{\textbf{Help Need Detection}} &
 \multicolumn{3}{c}{\textbf{Help Content Prediction}} \\
\cmidrule(lr){6-8}\cmidrule(lr){9-11}
 &  &  &
  &  &
 \textit{--} & \textit{+ Behv.} & \textit{+Behv.+Intent} &
 \textit{--} & \textit{+ Behv.} & \textit{+Behv.+Intent} \\
\midrule
Gemini-2.5-Flash~\cite{Gemini2025}        & 36.91 & 38.19 & 65.40 & 66.77 & 53.64 & 76.33 & 78.07 & 49.53 & 53.75 & 78.59 \\
Gemini-2.5-Pro~\cite{Gemini2025}          & 42.44 & 43.79 & 67.80 & 70.16 & \textbf{69.82} & 84.73 & 82.38 & 52.74 & 57.03 & 79.69 \\
GPT-4o-mini~\cite{gpt4o}                  & 17.65 & 17.07 & 60.76 & 62.19 & 46.05 & 78.92 & 82.26 & 31.32 & 42.86 & 79.84 \\
GPT-4o~\cite{gpt4o}                       & 36.32 & 37.24 & 61.19 & 62.58 & 49.69 & \textbf{87.79} & \textbf{87.91} & 45.95 & 48.37 & 79.78 \\
Claude-4.5-Sonnet~\cite{Anthropic2025Claude4.5} & \textbf{44.61} & \textbf{45.63} & \textbf{71.39} & \textbf{72.62} & 39.49 & 58.56 & 59.43 & \textbf{55.00} & \textbf{62.17} & \textbf{82.79} \\
Qwen3-VL-8B~\cite{QwenTeam2025Qwen3}      & 37.97 & 38.13 & 62.70 & 64.03 & 52.83 & 70.39 & 77.36 & 46.06 & 50.63 & 80.11 \\
InternVideo2.5-8B~\cite{Wang2025InternVideo2.5} & 21.57 & 27.02 & 43.79 & 45.13 & 34.36 & 35.35 & 35.25 & 23.67 & 29.15 & 73.86 \\
InternVL3-8B~\cite{Zhu2025InternVL3}      & 22.57 & 24.90 & 46.11 & 46.97 & 34.94 & 43.73 & 46.82 & 27.03 & 32.20 & 72.97 \\
\bottomrule
\end{tabular}
\caption{Evaluation results on accuracy across (1) Behavior State Detection, (2) Intent Prediction, and (3) Help Prediction.}
\label{tab:results_main}
\end{table*}

\subsection{Experimental Setup}
We evaluate a range of multimodal large language models (MLLMs) on our benchmark to assess their ability to understand, reason about, and assist users in open-ended software workflows. Our evaluation includes eight representative MLLMs spanning both proprietary and open-source models: \textbf{Gemini-2.5-Flash}~\cite{Gemini2025}, \textbf{Gemini-2.5-Pro}~\cite{Gemini2025}, \textbf{GPT-4o-mini}~\cite{gpt4o}, \textbf{GPT-4o}~\cite{gpt4o}, \textbf{Claude-4.5-Sonnet}~\cite{Anthropic2025Claude4.5}, \textbf{Qwen3-VL-8B}~\cite{QwenTeam2025Qwen3}, \textbf{InternVideo2.5-Chat-8B}~\cite{Wang2025InternVideo2.5}, and \textbf{InternVL3-8B}~\cite{Zhu2025InternVL3}. All models are evaluated in a zero-shot setting using publicly available APIs or checkpoints, without any additional fine-tuning.

For each test instance, we uniformly sample 32 frames from the corresponding video segment, providing only visual input (excluding narration audio) to simulate perception based solely on visual cues. To ensure consistency across models, we use standardized prompting templates (Section~\ref{sec:prompts}). We also prompt models to generate both a predicted label and supporting reasoning, a strategy shown to improve task performance~\cite{kojima22large}.

Our main experiments are conducted in an offline inference setting, where models solve the task given the full video. To approximate real-world proactive assistant scenarios, we additionally evaluate an online setting, where the model receives visual input progressively---at 25\%, 50\%, 75\%, and 100\% of the segment, we uniformly sample 32 frames from the corresponding prefix for inference.

\begin{table*}[ht]
\centering
\small
\setlength{\tabcolsep}{4pt}
\renewcommand{\arraystretch}{1.2}
\begin{tabular}{l cccc @{\hspace{8pt}} cccc @{\hspace{8pt}} cccc}
\toprule
\multirow{3}{*}{\textbf{Model}} &
\multicolumn{12}{c}{\textbf{Help Need Detection}} \\
\cmidrule(lr){2-13}
 & \multicolumn{4}{c}{\textit{--}} &
   \multicolumn{4}{c}{\textit{+ Behavior State}} &
   \multicolumn{4}{c}{\textit{+ Behavior State + Intent}} \\
\cmidrule(lr){2-5}\cmidrule(lr){6-9}\cmidrule(lr){10-13}
 & \textit{Acc} & \textit{Prec} & \textit{Rec} & \textit{F1} &
   \textit{Acc} & \textit{Prec} & \textit{Rec} & \textit{F1} &
   \textit{Acc} & \textit{Prec} & \textit{Rec} & \textit{F1} \\
\midrule
Gemini-2.5-Flash~\cite{Gemini2025}  & 53.64 & 83.27 & 36.62 & 50.87 & 76.33 & 97.67 & 65.47 & 78.39 & 78.07 & 94.56 & 70.62 & 80.86 \\
Gemini-2.5-Pro~\cite{Gemini2025}    & \textbf{69.82} & 76.42 & \textbf{78.09} & \textbf{77.42} & 84.73 & 93.61 & 82.34 & 87.61 & 82.38 & 91.20 & 80.94 & 86.76 \\
GPT-4o-mini~\cite{gpt4o}           & 46.05 & 83.03 & 22.31 & 35.17 & 76.73 & 97.61 & 66.23 & 78.92 & 79.71 & 97.20 & 71.29 & 82.26 \\
GPT-4o~\cite{gpt4o}      & 49.69 & 74.41 & 35.14 & 47.73 & \textbf{87.79} & 95.39 & \textbf{85.53} & \textbf{90.19} & \textbf{87.91} & 95.12 & \textbf{85.95} & \textbf{90.30} \\
Claude-4.5-Sonnet~\cite{Anthropic2025Claude4.5} & 39.49 & \textbf{87.69} & 8.92 & 16.19 & 58.56 & \textbf{99.16} & 37.09 & 53.99 & 59.43 & \textbf{99.19} & 38.44 & 55.41 \\
Qwen3-VL-8B~\cite{QwenTeam2025Qwen3} & 52.83 & 79.86 & 34.23 & 47.92 & 70.39 & 94.35 & 58.50 & 72.22 & 77.36 & 95.38 & 67.56 & 79.09 \\
InternVideo2.5-8B~\cite{Wang2025InternVideo2.5} & 34.36 & 33.33 & 0.16 & 0.31 & 35.35 & 90.91 & 1.56 & 3.07 & 35.25 & 83.33 & 1.56 & 3.07   \\
InternVL3-8B~\cite{Zhu2025InternVL3} & 34.94 & 72.73 & 1.25 & 2.46 & 43.73 & 98.88 & 15.77 & 27.20 & 46.82 & 98.40 & 19.22 & 32.16 \\
\bottomrule
\end{tabular}
\caption{Results for \textbf{Help Need Detection} on accuracy, precision, recall, and F1-score across three conditions (default, with behavior state, with behavior state and intent).}
\label{tab:help_need_detection}
\end{table*}

\subsection{Evaluation Tasks}
\paragraph{(1) Behavior State Detection.}
This task measures whether a model can identify the user's behavioral state from a given video segment. 
We provide each model with clips and ask it to classify them into one of nine taxonomy-defined states. 
Two configurations are tested: (\textit{i}) using only the current segment and (\textit{ii}) with prior history, where the model is given the immediately preceding segment's behavior state. 
This is framed as a multi-class classification problem, and performance is evaluated using accuracy.

\paragraph{(2) Intent Prediction.}
This task evaluates a model's ability to infer the user's underlying goal within a given video segment. Models are prompted to predict what the user's goal in two settings: (\textit{i}) using only the current segment, and (\textit{ii}) with additional behavior state context, where the model is also given the state label and its definition. 
We adopt a multiple-choice question (MCQ) format, where the model selects the most likely intent from four candidates. Performance is measured using accuracy. For the default setting (\textit{i}), \rr{to mitigate potential bias,} we additionally report multi-binary accuracy (MBAcc) following prior work~\cite{cai2024temporalbench, cho2025PerceptionLM,cheng2025tempura}, which evaluates whether the model correctly identifies the ground-truth intent in all three pairwise comparisons against incorrect alternatives.

\paragraph{(3) Help Prediction.}
The final task evaluates whether models can move beyond understanding and reasoning to provide actionable assistance. Given a video segment, models are asked to predict whether the user requires help (\textit{Need}), and if so, what kind of help would be most appropriate (\textit{Content}). \textbf{Help Need Detection} is framed as a binary classification task and evaluated using accuracy, precision, recall, and F1-score. \textbf{Help Content Prediction}, similar to Intent Prediction, uses a multiple-choice question (MCQ) format and is evaluated using accuracy and multi-binary accuracy (MBAcc) for the default setting.
We test three settings for both tasks: (\textit{i}) video only, (\textit{ii}) video + behavior state, where the model is given the behavior label and its definition for the current segment, and (\textit{iii}) video + behavior state + intent, where the model additionally receives the identified user intention. These settings progressively assess the model’s ability to leverage layered user context for meaningful, situation-aware assistance.

\subsection{Results}
Table~\ref{tab:results_main} presents the performance of baseline models on \sysname{} across the tasks, with accuracies reported under default and context-augmented settings. Overall, models performed weakest on Behavior State Detection and Help Prediction, with default-setting accuracies peaking at 44.61\% and 55.00\% for Behavior State Detection and Help Content Prediction, respectively, both from Claude-4.5-Sonnet~\cite{Anthropic2025Claude4.5}. While Gemini-2.5-Pro~\cite{Gemini2025} reached nearly 70\% accuracy on Help Need Detection, most other models showed substantially lower performance across both Help sub-tasks. Across tasks, we observe that models generally benefit from added behavioral and intent context, with particularly notable improvements in help-related predictions. We report the main findings below.

\subsubsection{Behavior State Detection}\label{sec:results_behavior_state}

\paragraph{Behavior State Detection remains highly challenging.}
All models struggled to accurately infer the user’s behavioral state from video segments, underscoring the difficulty of the 9-way classification task. While proprietary models such as Claude-4.5-Sonnet~\cite{Anthropic2025Claude4.5} and Gemini-2.5-Pro~\cite{Gemini2025} performed best, no model surpassed 45\% accuracy, and most fell below 40\%.

\paragraph{Models often misinterpret signals of struggle.}
The most common failure was misclassifying \textit{Frustration} or \textit{Debugging} as \textit{Performing Actions} or \textit{Exploration and Decision-Making}\rr{, as shown in the confusion matrix} (Figure~\ref{fig:taxonomy_result_confusion} in supp.). 
\rr{ These errors suggest that models overlook subtle indicators of user difficulty---e.g., repeated clicks, hesitation, or undoing actions---instead interpreting them as deliberate progress, revealing a lack of nuanced understanding of user frustration signals.}

\paragraph{Temporal context shows modest potential.}
Incorporating the prior behavior state led to small but consistent gains across models. While most improvements were marginal, the largest gain was observed for InternVideo2.5-8B~\cite{Wang2025InternVideo2.5} with 5.45 percentage points, suggesting that temporal context holds value and may be more effectively utilized with improved temporal reasoning capabilities.



\begin{figure*}[ht]
  \centering
  \includegraphics[width=\textwidth]{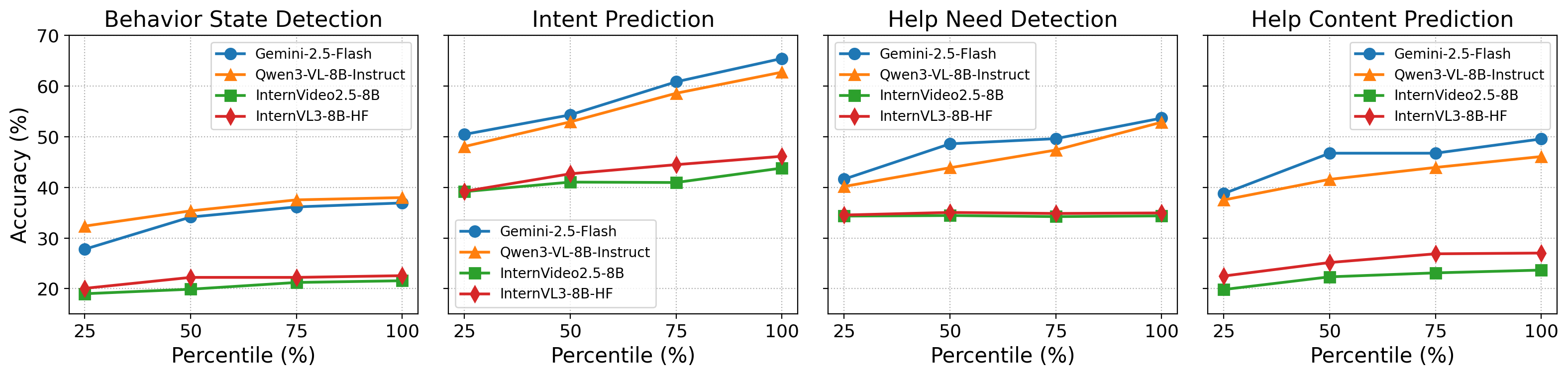}
  \caption{\rr{Accuracy trends across the tasks in the online setting, where models are given progressively more of the video segment (25\%, 50\%, 75\%, and 100\%). Models show consistent improvement as they see more segments, with Gemini-2.5-Flash~\cite{Gemini2025} and Qwen3-VL-8B~\cite{QwenTeam2025Qwen3} showing larger and more consistent gains across all four tasks compared to the smaller open-source models.}}
  \label{fig:online}
\end{figure*}

\begin{table}[ht]
\centering
\small
\setlength{\tabcolsep}{4pt}
\renewcommand{\arraystretch}{1.2}
\begin{tabular}{l cc @{\hspace{8pt}} cc}
\toprule
\multirow{2}{*}{\textbf{Model}} &
\multicolumn{2}{c}{\textbf{Intent Prediction}} &
\multicolumn{2}{c}{\textbf{Help Prediction}} \\
\cmidrule(lr){2-3}\cmidrule(lr){4-5}
 & \textit{Acc} & \textit{MBAcc} & \textit{Acc} & \textit{MBAcc} \\
\midrule
Gemini-2.5-Flash~\cite{Gemini2025}        & 65.40 & 59.09 & 49.53 & 44.69 \\
Gemini-2.5-Pro~\cite{Gemini2025}          & 67.80 & 64.34 & 52.74 & 45.31 \\
GPT-4o-mini~\cite{gpt4o}                  & 60.76 & 50.24 & 31.32 & 28.59  \\
GPT-4o~\cite{gpt4o}                       & 61.19 & 56.58 & 45.95 & 41.25 \\
Claude-4.5-Sonnet~\cite{Anthropic2025Claude4.5} & \textbf{71.39} & \textbf{65.44} & \textbf{55.00} & \textbf{50.78} \\
Qwen3-VL-8B~\cite{QwenTeam2025Qwen3} & 62.70 & 58.07 & 46.06 & 44.69 \\
InternVideo2.5-8B~\cite{Wang2025InternVideo2.5} & 43.79 & 27.98 & 23.67 & 18.75 \\
InternVL3-8B~\cite{Zhu2025InternVL3} & 46.11 & 40.75 & 27.03 & 23.75 \\
\bottomrule
\end{tabular}
\caption{Evaluation of \textbf{Intent Prediction} and \textbf{Help Content Prediction}, with Accuracy (Acc) and Multi-Binary Accuracy (MBAcc).}
\label{tab:results_mbacc}
\end{table}

\subsubsection{Intent Prediction}

\paragraph{Intent Prediction is the most tractable task, but still imperfect.} 
Among the three tasks, models achieved the highest performance on intent prediction, with several surpassing 60\% accuracy. However, performance drops under the stricter MBAcc metric, which requires consistent discrimination across all answer pairs. This indicates that while models can often select a plausible intent, they still struggle with reliably identifying the correct one over all distractors (Table~\ref{tab:results_mbacc}).


\paragraph{Behavior context helps, but only slightly.} 
Incorporating behavior state context (i.e., the user’s behavioral label and definition) consistently improved performance, but the gains were relatively modest across all models. This suggests that while such context may offer useful cues, it does not provide sufficient information on its own or is not yet effectively leveraged by current models for intent inference.

\subsubsection{Help Prediction}

\paragraph{High variance and missed help cases in Need Detection.}
Table~\ref{tab:help_need_detection} shows the full performance results for Help Need Detection. This subtask exhibited the most variance across models, with F1 scores ranging from 0.31 (InternVideo2.5-8B~\cite{Wang2025InternVideo2.5}) to 77.42 (Gemini-2.5-Pro~\cite{Gemini2025}). Notably, recall was particularly low across most models---except for Gemini-2.5-Pro, all others had recall under 37\%. This indicates that many instances where users actually needed help were misclassified as not needing it, echoing similar trends in Behavior State Detection (Section~\ref{sec:results_behavior_state}) where models frequently misinterpreted signals of struggle.

\paragraph{Behavior context improves Help Need Detection.}
Providing the user’s behavior state led to consistent and significant improvements in Help Need Detection across all models, with the largest gain observed in GPT-4o~\cite{gpt4o}, with a 42.46-point increase in F1 score. This suggests that context, such as whether a user is exploring or showing signs of frustration, provides strong cues for determining help needs.

\paragraph{Help Content Prediction remains challenging, but benefits from intent context.}
Help Content Prediction proved particularly challenging, with all models struggling and the top accuracy reaching only 55\% from Claude-4.5-Sonnet~\cite{Anthropic2025Claude4.5}, which further declined to around 50\% under the stricter MBAcc evaluation. However, incorporating user intent led to substantial improvements across models, with the largest gain in InternVideo2.5-8B~\cite{Wang2025InternVideo2.5} at 50.19 percentage points, highlighting the importance of understanding both user state and intent for providing targeted support.

\subsubsection{Other Findings}

\paragraph{Online vs. Offline Setting: models benefit more from temporal context.}
In our online simulation experiment, where models are given progressively more of the video segment (25\%, 50\%, 75\%, and 100\%), we observe consistent performance gains across all four tasks (Figure~\ref{fig:online}). 
\rr{
Gemini-2.5-Flash~\cite{Gemini2025} and Qwen3-VL-8B~\cite{QwenTeam2025Qwen3} show larger and more consistent gains across all tasks, compared to the smaller open-source models, indicating a strong ability to integrate growing context into more accurate predictions.}
These findings suggest that gathering appropriate context over time is crucial for proactive AI assistance, where systems must not only react but also anticipate user needs based on incomplete and evolving information.

\rr{
\paragraph{Outlook for Model Improvements.}
Together, these results suggest that incorporating structured user context (behavior state and intent) and temporal context consistently improves help prediction. Recent work on agents demonstrates the effectiveness of context engineering via stratified memory, where interaction history is selectively structured rather than treated as a flat sequence~\cite{yang2026groundingagentmemorycontextual}. Applying this idea to GUI assistance is a promising direction for better leveraging long-horizon user context.}
\section{Conclusion}
We introduced a benchmark for evaluating models in understanding, reasoning about, and assisting users in open-ended GUI-based workflows. Grounded in real-world novice user demonstrations, our tasks---behavior state detection, intent prediction, and help prediction---capture core capabilities needed for collaborative GUI agents.
Evaluation across state-of-the-art MLLMs revealed that models struggle to interpret nuanced user behavior and accurately infer assistance needed in open-ended scenarios. However, when provided with appropriate user context, 
models showed consistent improvements, highlighting the value of structured user understanding.
Overall, our benchmark provides a foundation for user-aware agents that support human workflows.


\section*{Acknowledgements}
\rr{This work was supported by the Institute of Information \& Communications Technology Planning \& Evaluation (IITP) grant funded by the Korean government (MSIT) (No. 2021-0-01347, Video Interaction Technologies Using Object-Oriented Video Modeling and No. RS-2024-00443251, Accurate and Safe Multimodal, Multilingual Personalized AI Tutors).}

{
    \small
    \bibliographystyle{ieeenat_fullname}
    \bibliography{main}
}

\clearpage
\setcounter{page}{1}
\maketitlesupplementary

\renewcommand{\thesection}{\Alph{section}}
\renewcommand{\thetable}{\Alph{section}\arabic{table}}
\renewcommand{\thefigure}{\Alph{section}\arabic{figure}}
\setcounter{section}{0}
\setcounter{table}{0}
\setcounter{figure}{0}

\section{Detailed Evaluation Metrics}
\label{sec:supp_metrics}

In this section, we provide the formal definitions for the evaluation metrics used across our four evaluation tasks: Behavior State Detection, Intent Prediction, Help Need Detection, and Help Content Prediction. Let $N$ denote the total number of test samples in the dataset. For the $i$-th sample, let $y_i$ denote the ground-truth label and $\hat{y}_i$ denote the model's predicted label. $\mathbb{I}(\cdot)$ denotes the indicator function, which equals 1 if the condition inside is true and 0 otherwise.

\subsection{Metric Definitions by Task}

\subsubsection{Task 1: Behavior State Detection}
This task is formulated as a multi-class classification problem where the model must classify a video segment into one of 9 distinct behavioral states. We evaluate performance using standard \textbf{Accuracy}.

\begin{equation}
    \text{Accuracy} = \frac{1}{N} \sum_{i=1}^{N} \mathbb{I}(\hat{y}_i = y_i)
\end{equation}

\subsubsection{Task 2: Intent Prediction}
This task is framed as a Multiple-Choice Question (MCQ) task with 4 options (1 correct answer and 3 distractors). We use two metrics:

\paragraph{Accuracy.}
Measures the proportion of instances where the model selects the correct intent option from the four candidates.
\begin{equation}
    \text{Accuracy} = \frac{1}{N} \sum_{i=1}^{N} \mathbb{I}(\hat{y}_i = y_i)
\end{equation}

\paragraph{Multi-Binary Accuracy (MBAcc).}
Following prior work~\cite{cai2024temporalbench, cho2025PerceptionLM}, we employ MBAcc to evaluate robustness against distractors. For a given sample $i$, let $y_i$ be the correct option and $\mathcal{C}_i^{-} = \{c_{i,1}, c_{i,2}, c_{i,3}\}$ be the set of three incorrect distractor options. The model performs a pairwise comparison function $f(x, \text{opt}_A, \text{opt}_B)$ which returns the chosen option between A and B. A prediction is considered correct under MBAcc only if the model prefers the ground truth $y_i$ over \textit{every} distractor in $\mathcal{C}_i^{-}$.

\begin{equation}
    \text{MBAcc} = \frac{1}{N} \sum_{i=1}^{N} \left( \prod_{c \in \mathcal{C}_i^{-}} \mathbb{I}(f(x_i, y_i, c) = y_i) \right)
\end{equation}

\subsubsection{Task 3-1: Help Prediction (Need Detection)}
This sub-task is a binary classification problem (Help Needed vs. Not Needed). We evaluate this using Accuracy, Precision, Recall, and F1-Score. Let $TP$ (True Positive), $TN$ (True Negative), $FP$ (False Positive), and $FN$ (False Negative) denote the classification counts.

\begin{itemize}
    \item \textbf{Accuracy}: The ratio of correctly predicted observations to total observations.
    \begin{equation}
        \text{Accuracy} = \frac{TP + TN}{TP + TN + FP + FN}
    \end{equation}

    \item \textbf{Precision}: The ratio of correctly predicted positive observations to the total predicted positives.
    \begin{equation}
        \text{Precision} = \frac{TP}{TP + FP}
    \end{equation}

    \item \textbf{Recall}: The ratio of correctly predicted positive observations to the all observations in the actual class.
    \begin{equation}
        \text{Recall} = \frac{TP}{TP + FN}
    \end{equation}

    \item \textbf{F1-Score}: The harmonic mean of Precision and Recall.
    \begin{equation}
        \text{F1} = 2 \cdot \frac{\text{Precision} \cdot \text{Recall}}{\text{Precision} + \text{Recall}}
    \end{equation}
\end{itemize}

\subsubsection{Task 3-2: Help Prediction (Content Prediction)}
Similar to Intent Prediction, this sub-task is an MCQ task where the model must select the appropriate help content. It is evaluated using \textbf{Accuracy} and \textbf{Multi-Binary Accuracy (MBAcc)}.

\paragraph{Accuracy.}
\begin{equation}
    \text{Accuracy} = \frac{1}{N} \sum_{i=1}^{N} \mathbb{I}(\hat{y}_i = y_i)
\end{equation}

\paragraph{Multi-Binary Accuracy (MBAcc).}
Defined identically to the Intent Prediction task. Let $\mathcal{C}_i^{-}$ be the set of incorrect help content options for the $i$-th sample.
\begin{equation}
    \text{MBAcc} = \frac{1}{N} \sum_{i=1}^{N} \left( \prod_{c \in \mathcal{C}_i^{-}} \mathbb{I}(f(x_i, y_i, c) = y_i) \right)
\end{equation}

\clearpage
\section{Dataset Details}
We provide a comprehensive overview of the \sysname{} dataset, detailing its statistical properties, task granularity, and the diverse range of software workflows it encompasses.

\subsection{Dataset Statistics}
\label{sec:dataset_stats}

\sysname{} comprises a comprehensive collection of 120 screen recording videos, totaling approximately 67.5 hours of footage. A key characteristic of our dataset is the inclusion of rich verbal narration; as shown in Table~\ref{tab:dataset_stats}, think-aloud narration covers \textbf{78\% of the total video duration}, providing high-quality ground truth for annotating user intent and mental states.

\begin{table}[h]
\centering
    \begin{tabular}{lc}
    \toprule
    \textbf{Variable} & \textbf{Value} \\
    \midrule
    \# Videos & 120 \\
    Total Duration & 67.5 hours \\
    Avg. Duration & 33 min 44 sec \\
    Max Duration & 1 hour 23 min 50 sec \\
    Min Duration & 16 min 42 sec \\
    Think-Aloud Narration Ratio & 78\% \\
    \midrule
    \multicolumn{2}{l}{\textit{Task Samples \& Granularity}} \\
    (1) Behavior State Detection & 1.8K \\
    \quad \quad \textit{Avg. Segment Length} & 14.16s \\
    (2) Intent Prediction & 1.3K \\
    \quad \quad \textit{Avg. Segment Length} & 25.40s \\
    (3) Help Prediction & 1K \\
    \quad \quad \textit{Avg. Segment Length} & 25.56s \\
    \bottomrule
    \end{tabular}
    \caption{Statistics of the \sysname{} dataset.}
    \label{tab:dataset_stats}
\end{table}

\begin{figure}[h]
    \centering
    \includegraphics[width=\columnwidth]{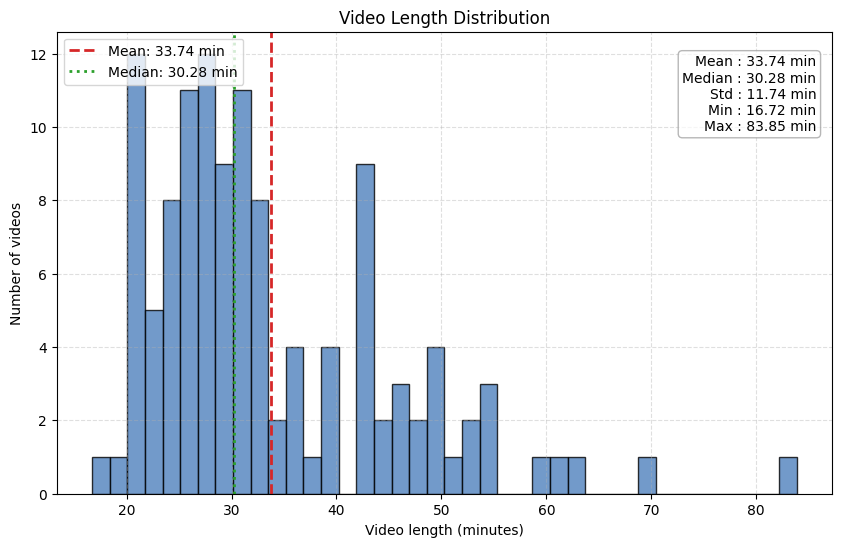}
    \caption{Distribution of screen recording video lengths in the dataset.}
    \label{fig:video_dist}
\end{figure}

The dataset focuses on long-horizon, open-ended workflows. The average video duration is 33 minutes and 44 seconds, with sessions ranging from approximately 16 minutes to over 1 hour and 23 minutes (Figure~\ref{fig:video_dist}). This extended duration ensures that the dataset captures the full evolution of user tasks, including periods of exploration, struggle, and error recovery.

\paragraph{Task Granularity.}
From these raw videos, we extracted varying numbers of instances for our three evaluation tasks. We collected \textbf{1.8K samples} for Behavior State Detection, \textbf{1.3K samples} for Intent Prediction, and \textbf{1K samples} for Help Prediction. Notably, the average segment length for behavior detection is shorter (14.16s) compared to Intent Prediction (25.4s) and Help Prediction (25.56s). This is because when annotating behavior states from narration-aligned segments, we instructed the model to split the clip if two or more states were identified.

\begin{figure}[h]
    \centering
    \includegraphics[width=0.9\linewidth]{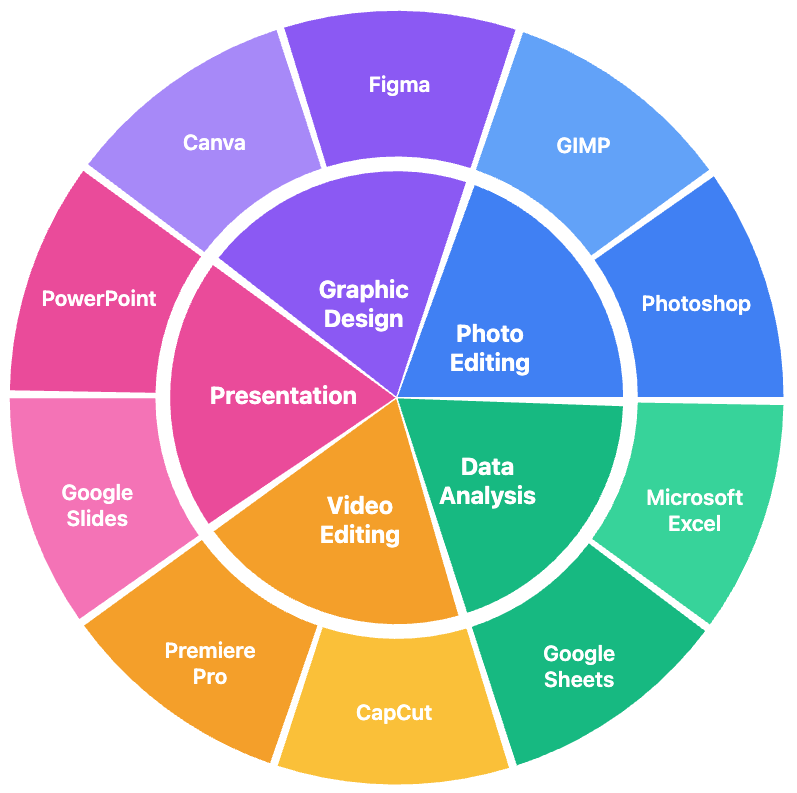}
    \caption{Software categories represented in the dataset.}
    \label{fig:software_cat}
\end{figure}

\paragraph{Diversity.}
To ensure generalizability, the dataset spans a wide variety of software domains. As illustrated in Figure~\ref{fig:software_cat}, users interacted with diverse applications ranging from creative design tools to analytical software.

\onecolumn
\subsection{Task Composition}
To ensure our benchmark captures a comprehensive range of user behaviors, we designed a set of 20 open-ended tasks across five distinct software categories: Photo Editing, Graphic Design, Presentation Design, Video Editing, and Data Analysis. Table~\ref{tab:task_descriptions} provides a detailed overview of these categories and their corresponding tasks.

\paragraph{Open-Ended Task Design.}
Unlike rigid, step-by-step tutorials that result in linear behavior, our tasks are designed to be goal-oriented and open-ended. For instance, while we provided users with necessary materials (e.g., raw video clips, images) and suggested specific software features to utilize, we did not prescribe a fixed execution path or a target reference outcome. This semi-structured ambiguity is intentional; it forces users to engage in high-level planning, trial-and-error exploration, and problem-solving. Consequently, this setup naturally elicits the complex behavior states---such as \textit{Exploration}, \textit{Debugging}, and \textit{Frustration}---that \sysname{} aims to detect.

\paragraph{Domain Diversity.}
The selected software categories cover a broad spectrum of software domains, ensuring comprehensive coverage of diverse GUI workflows. Our dataset spans \textbf{creative domains} (Photo Editing, Graphic Design) that rely on visual manipulation and aesthetic decisions, \textbf{analytical domains} (Data Analysis) focused on data processing and logic, and hybrid tasks like \textbf{Presentation Design} or \textbf{Video Editing}. This variety ensures that our models are evaluated on their ability to generalize across diverse user interfaces, toolsets, and workflow paradigms.


\begin{table*}[h]
\centering
\vspace{1em}
\small
\begin{tabular}{lp{3.5cm}p{9cm}}
\toprule
\textbf{Category} & \textbf{Software} & \textbf{Tasks} \\
\midrule

\textbf{Photo Editing} & Photoshop, GIMP & 
1. Create a composite from two images. \newline
2. Create a bakery logo with a warm, friendly identity. \newline
3. Replace a photo’s background with a custom-designed pattern. \newline
4. Design a movie poster. \\

\midrule
\textbf{Graphic Design} & Figma, Canva & 
1. Design a mobile sign-up screen for a fictional app. \newline
2. Design a custom 404 error page with a visual and animated element. \newline
3. Design compact profile cards that display personal user details. \newline
4. Design an event poster for a music festival. \\

\midrule
\textbf{Presentation Design} & PowerPoint, Google Slides & 
1. Create a product pitch deck that highlights the MacBook's key features. \newline
2. Create an interactive timeline presenting a company’s history. \newline
3. Create a 5-slide nature-themed shape-masked photo scrapbook. \newline
4. Create a quiz deck with 3 multiple-choice questions. \\

\midrule
\textbf{Video Editing} & Premiere Pro, CapCut & 
1. Edit a short interview to improve clarity and engagement. \newline
2. Design a creative intro using animated text. \newline
3. Edit a short instructional video to clearly guide a process. \newline
4. Transform a long video into a highly engaging short-form clip. \\

\midrule
\textbf{Data Analysis} & Microsoft Excel, Google Sheets & 
1. Design a Gantt chart for a mini project. \newline
2. Summarize and visualize responses from a survey. \newline
3. Visualize student performance across subjects. \newline
4. Summarize and visualize product sales by category or region. \\

\bottomrule
\end{tabular}
\caption{Overview of open-ended tasks across software categories. Each category includes two software applications and four tasks designed to elicit natural and diverse user behaviors.}
\label{tab:task_descriptions}
\end{table*}

\clearpage
\section{User Behavior Taxonomy}
\label{sec:taxonomy}

To effectively assist users, an agent must understand not just \textit{what} the user is doing (e.g., clicking a mouse), but \textit{why} they are doing it and what their current cognitive and behavior state is. We introduce a hierarchical taxonomy of 9 user behavior states, organized into four high-level phases of the software workflow: \textbf{Planning}, \textbf{Execution}, \textbf{Problem-Solving}, and \textbf{Evaluation}. Table~\ref{tab:taxonomy_descriptions} provides detailed definitions and examples for each state.

\begin{table*}[h]
\vspace{1em}
\centering
\small
\begin{tabular}{p{2.5cm} p{7.2cm} p{6cm}}
\toprule
\textbf{Behavior State} & \textbf{Description} & \textbf{Examples} \\
\midrule

\multicolumn{3}{l}{\textbf{Planning}} \\
\midrule
Task Understanding and Preparation &
The user is focused on the logistics of the task. This includes interpreting the task, gathering necessary digital assets, and configuring the software environment. Their goal is to set up the conditions needed to begin the work. &
Reading task instructions, opening required software/files/templates, arranging workspace (resizing windows, organizing directories), downloading images for photo editing. \\

Ideation and Planning &
The user is engaged in high-level conceptual work. They are brainstorming ideas, outlining the structure of the outcome, or creating a plan for how to approach the task. This often involves creating preliminary, non-final content that serves as a guide. &
Formulating high-level strategy, creating step lists, sketching rough layouts or wireframes. The output is a plan or outline, not the final polished product. \\

\midrule
\multicolumn{3}{l}{\textbf{Execution}} \\
\midrule
Exploration and Decision-Making &
The user experiments with different options or features to understand their effects and decide which one to use. This exploratory phase involves deliberate trial and comparison, often pausing forward progress to evaluate alternatives. &
Applying effects and undoing them, hovering over tools to see what they do, testing multiple font sizes to decide which fits best. \\

Performing Actions &
The user is confidently using the software to make progress on the task. These actions are purposeful and executed with little hesitation. &
Typing/deleting text, inserting and resizing images, applying formatting with clear intent, searching for functions to use. \\

\midrule
\multicolumn{3}{l}{\textbf{Problem-Solving}} \\
\midrule
Frustration &
The user encounters a blocker and shows signs of being stuck, confused, or annoyed. The system may not behave as expected, or the user cannot find a way to perform a desired action, leading to repetitive or unproductive behavior. &
Sighing, pausing for long periods, undoing repeatedly, clicking unresponsive elements, complaining about slow system behavior. \\

Debugging &
The user moves beyond frustration and begins to actively investigate the cause of a problem. They form and test hypotheses to diagnose and fix an issue. &
Testing alternative approaches, undoing recent actions step by step, forming hypotheses about causes, adjusting settings to identify errors. \\

Seeking External Help &
The user recognizes a gap in their own knowledge and turns to an external resource for assistance or procedural guidance. &
Switching to a web browser for solutions, opening tutorials/documentation, consulting AI assistants or colleagues, posting questions in forums. \\

\midrule
\multicolumn{3}{l}{\textbf{Evaluation}} \\
\midrule
Waiting and Monitoring &
The user is in a passive state, waiting for a system-controlled process to complete before continuing their work. They are unable to take meaningful action and typically observe progress indicators. &
Watching loading bars or spinners, waiting for exports or rendering to complete. \\

Assessment &
The user intentionally pauses their work to review and evaluate their output. They examine the result for quality, accuracy, or aesthetics. &
Zooming in/out to inspect fine details, replaying video snippets for review, comparing results to reference images or previous versions. \\

\bottomrule
\end{tabular}
\caption{Taxonomy of user behavior states in open-ended GUI workflows.}
\label{tab:taxonomy_descriptions}
\end{table*}

\clearpage
\subsection{Behavior State Distribution}
\begin{figure*}[h]
  \centering
  \includegraphics[width=0.8\linewidth]{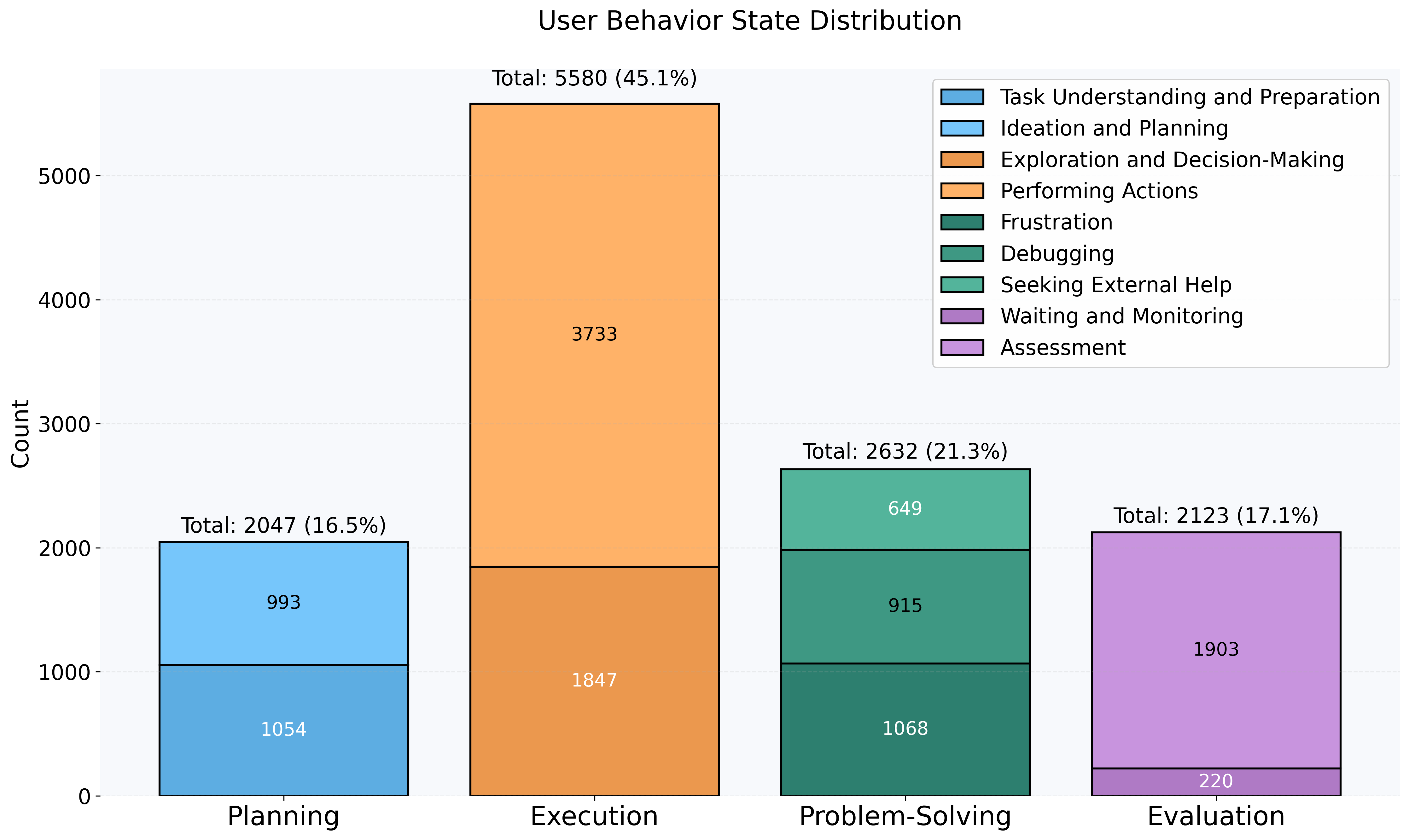}
  \caption{Distribution of user behavior states across Planning, Execution, Problem-Solving, and Evaluation phases across the videos in the dataset.}
  \label{fig:state_distribution}
\end{figure*}

Figure~\ref{fig:state_distribution} illustrates the overall distribution of user behavior states across the four high-level phases defined in our taxonomy. 
Table~\ref{tab:label_distribution} provides a granular breakdown of these states across the specific evaluation tasks.

\begin{table*}[h]
\centering
\small
\caption{Distribution of behavior state labels across the full dataset and specific evaluation tasks. Note that annotated instances used in the evaluation tasks may involve two or more states (e.g., a single segment containing both \textit{Debugging} and \textit{Seeking External Help}). Behavior State Detection uniformly sampled 200 instances from each class.}
\label{tab:label_distribution}
\begin{tabular}{l rr rr rr rr}
\toprule
& \multicolumn{2}{c}{\textbf{Dataset}} & \multicolumn{2}{c}{\textbf{Intent Prediction}} & \multicolumn{2}{c}{\textbf{Help Need Detection}} & \multicolumn{2}{c}{\textbf{Help Content Prediction}} \\
\cmidrule(lr){2-3} \cmidrule(lr){4-5} \cmidrule(lr){6-7} \cmidrule(lr){8-9}
\textbf{Behavior State} & \textbf{Count} & \textbf{(\%)} & \textbf{Count} & \textbf{(\%)} & \textbf{Count} & \textbf{(\%)} & \textbf{Count} & \textbf{(\%)} \\
\midrule
\multicolumn{9}{l}{\textit{Planning}} \\
Task Understanding and Preparation & 1054 & 8.51\% & 216 & 9.85\% & 103 & 5.65\% & 36 & 2.82\% \\
Ideation and Planning & 993 & 8.02\% & 282 & 12.86\% & 84 & 4.61\% & 41 & 3.21\% \\
\midrule
\multicolumn{9}{l}{\textit{Execution}} \\
Exploration and Decision-Making & 1847 & 14.92\% & 289 & 13.18\% & 162 & 8.89\% & 114 & 8.92\% \\
Performing Actions & 3733 & 30.15\% & 697 & 31.80\% & 474 & 26.00\% & 228 & 17.84\% \\
\midrule
\multicolumn{9}{l}{\textit{Problem-Solving}} \\
Frustration & 1068 & 8.63\% & 131 & 5.98\% & 416 & 22.82\% & 415 & 32.47\% \\
Debugging & 915 & 7.39\% & 103 & 4.70\% & 259 & 14.21\% & 252 & 19.72\% \\
Seeking External Help & 649 & 5.24\% & 114 & 5.20\% & 85 & 4.66\% & 83 & 6.49\% \\
\midrule
\multicolumn{9}{l}{\textit{Evaluation}} \\
Waiting and Monitoring & 220 & 1.78\% & 33 & 1.51\% & 17 & 0.93\% & 9 & 0.70\% \\
Assessment & 1903 & 15.37\% & 327 & 14.92\% & 223 & 12.23\% & 100 & 7.82\% \\
\bottomrule
\end{tabular}
\end{table*}
\clearpage
\subsection{Error Analysis: Behavior State Detection}

\begin{figure*}[h]
  \centering
  \includegraphics[width=0.9\linewidth]{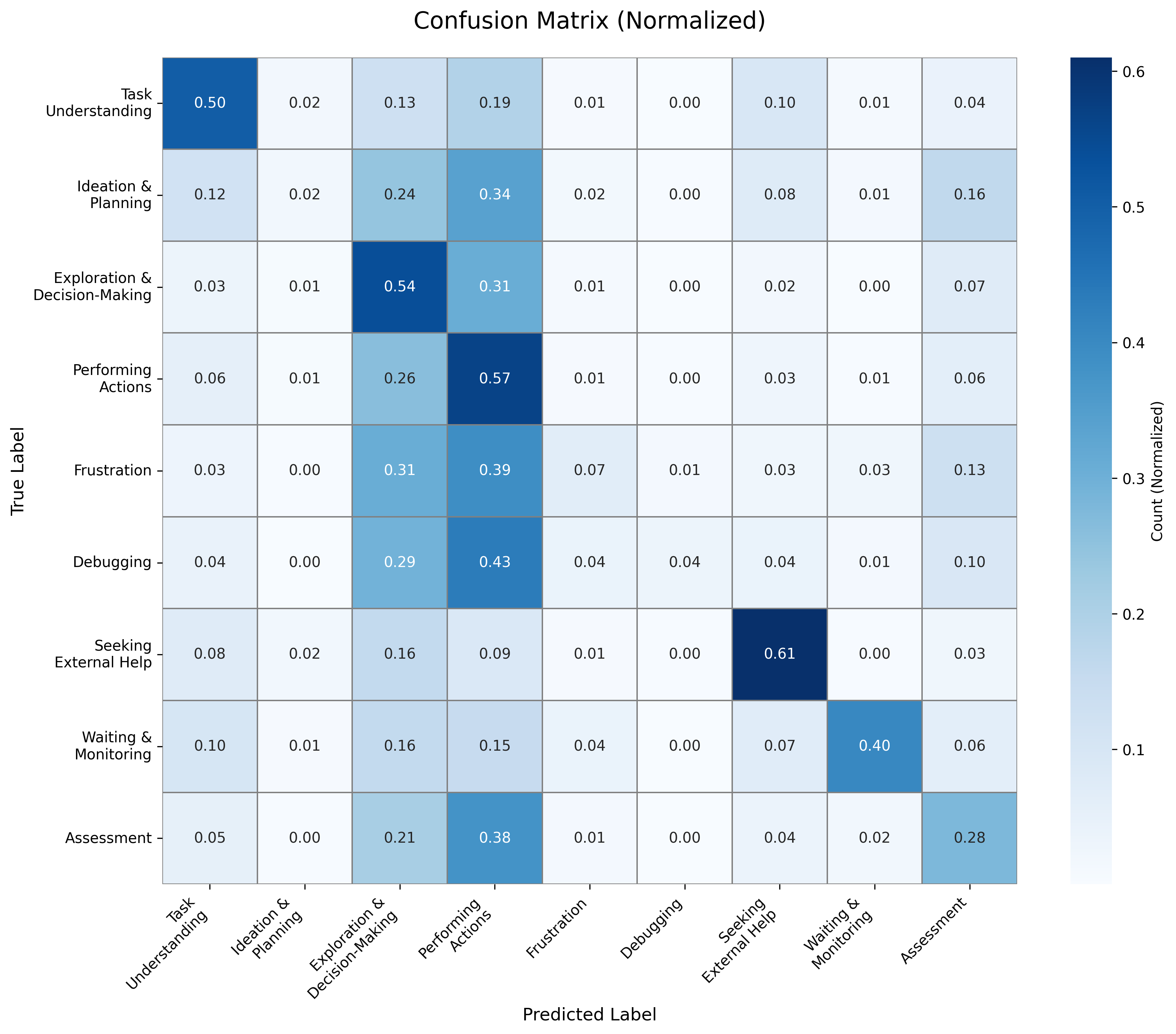}
  \caption{Normalized confusion matrix for user behavior state classification. The most common errors occur when \textit{Frustration} or \textit{Debugging} is misclassified as 
  \textit{Performing Actions} or \textit{Exploration and Decision-Making.}}
  \label{fig:taxonomy_result_confusion}
\end{figure*}

Figure~\ref{fig:taxonomy_result_confusion} presents the normalized confusion matrix for Behavior State Detection (\cref{sec:results_behavior_state}). The results reveal a critical limitation in current MLLMs: a systemic bias toward interpreting interactions as productive execution while failing to recognize signs of struggle or hesitation. 
While models achieve reasonable accuracy for visually distinct states like \textit{Seeking External Help} (0.61) and \textit{Performing Actions} (0.57), they show near-zero capability in detecting \textit{Frustration} (0.07) and \textit{Debugging} (0.04). Instead, these negative states are overwhelmingly misclassified as \textit{Performing Actions} (39\% and 43\%, respectively) or \textit{Exploration and Decision-Making} (31\% and 29\%). This suggests that models perceive the visual activity of a struggling user---such as repeated clicking or rapid mouse movements---as deliberate progress, lacking the temporal understanding to distinguish between trial-and-error and confident execution.


\clearpage
\section{Screen Recording Video Examples}
We present qualitative examples to illustrate the richness of the multimodal data in \sysname{}.

\begin{table*}[h]
\centering
\small
\begin{tabular}{m{0.50\textwidth} m{0.50\textwidth}}
\toprule
\multicolumn{2}{c}{\textbf{Software}: Canva, \textbf{Task}: Design a mobile sign-up screen for a fictional app} \\
\midrule

\begin{minipage}{0.50\textwidth}
 \vspace{8pt}
    \centering
    \includegraphics[width=0.9\linewidth]{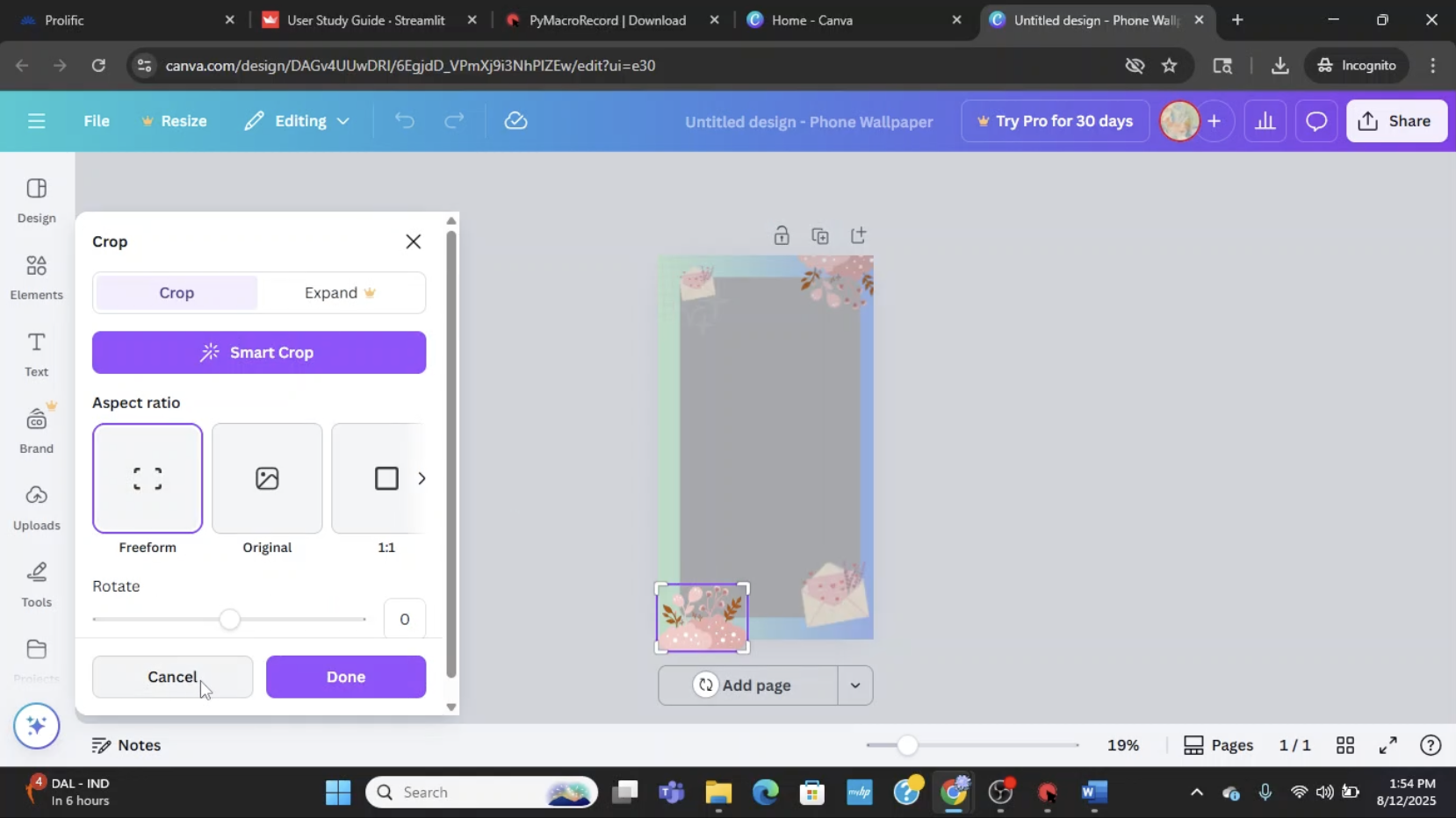}\\[3pt]
    
    \small \textit{(4:11) ``Nope, that's not what I wanted to do. Try again. Alright, let's do a text box.''}
\end{minipage}
&
\begin{minipage}{0.50\textwidth}
 \vspace{8pt}
    \centering
    \includegraphics[width=0.9\linewidth]{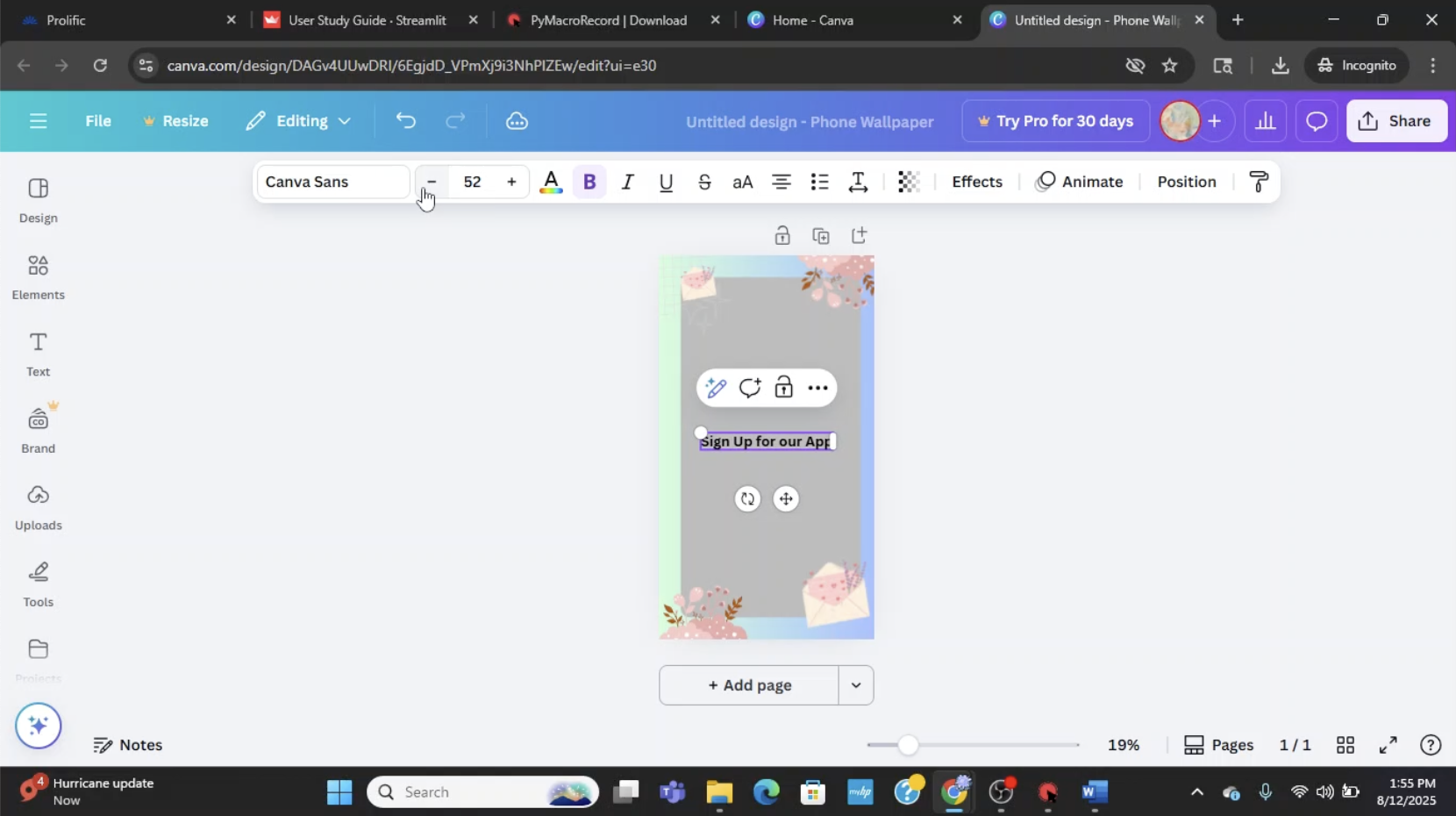}\\[3pt]
    
    \small \textit{(4:40) ``And we need to make this much smaller so it fits there at the top.''}
\end{minipage}
\\

\begin{minipage}{0.50\textwidth}
 \vspace{8pt}
    \centering
    \includegraphics[width=0.9\linewidth]{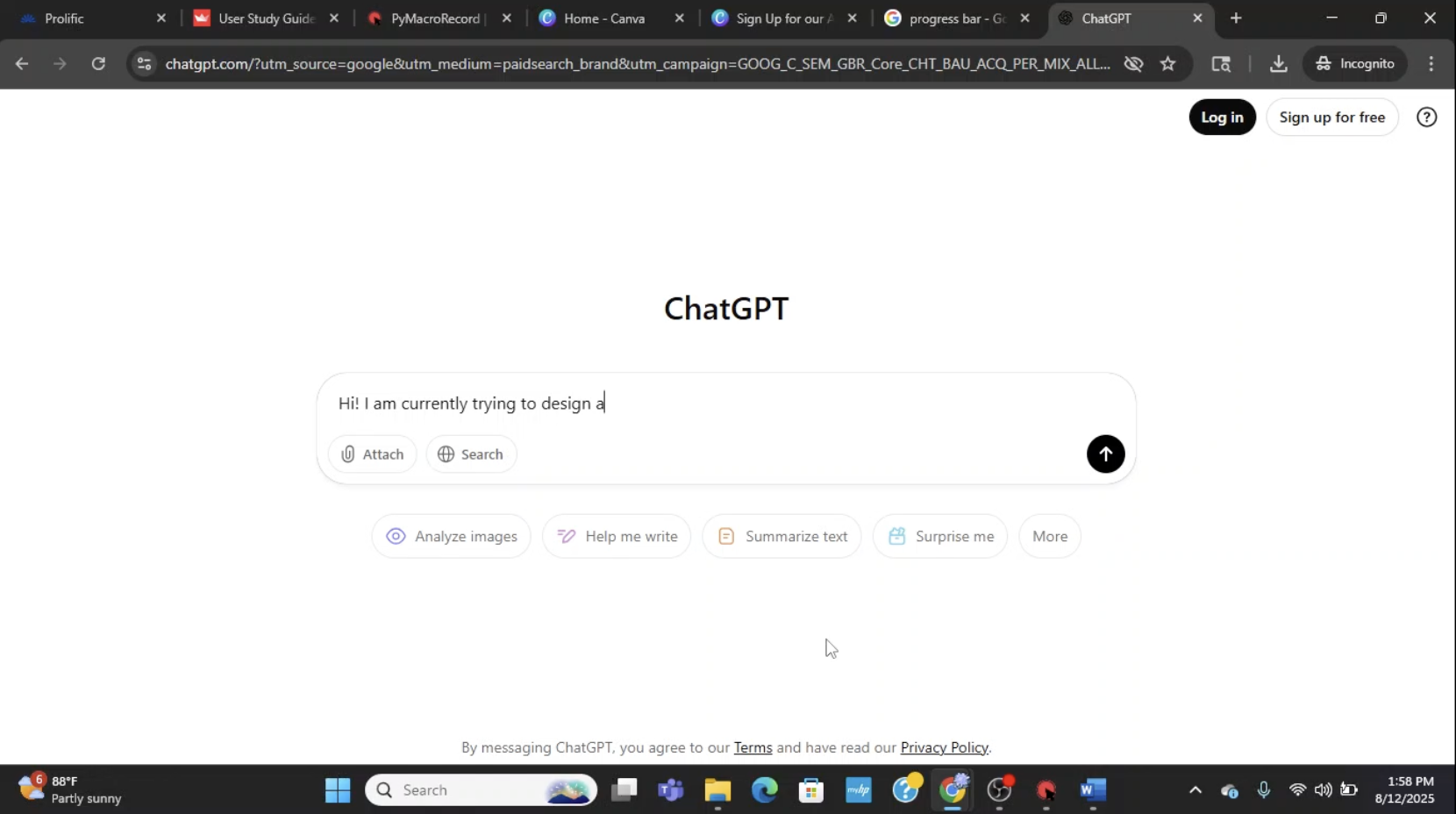}\\[3pt]
    
    \small \textit{(6:50) ``Progress bar for... what? I don't know. Just do progress bar. But you know what? Let's try ChatGPT because maybe they can help us.''}
\end{minipage}
&
\begin{minipage}{0.50\textwidth}
 \vspace{8pt}
    \centering
    \includegraphics[width=0.9\linewidth]{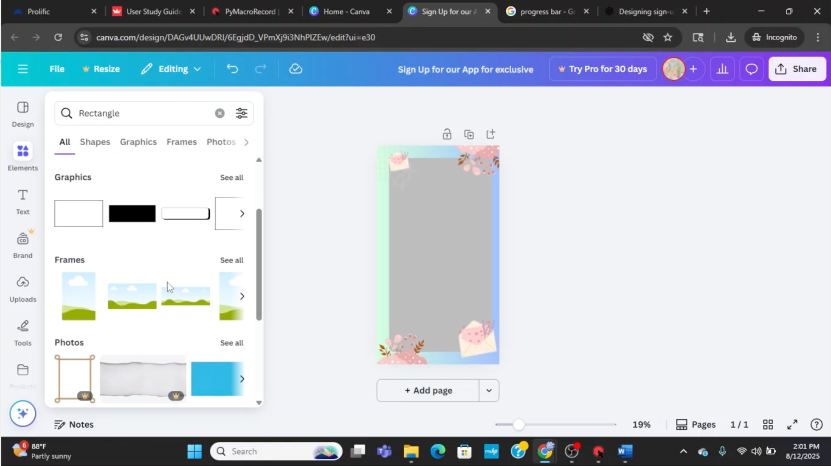}\\[3pt]
    
    \small \textit{(10:07) ``Oh, that's cool, okay. So I can create the progress bar using the free elements with the shapes. Alright, so let's try and do that. Alright. Elements.''}
\end{minipage}
\\

\begin{minipage}{0.50\textwidth}
 \vspace{8pt}
    \centering
    \includegraphics[width=0.9\linewidth]{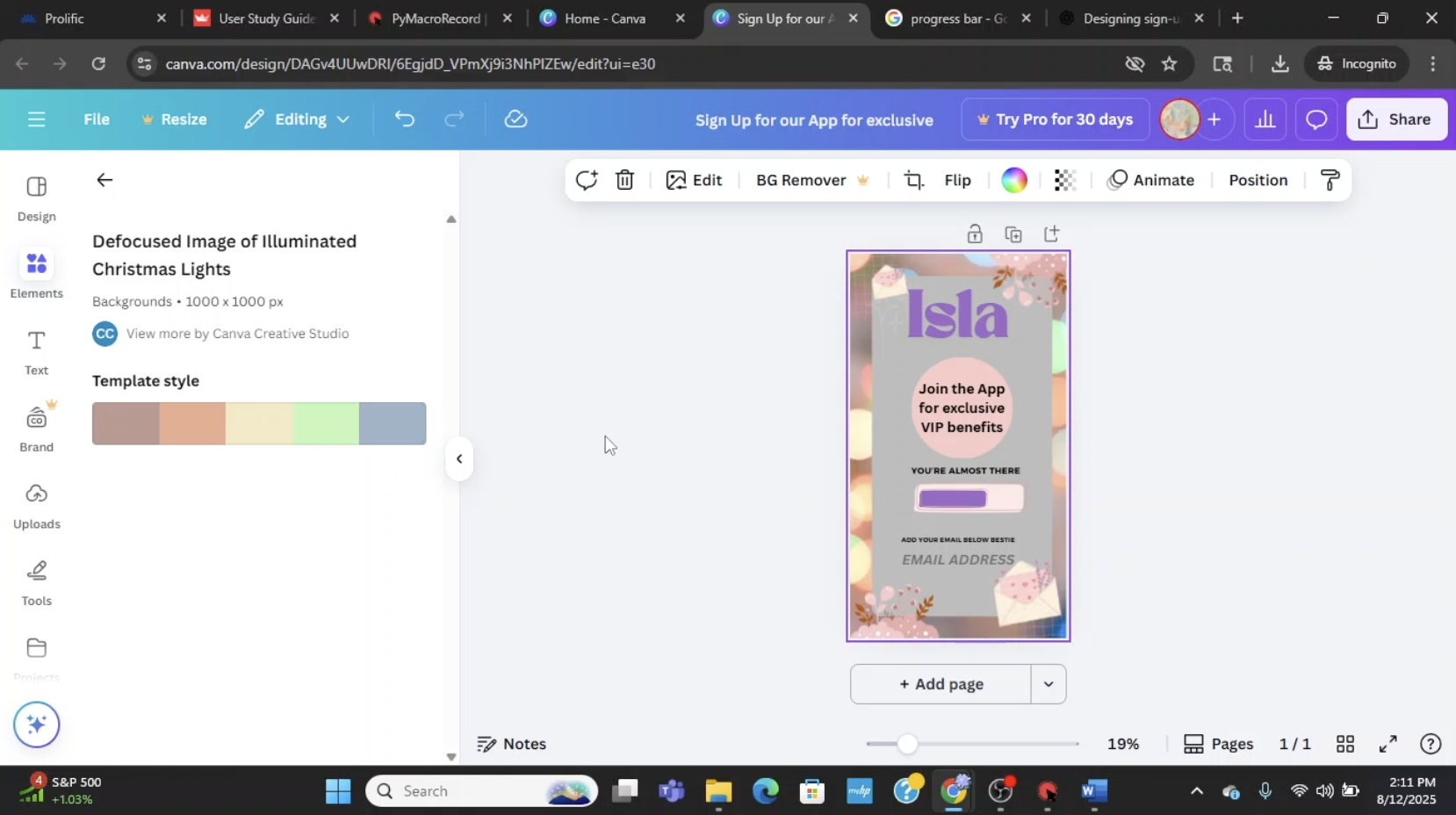}\\[3pt]
    
    \small \textit{(21:01) ``Ooh, what is this? Ooh, I like that so much better. All right, we're gonna keep this. Yes, we're gonna keep this.''}
\end{minipage}
&
\begin{minipage}{0.50\textwidth}
 \vspace{8pt}
    \centering
    \includegraphics[width=0.9\linewidth]{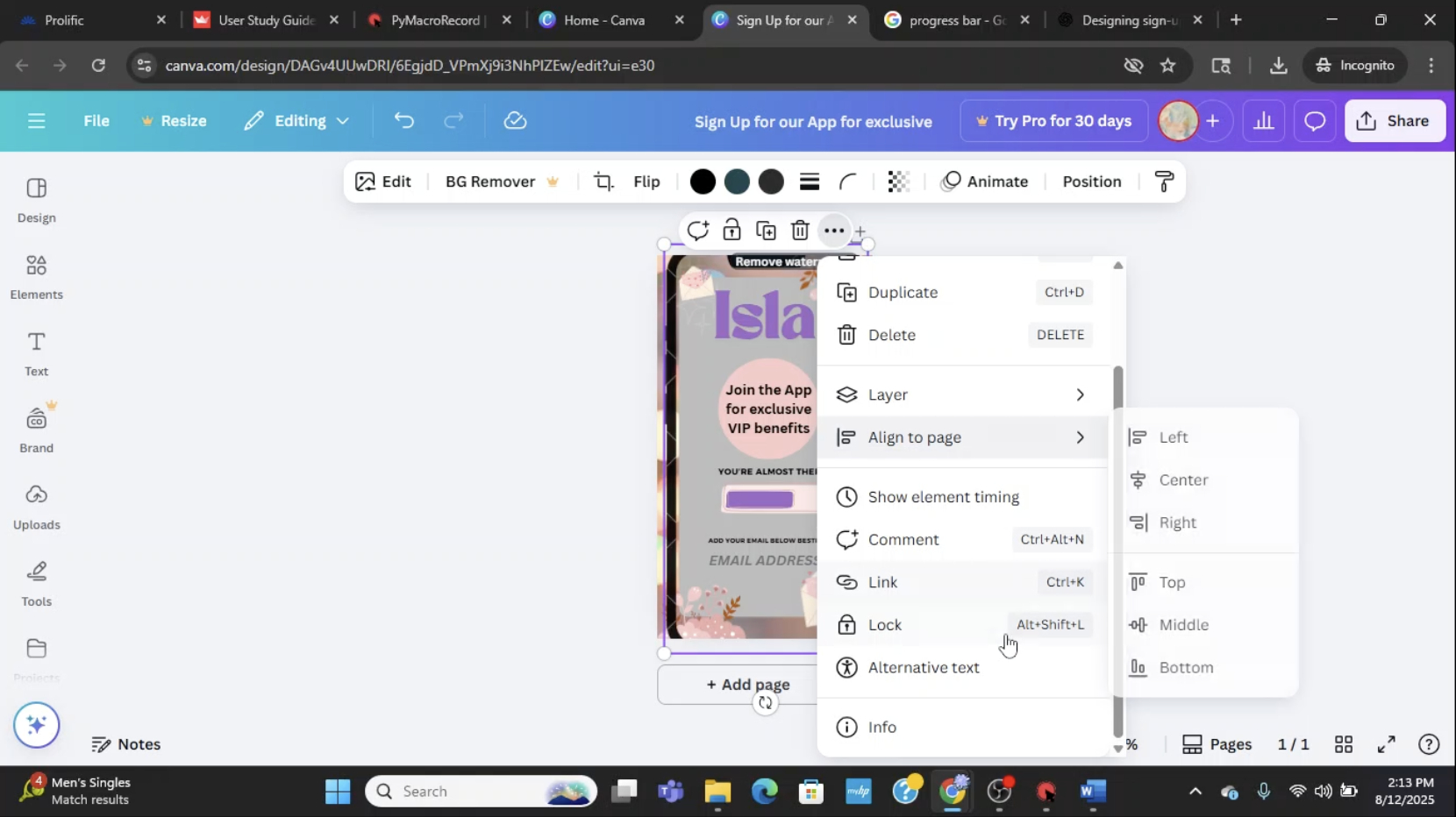}\\[3pt]
    
    \small \textit{(22:05) ``Not necessarily sure how I can... isn't there a way to like make it a certain size? How do I do that?''}
\end{minipage}
\\

\bottomrule
\end{tabular}
\caption{Example video illustrating the user’s on-screen actions accompanied by think-aloud narration.}
\end{table*}

\begin{table*}[t]
\centering
\small
\begin{tabular}{m{0.50\textwidth} m{0.50\textwidth}}
\toprule
\multicolumn{2}{c}{\textbf{Software}: CapCut, \textbf{Task}: Design a creative intro using animated text.} \\
\midrule

\begin{minipage}{0.50\textwidth}
 \vspace{8pt}
    \centering
    \includegraphics[width=0.9\linewidth]{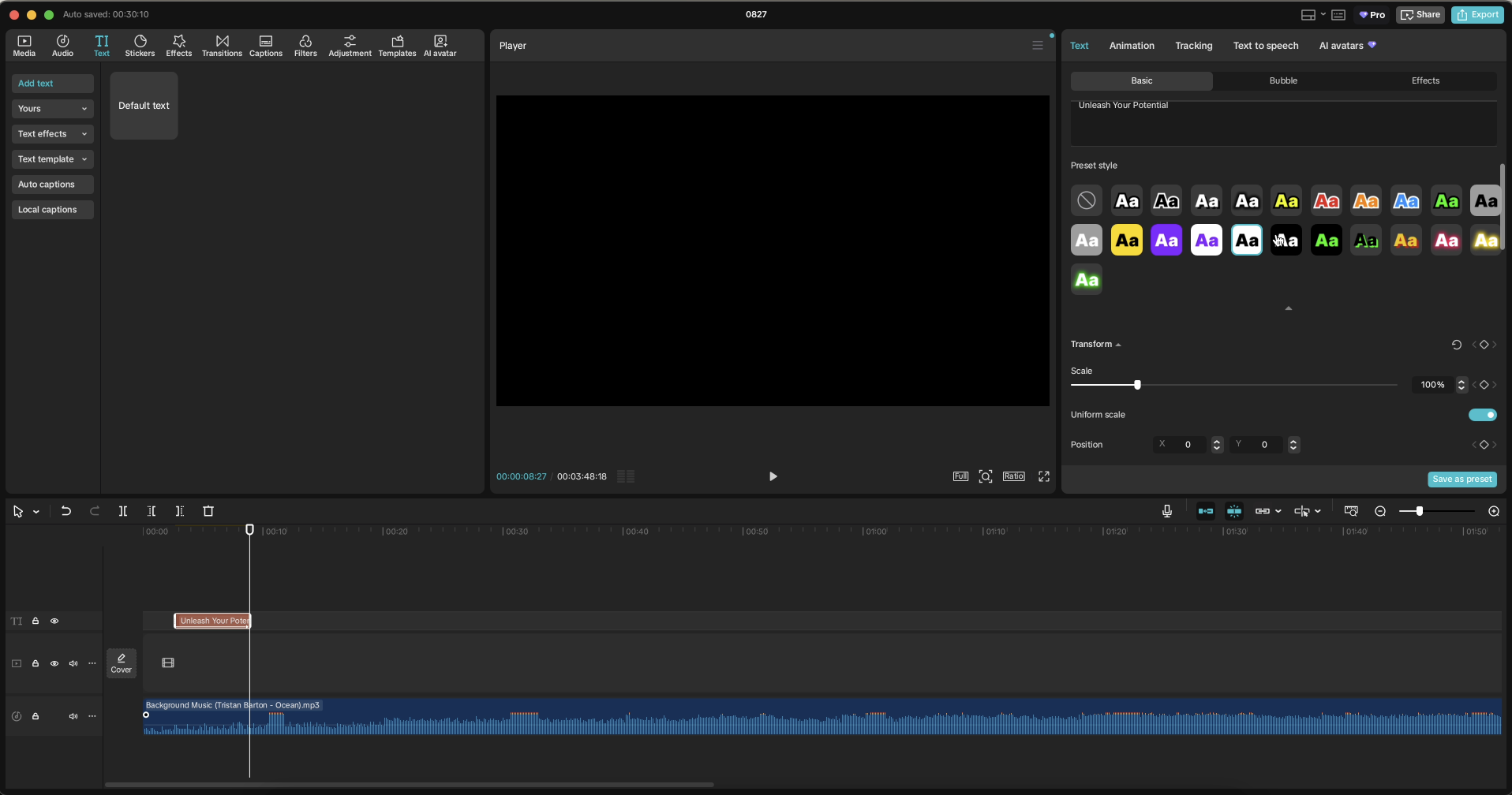}\\[3pt]
    
    \small \textit{(4:56)``but I want to make it a bit more dynamic so all right.''}
\end{minipage}
&
\begin{minipage}{0.50\textwidth}
 \vspace{8pt}
    \centering
    \includegraphics[width=0.9\linewidth]{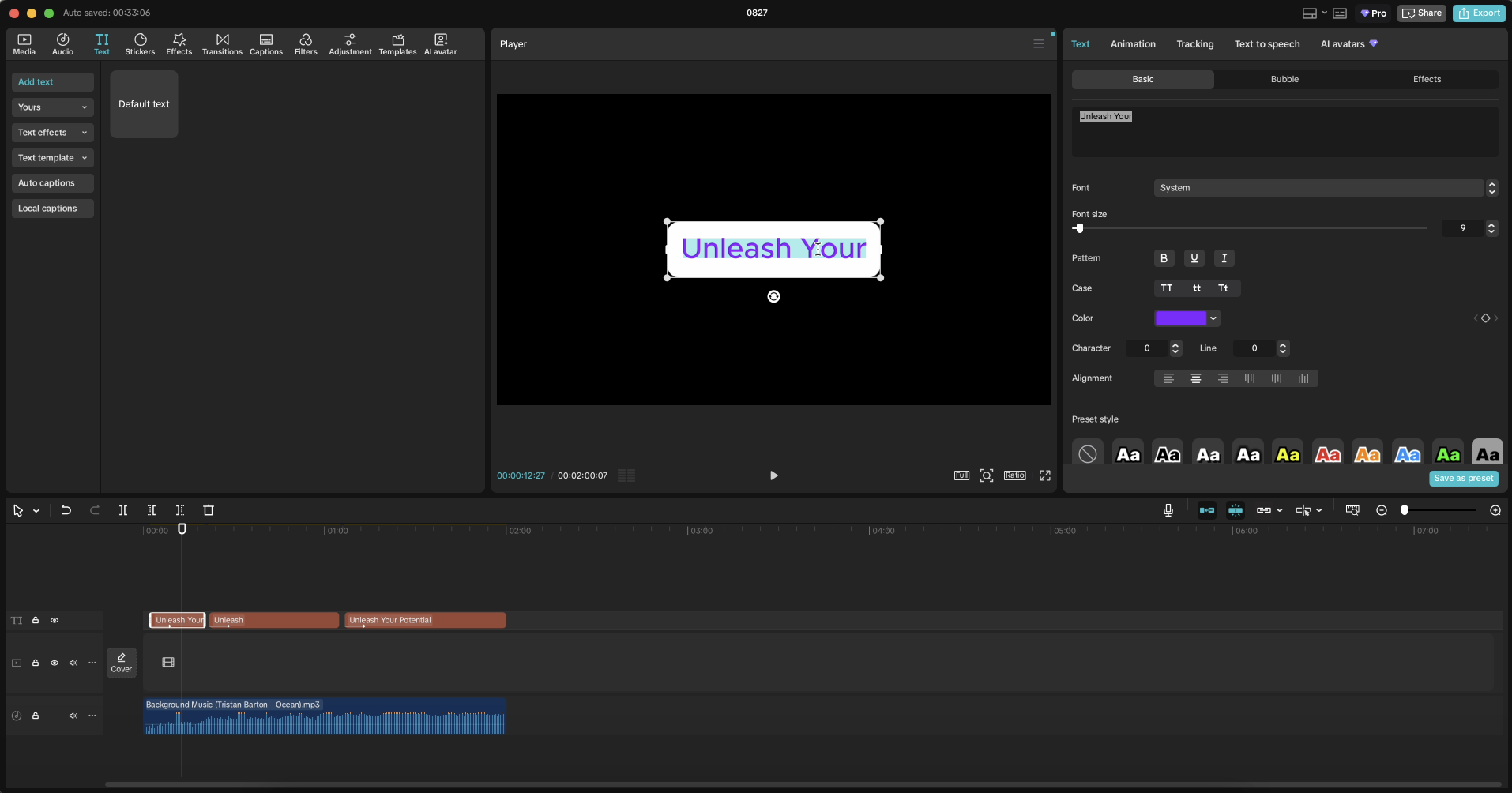}\\[3pt]
    
    \small \textit{(7:55)``oh I think here it should be only one word appearing at a time.''}
\end{minipage}
\\

\begin{minipage}{0.50\textwidth}
 \vspace{8pt}
    \centering
    \includegraphics[width=0.9\linewidth]{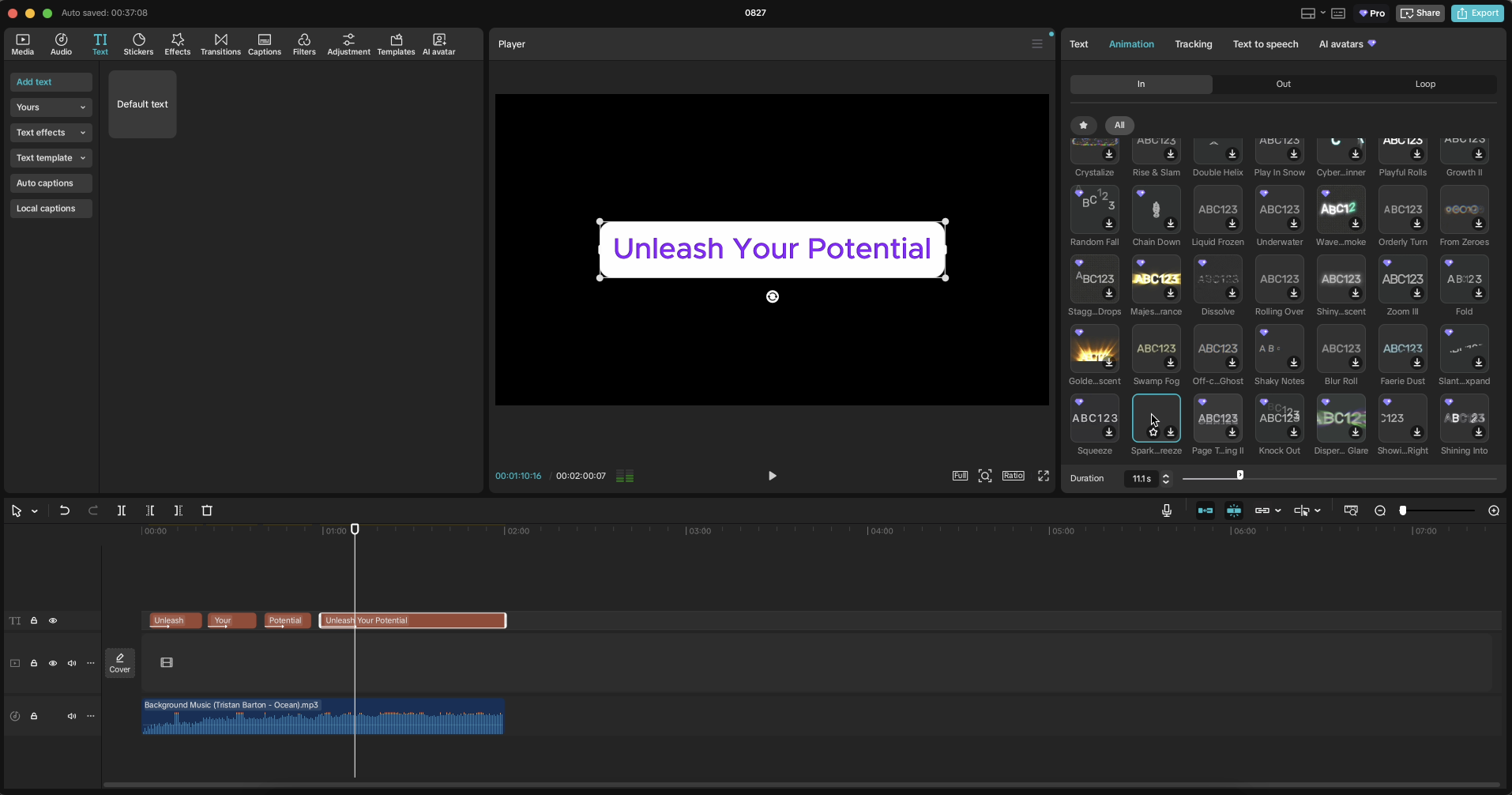}\\[3pt]
    
    \small \textit{(11:39) ``there are so many effects that I get a little bit too overwhelmed to see so many.''}
\end{minipage}
&
\begin{minipage}{0.50\textwidth}
 \vspace{8pt}
    \centering
    \includegraphics[width=0.9\linewidth]{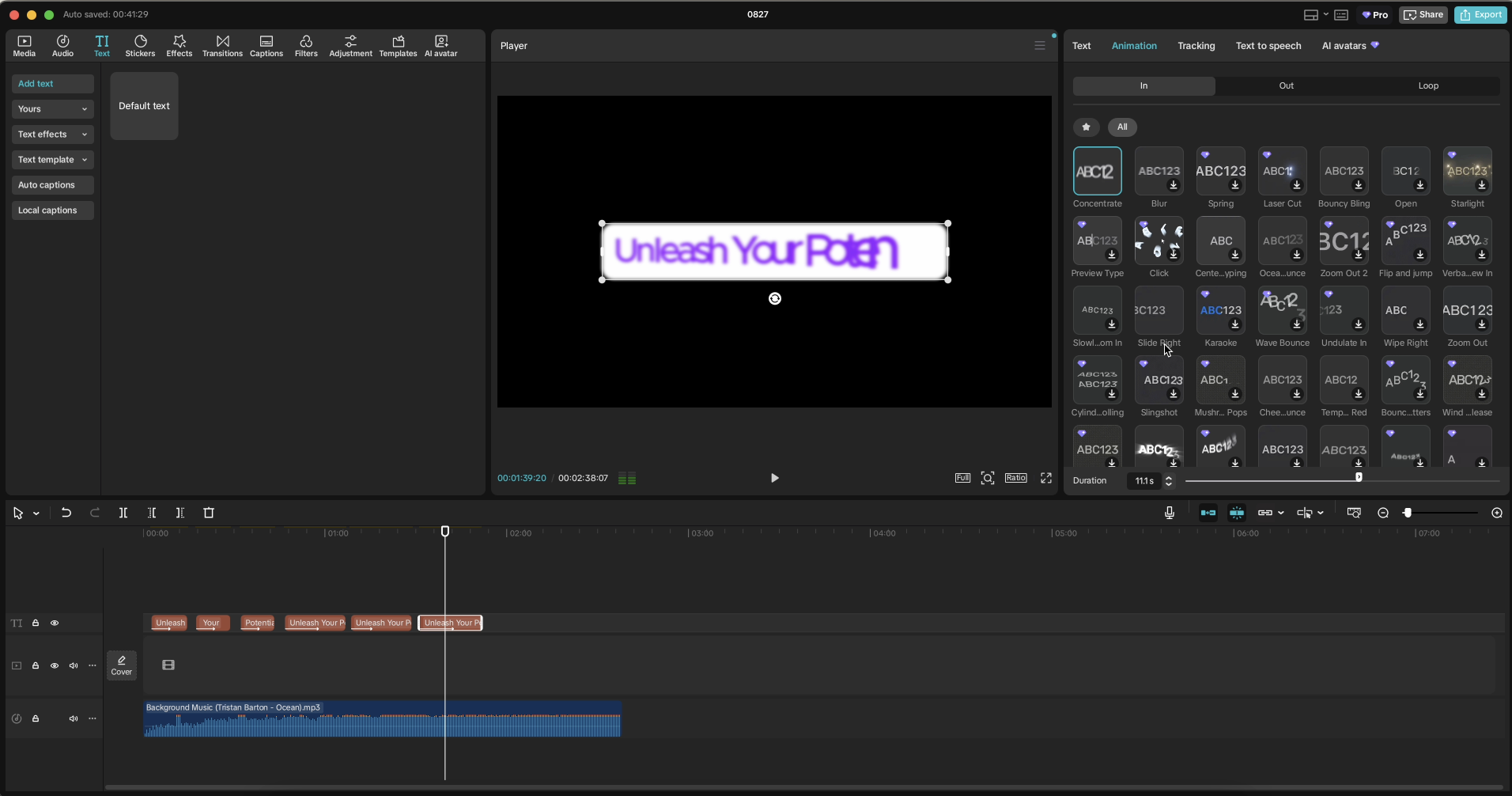}\\[3pt]
    
    \small \textit{(17:55) ``I would like to add additional different animation for this one so I like to keep this.''}
\end{minipage}
\\

\begin{minipage}{0.50\textwidth}
 \vspace{8pt}
    \centering
    \includegraphics[width=0.9\linewidth]{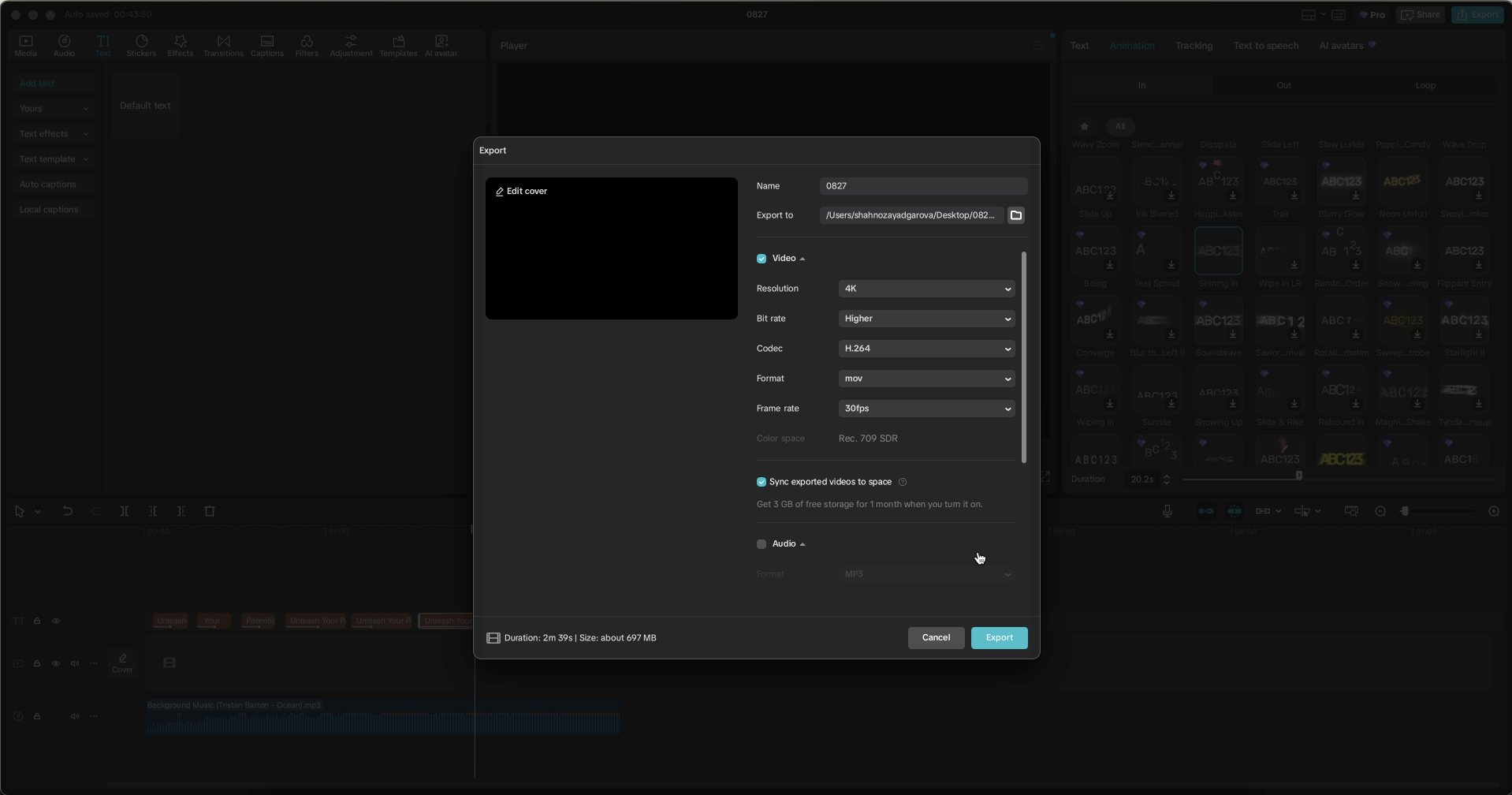}\\[3pt]
    
    \small \textit{(18:42) ``I am satisfied with the results so I will export this.''}
\end{minipage}
&
\begin{minipage}{0.50\textwidth}
 \vspace{8pt}
    \centering
    \includegraphics[width=0.9\linewidth]{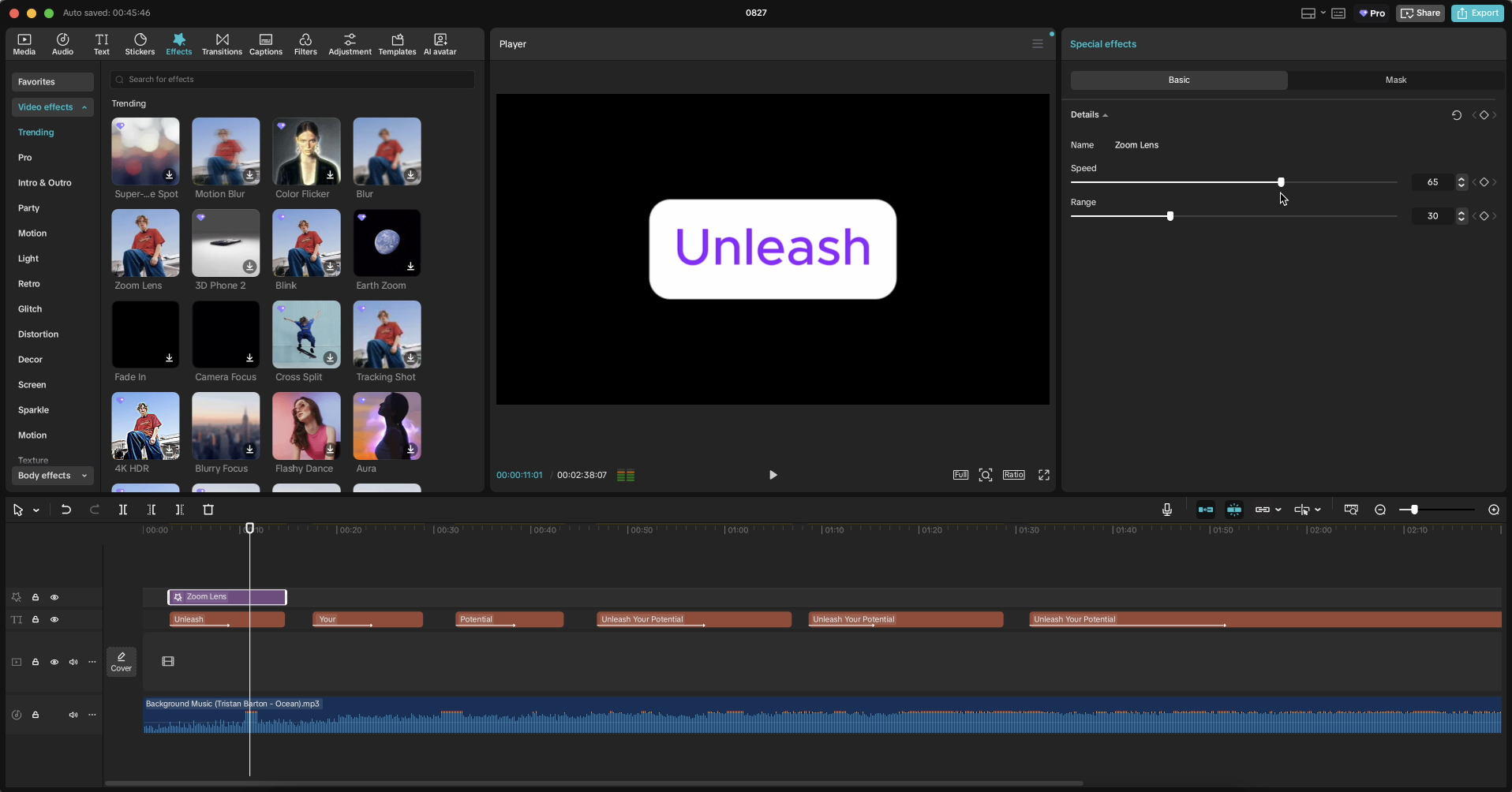}\\[3pt]
    
    \small \textit{(20:14) ``Maybe I would like to make it a bit slower.''}
\end{minipage}
\\

\bottomrule
\end{tabular}
\caption{Example video illustrating the user’s on-screen actions accompanied by think-aloud narration.}
\end{table*}

\clearpage
\section{Benchmark Task Examples}
\subsection{Behavior State Detection}
\begin{center}


\small
\begin{tabularx}{\textwidth}{m{0.50\textwidth} >{\raggedright\arraybackslash}X}
\toprule
\textbf{Screenshot} & \textbf{User Behavior State} \\
\midrule

\begin{minipage}{0.50\textwidth}
 \vspace{8pt}
    \centering
    \includegraphics[width=0.8\linewidth]{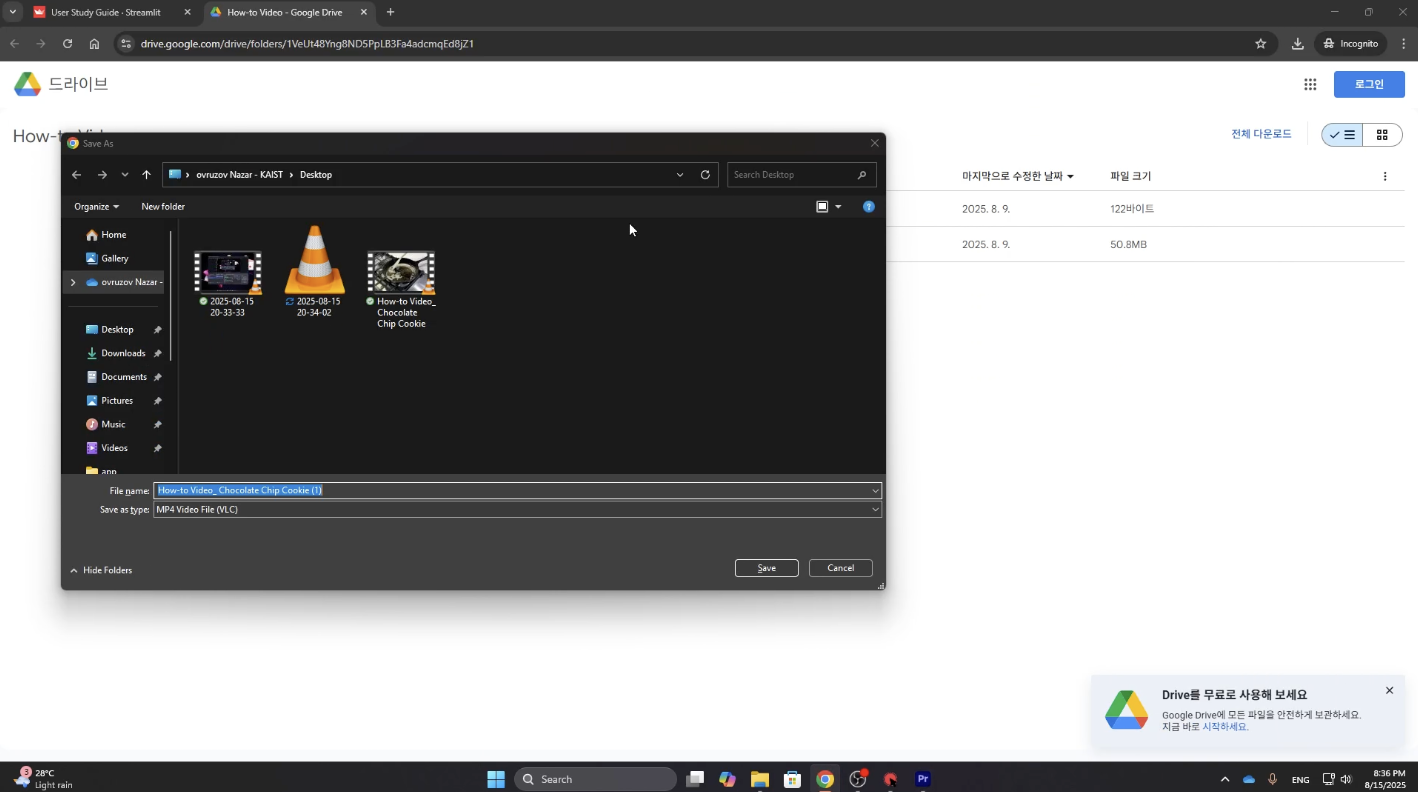}\\[3pt]
    
    \small \textit{``Okay, I downloaded it already. Delete my test, so I don't get confused. I have the video.''}
\end{minipage}
&
\textbf{Software}: Premiere Pro

\textbf{Task}: Edit a short instructional video to clearly guide a process.

\textbf{Behavior State}:
\textbf{Task Understanding and Preparation}

The user is preparing their digital workspace before starting the editing task. They locate the necessary video file on their desktop and delete a superfluous 'test' file to prevent confusion.
\\[8pt]



\begin{minipage}{0.50\textwidth}
 \vspace{8pt}
    \centering
    \includegraphics[width=0.8\linewidth]{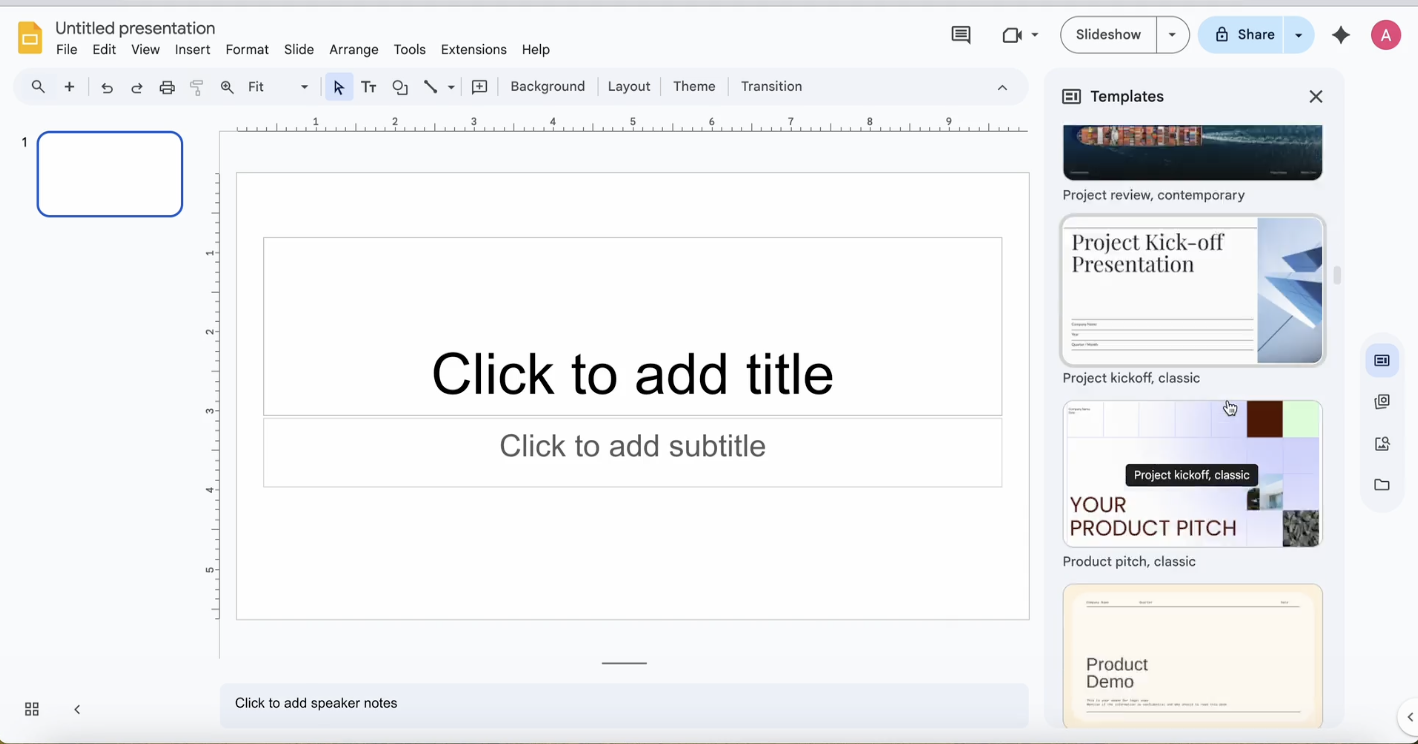}\\[3pt]
    \small \textit{``I would like to just use this design or the white some minimalistic like iOS design. Oh, this one. This one looks good. Okay, let's just...''}
\end{minipage}
&
\textbf{Software}: Google Slides

\textbf{Task}: Create a product pitch deck highlighting a product’s key features.

\textbf{Behavior State}:
\textbf{Exploration and Decision-Making}

The user is actively browsing and comparing different templates, as shown by the scrolling and hovering behavior. The narration ('This one looks good') confirms they are evaluating options to make a final decision.
\\[8pt]

\begin{minipage}{0.50\textwidth}
 \vspace{8pt}
    \centering
    \includegraphics[width=0.8\linewidth]{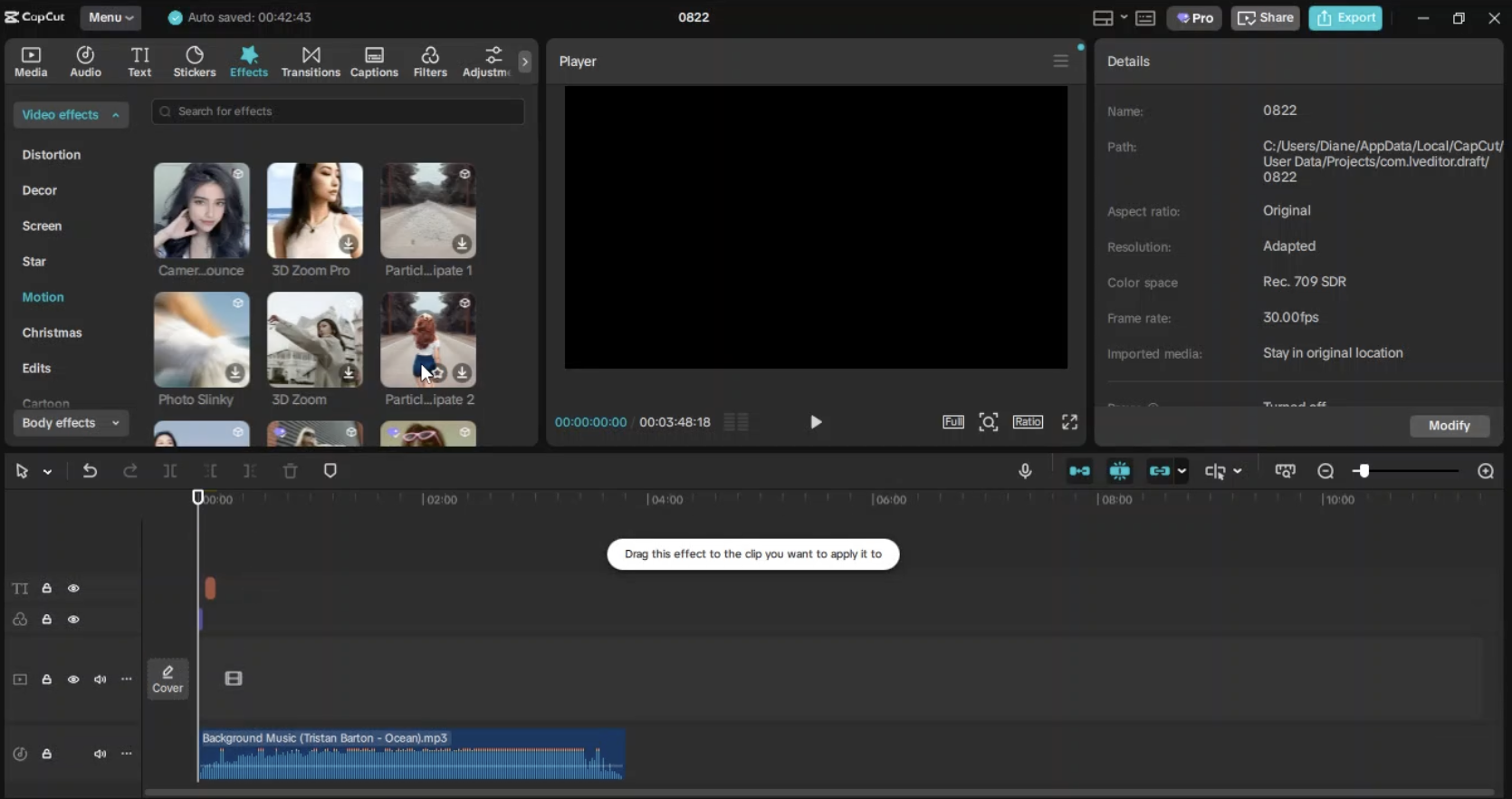}\\[3pt]
    \small  \textit{``Okay, that's strange. That's very strange, honestly.''}
\end{minipage}
&
\textbf{Software}: CapCut

\textbf{Task}: Design a creative intro using animated text.

\textbf{Behavior State}:
\textbf{Frustration}

The user verbally expresses confusion ('that's strange') after the software behaved in an unexpected way. They are momentarily paused, indicating a blocker in their workflow before they decide on a new course of action.
\\[8pt]

\begin{minipage}{0.50\textwidth}
 \vspace{8pt}
    \centering
    \includegraphics[width=0.8\linewidth]{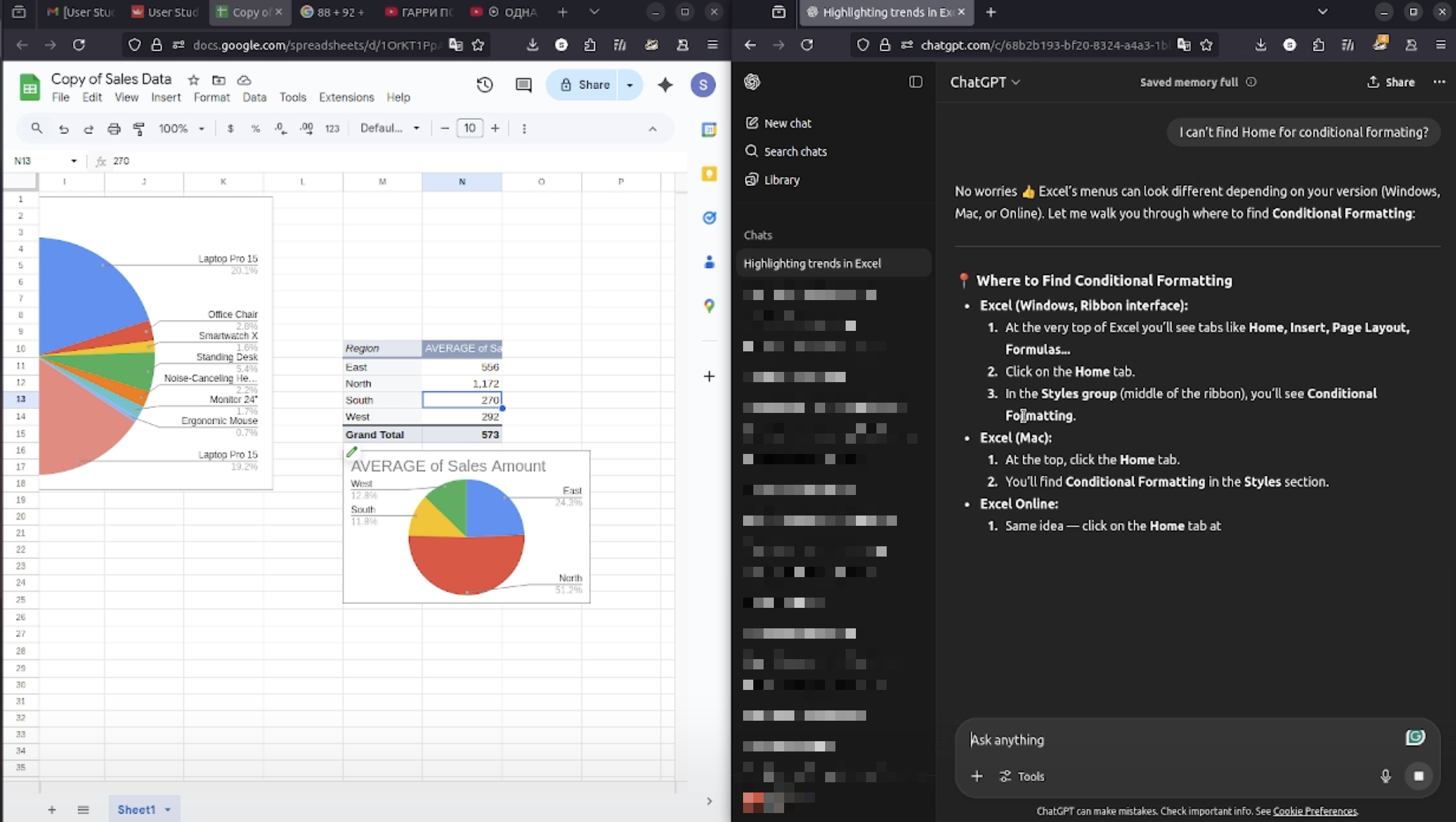}\\[3pt]
    \small \textit{(no narration)}
\end{minipage}
&
\textbf{Software}: Google Sheets

\textbf{Task}: Summarize and visualize product sales by category or region.

\textbf{Behavior State}:
\textbf{Seeking External Help}

The user is unable to find a feature and turns to ChatGPT for assistance. They type a question clarifying their problem, wait for the response, and then read the provided instructions.
\\

\bottomrule
\end{tabularx}
\captionof{table}{Example instances for the (1) User Behavior State Detection task, showing screenshots, think-aloud narration, and the corresponding behavior state.}
\end{center}

\clearpage

\subsection{Intent Prediction}
\begin{center}
\small
\begin{tabularx}{\textwidth}{m{0.50\textwidth} >{\raggedright\arraybackslash}X}
\toprule
\textbf{Screenshot} & \textbf{Intent} \\
\midrule

\begin{minipage}{0.50\textwidth}
 \vspace{8pt}
    \centering
    \includegraphics[width=0.85\linewidth]{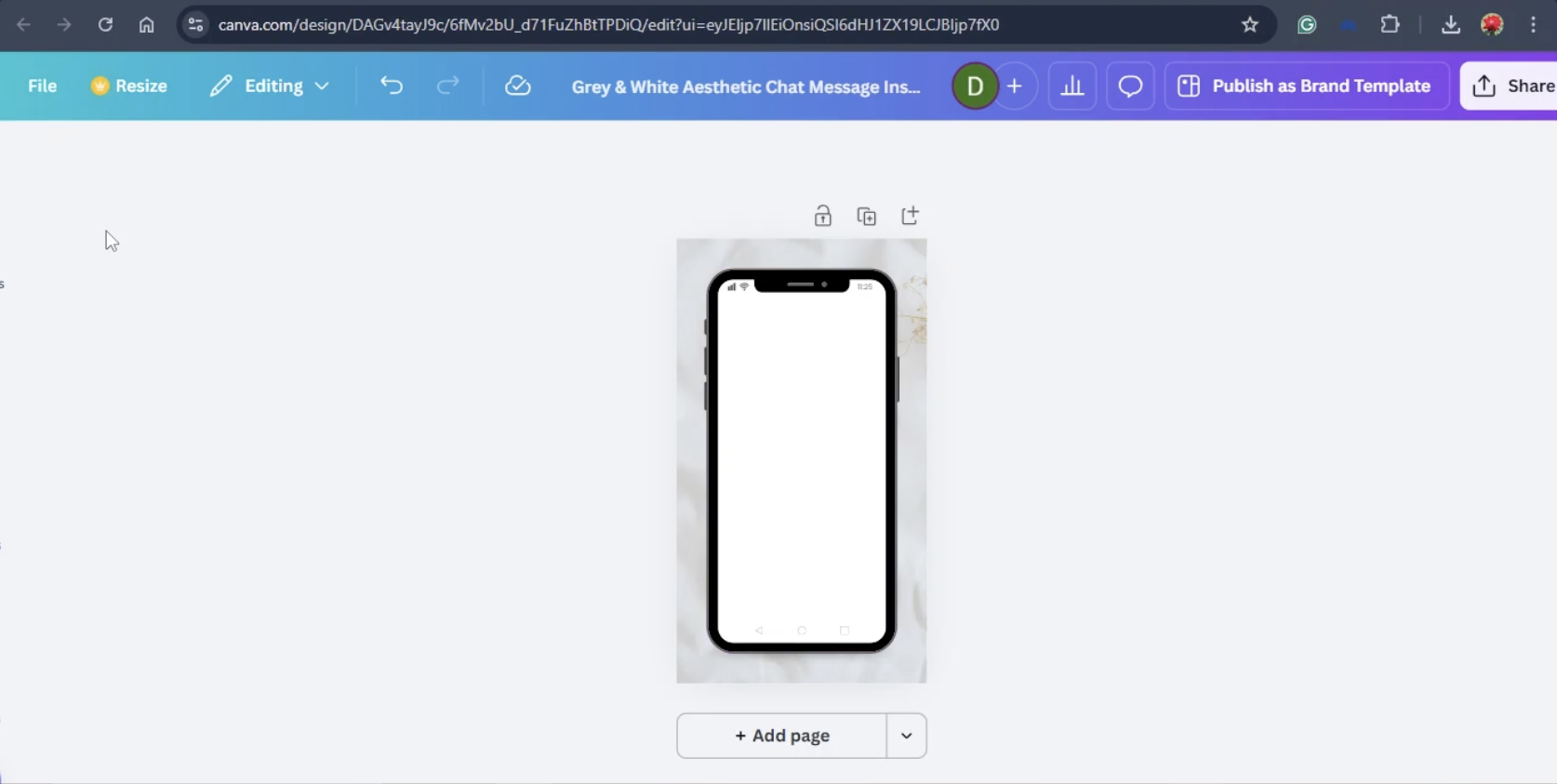}\\[3pt]
    
    \small \textit{``So now that I have the frame as a design base, I need to include the input field for name, email.''}
\end{minipage}
&
\textbf{Software}: Canva

\textbf{Task}: Design a mobile sign-up screen for a fictional app.

\textbf{Intent}:

A: Rename the design file to reflect the new project

\textbf{B: Add the required input fields to the design}

C: Search for a suitable illustration to use as a header

D: Resize the canvas to a custom dimension
\\

\begin{minipage}{0.50\textwidth}
 \vspace{8pt}
    \centering
    \includegraphics[width=0.85\linewidth]{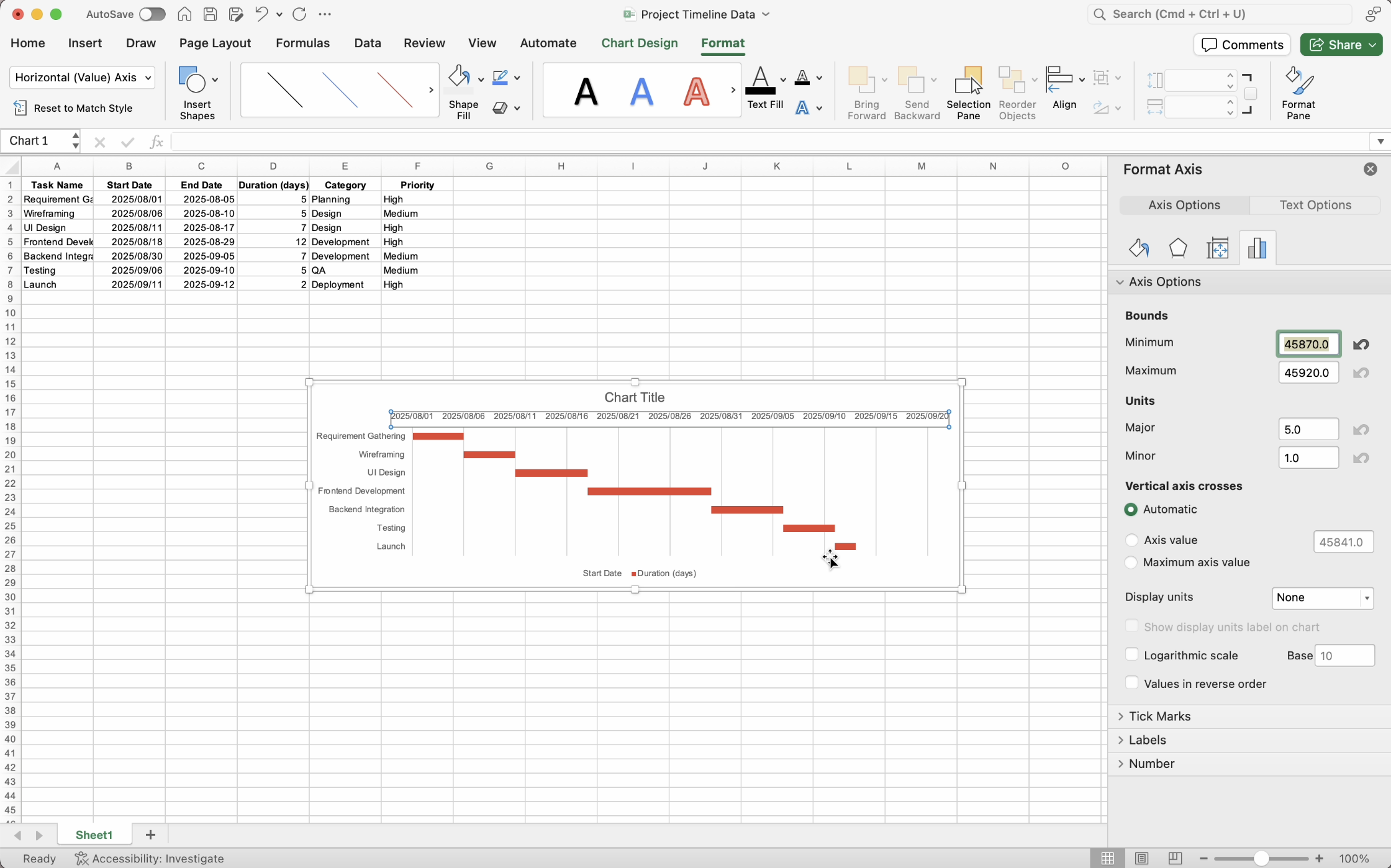}\\[3pt]
    \small \textit{``okay looks perfect, I need to adjust the end date as well''}
\end{minipage}
&
\textbf{Software}: Excel

\textbf{Task}: Design a Gantt chart for a mini project.

\textbf{Intent}:

\textbf{A: Adjust the end date of the chart's horizontal axis}

B: Adjust the date interval of the chart's horizontal axis

C: Reverse the order of the chart's vertical axis

D: Adjust the start date of the chart's horizontal axis
\\

\begin{minipage}{0.50\textwidth}
 \vspace{8pt}
    \centering
    \includegraphics[width=0.85\linewidth]{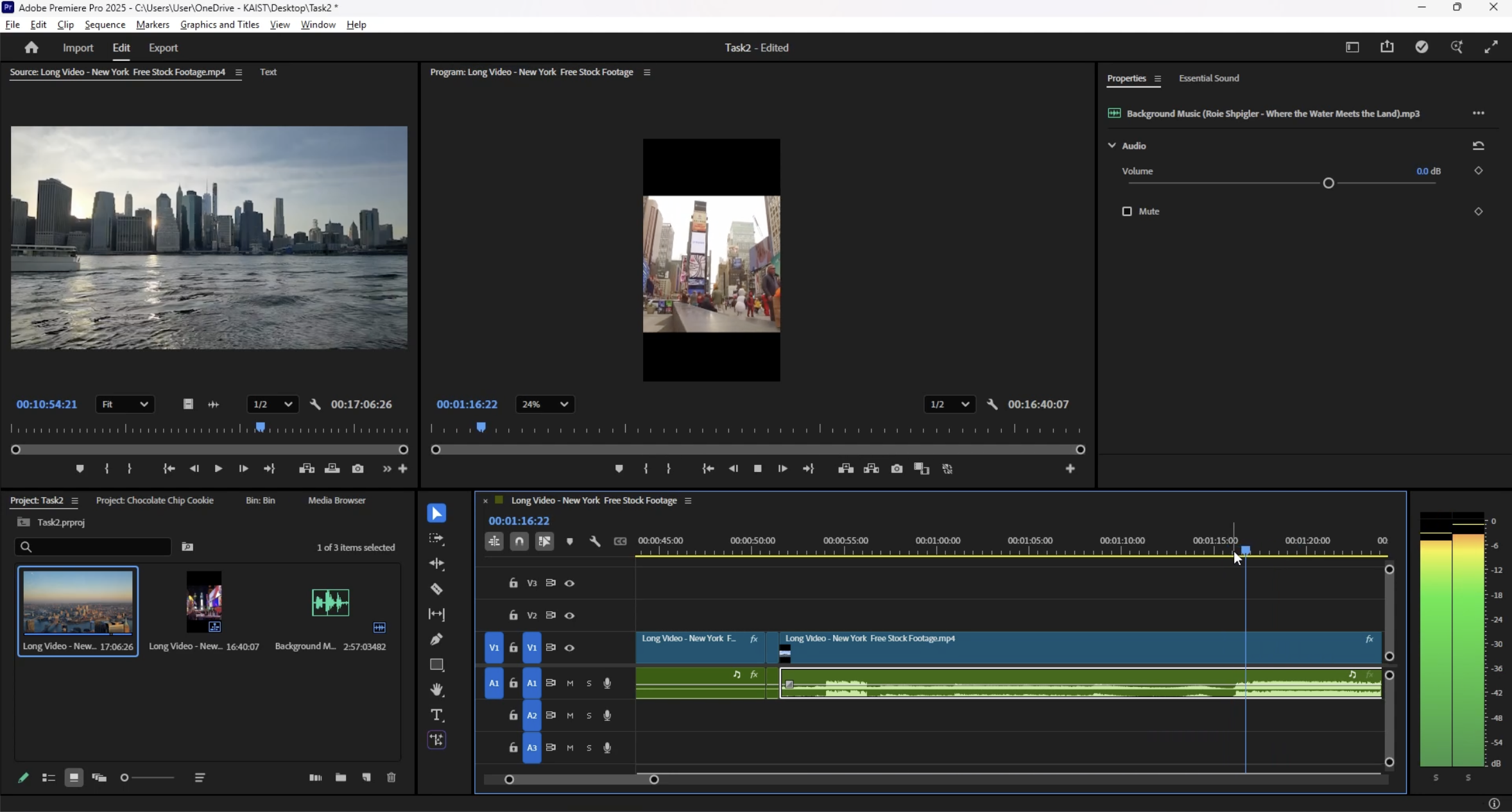}\\[3pt]
    \small  \textit{``When this slot comes, we should put some kind of image here.''}
\end{minipage}
&
\textbf{Software}: Premiere Pro

\textbf{Task}: Transform a long video into a short-form clip.

\textbf{Intent}:

A: Create a new text layer above the existing video track

\textbf{B: Add an image to a specific empty slot in the timeline}

C: Apply a transition effect to the end of a video clip

D: Add a video clip to the end of the current sequence
\\

\begin{minipage}{0.50\textwidth}
 \vspace{8pt}
    \centering
    \includegraphics[width=0.85\linewidth]{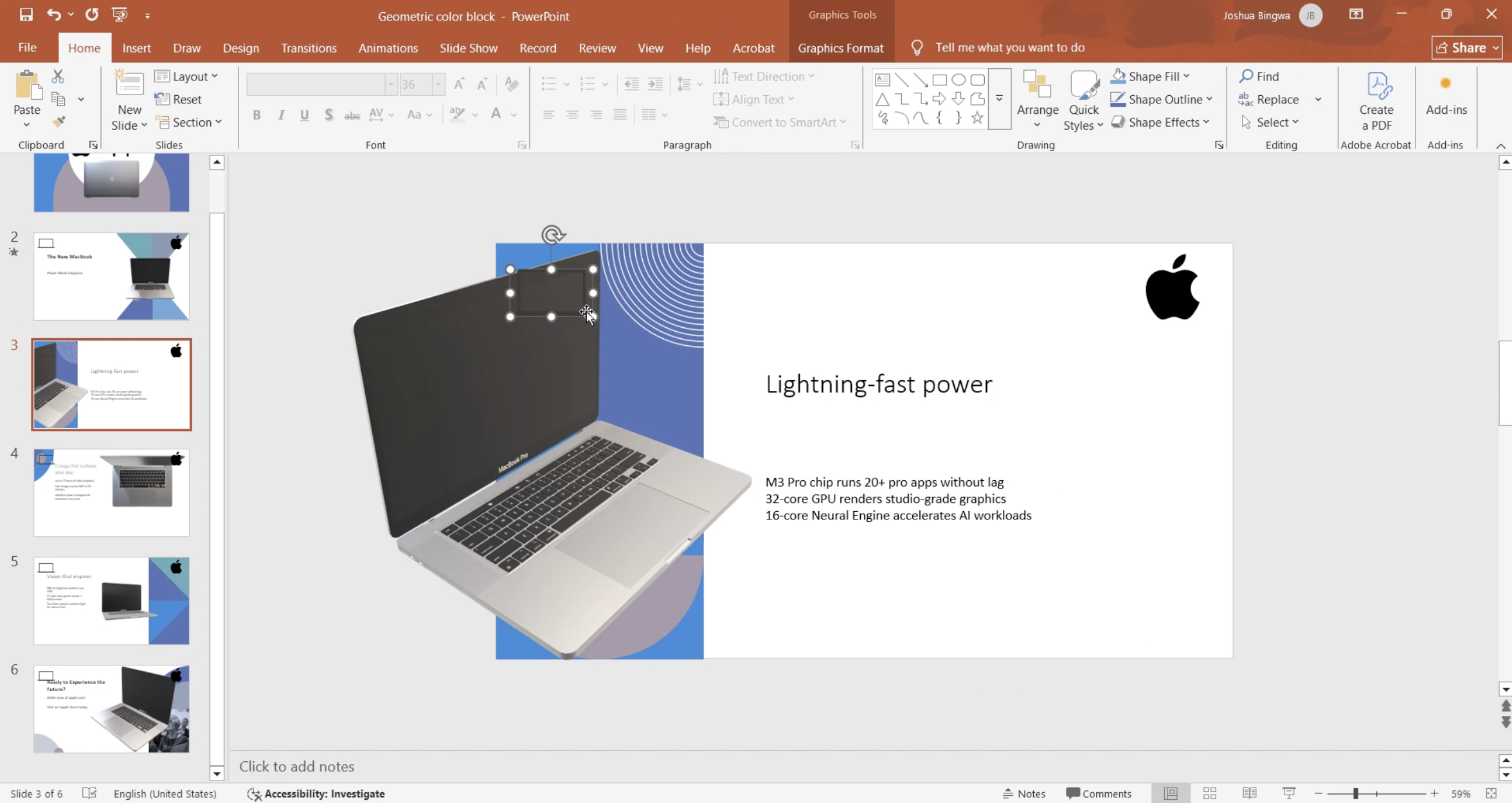}\\[3pt]
    \small \textit{``Paste, paste, paste, paste. Done. Done.''}
\end{minipage}
&
\textbf{Software}: PowerPoint

\textbf{Task}: Create a product pitch deck highlighting a product’s key features.

\textbf{Intent}:

A: Align the logos with the main text boxes.

B: Delete the logos from all the slides.

\textbf{C: Duplicate the logos onto the remaining slides.}

D: Change the color of the logos on all slides.
\\

\bottomrule
\end{tabularx}
\captionof{table}{Example instances for the (2) Intent Prediction task, showing screenshots, think-aloud narration, and the corresponding intent.}
\end{center}

\clearpage

\subsection{Help Prediction}
\begin{center}

\small
\begin{tabularx}{\textwidth}{m{0.50\textwidth} >{\raggedright\arraybackslash}X}
\toprule
\textbf{Screenshot} & \textbf{Help} \\
\midrule

\begin{minipage}{0.50\textwidth}
 \vspace{8pt}
    \centering
    \includegraphics[width=0.85\linewidth]{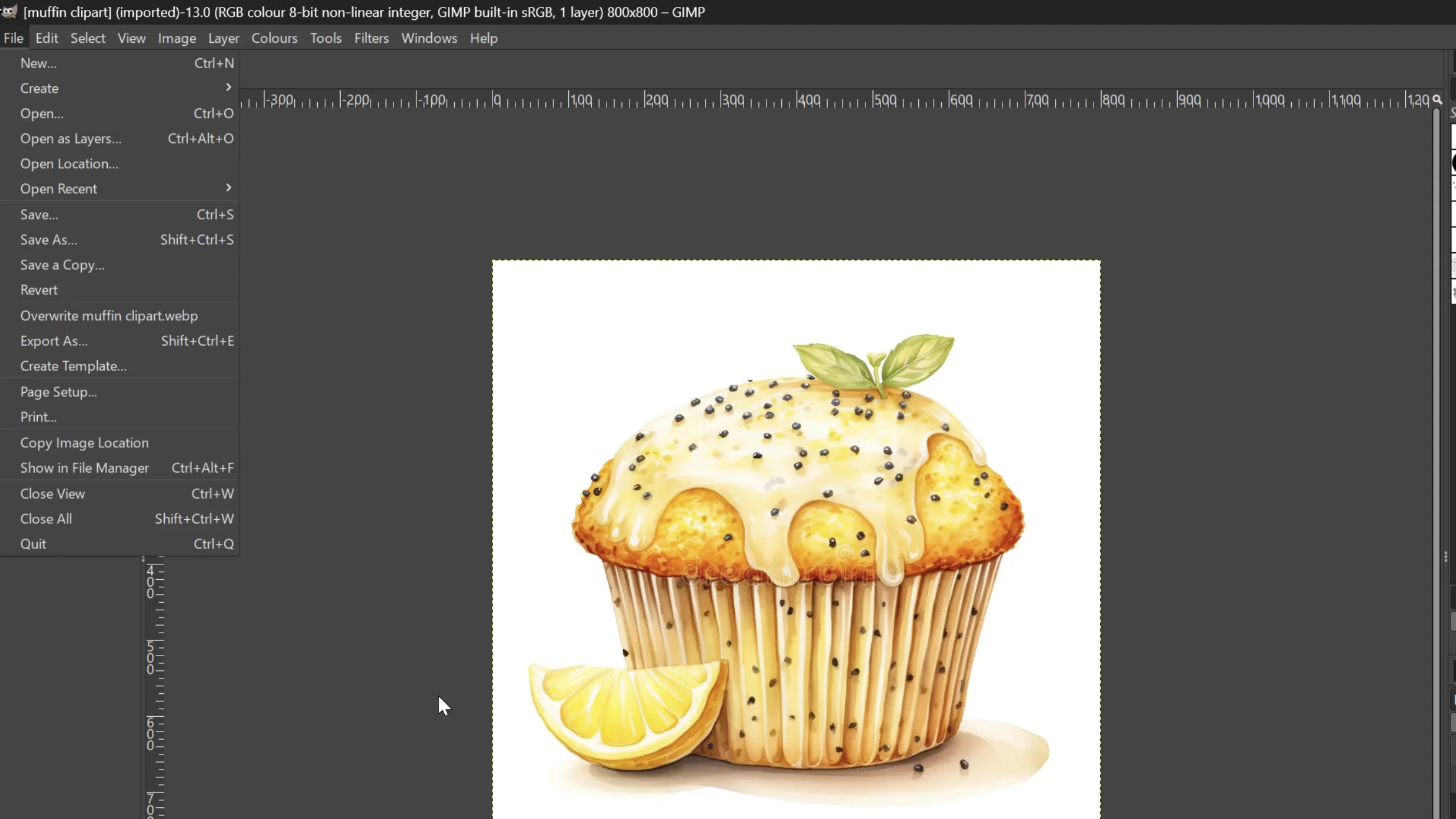}\\[3pt]
    
    \small \textit{``Where could I insert the text? [...] I'm just going to, because the help function I don't quite understand, but I can see if I can add it. Find it in Google.''}
\end{minipage}
&
\textbf{Software}: GIMP

\textbf{Task}: Create a bakery logo with a warm, friendly identity.

\textbf{Help Content}:

A: how to add another image as a layer      

\textbf{B: find the tool to add text}

C: remove the image background

D: add a background color or shape
\\

\begin{minipage}{0.50\textwidth}
 \vspace{8pt}
    \centering
    \includegraphics[width=0.85\linewidth]{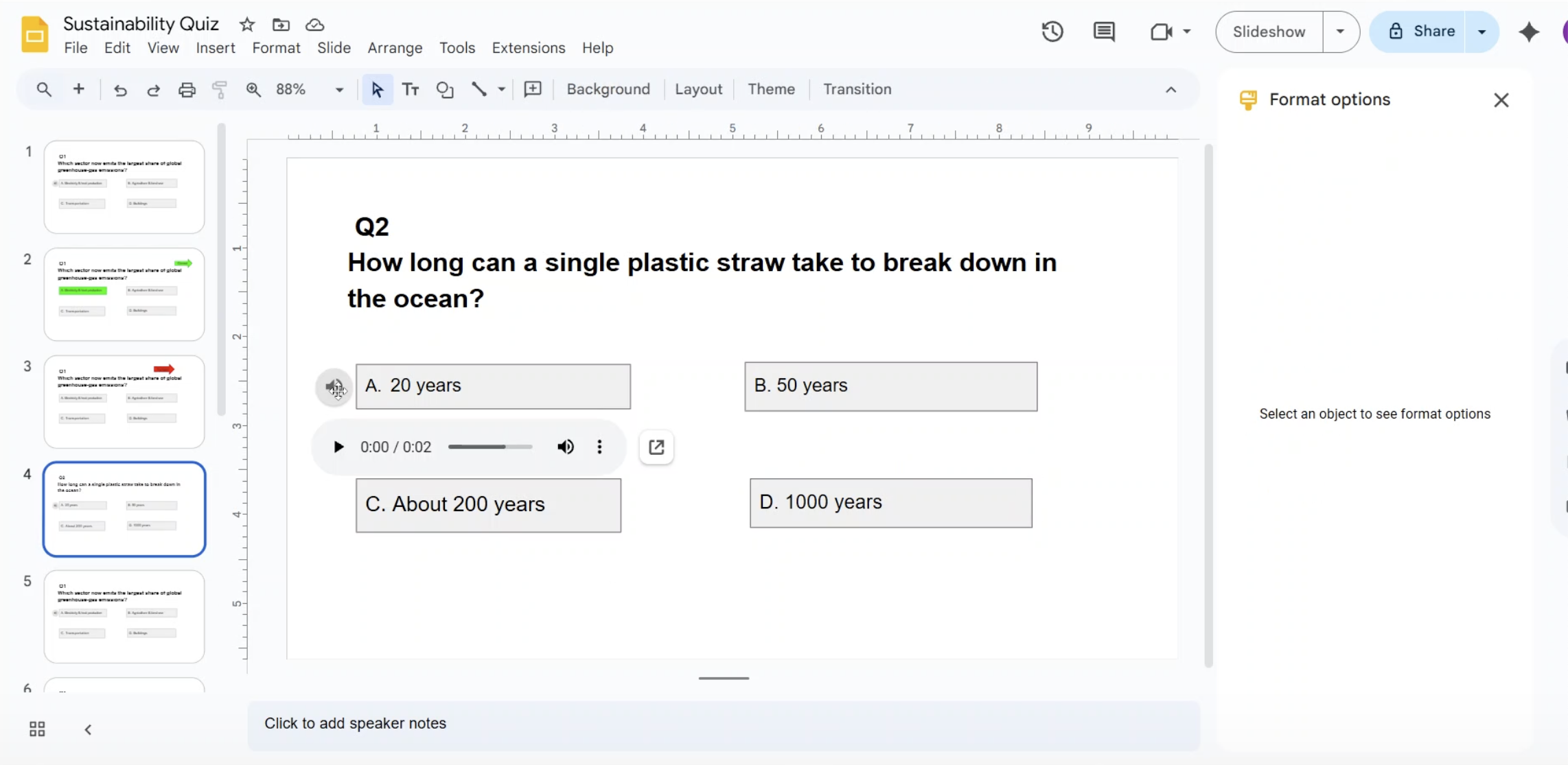}\\[3pt]
    \small \textit{``I think I made a mistake here and I need to rectify this.''}
\end{minipage}
&
\textbf{Software}: Google Slides

\textbf{Task}: Create a quiz deck with multiple-choice questions testing sustainability facts

\textbf{Help Content}:

A: align the answer choice boxes

B: how to create a quiz slide template

\textbf{C: how to fix a self-identified audio related error}

D: add animation to reveal the correct answer
\\

\begin{minipage}{0.50\textwidth}
 \vspace{8pt}
    \centering
    \includegraphics[width=0.85\linewidth]{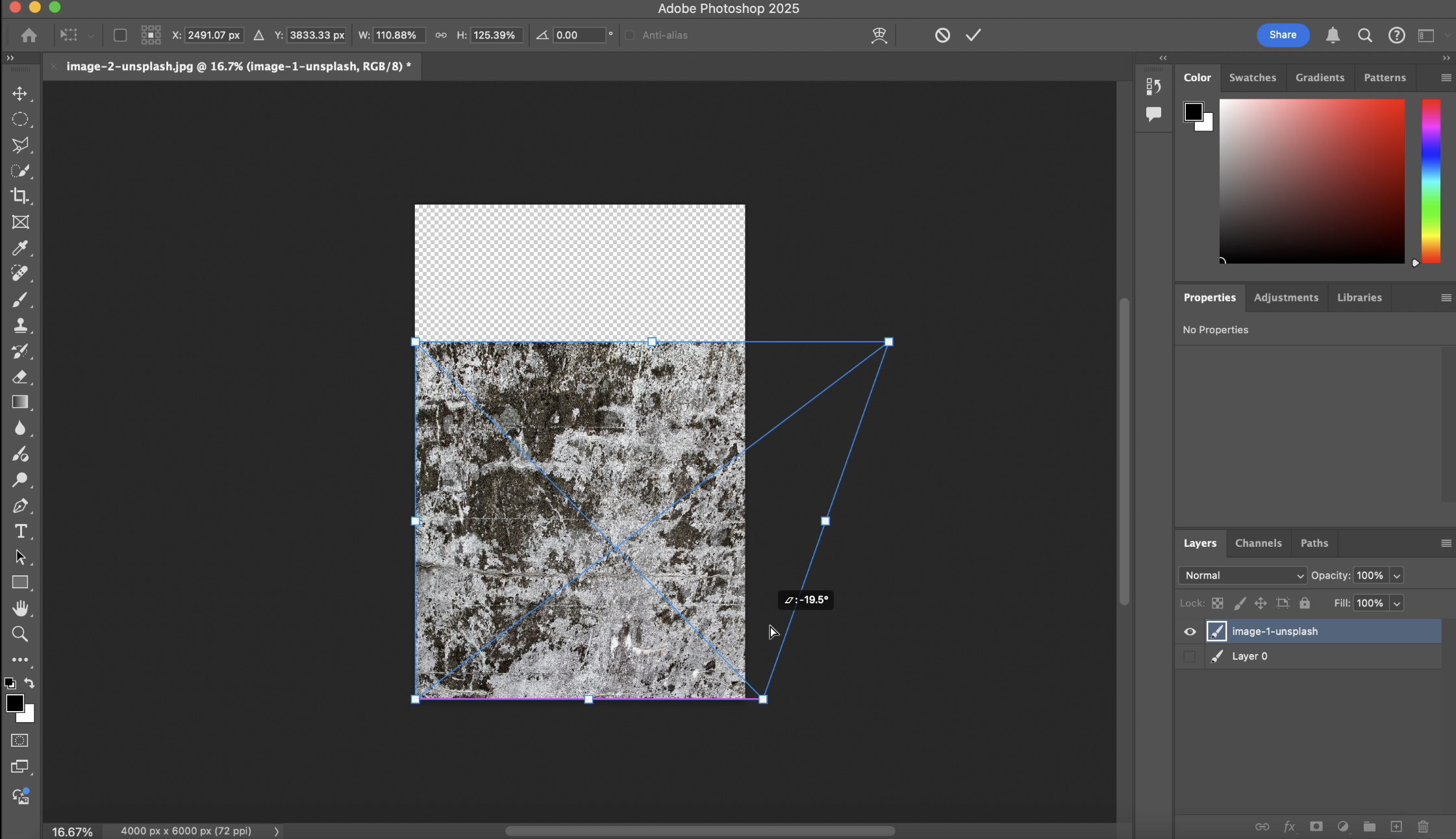}\\[3pt]
    \small  \textit{``I'll scale it. I just want to scale this up. How do I keep it?''}
\end{minipage}
&
\textbf{Software}: Photoshop

\textbf{Task}: Create a composite from two images.

\textbf{Help Content}:

A: how to use the perspective or warp transform tools   

B: center the new layer on the canvas

C: how to use layer blend modes

\textbf{D: maintain aspect ratio while scaling}
\\

\begin{minipage}{0.50\textwidth}
 \vspace{8pt}
    \centering
    \includegraphics[width=0.85\linewidth]{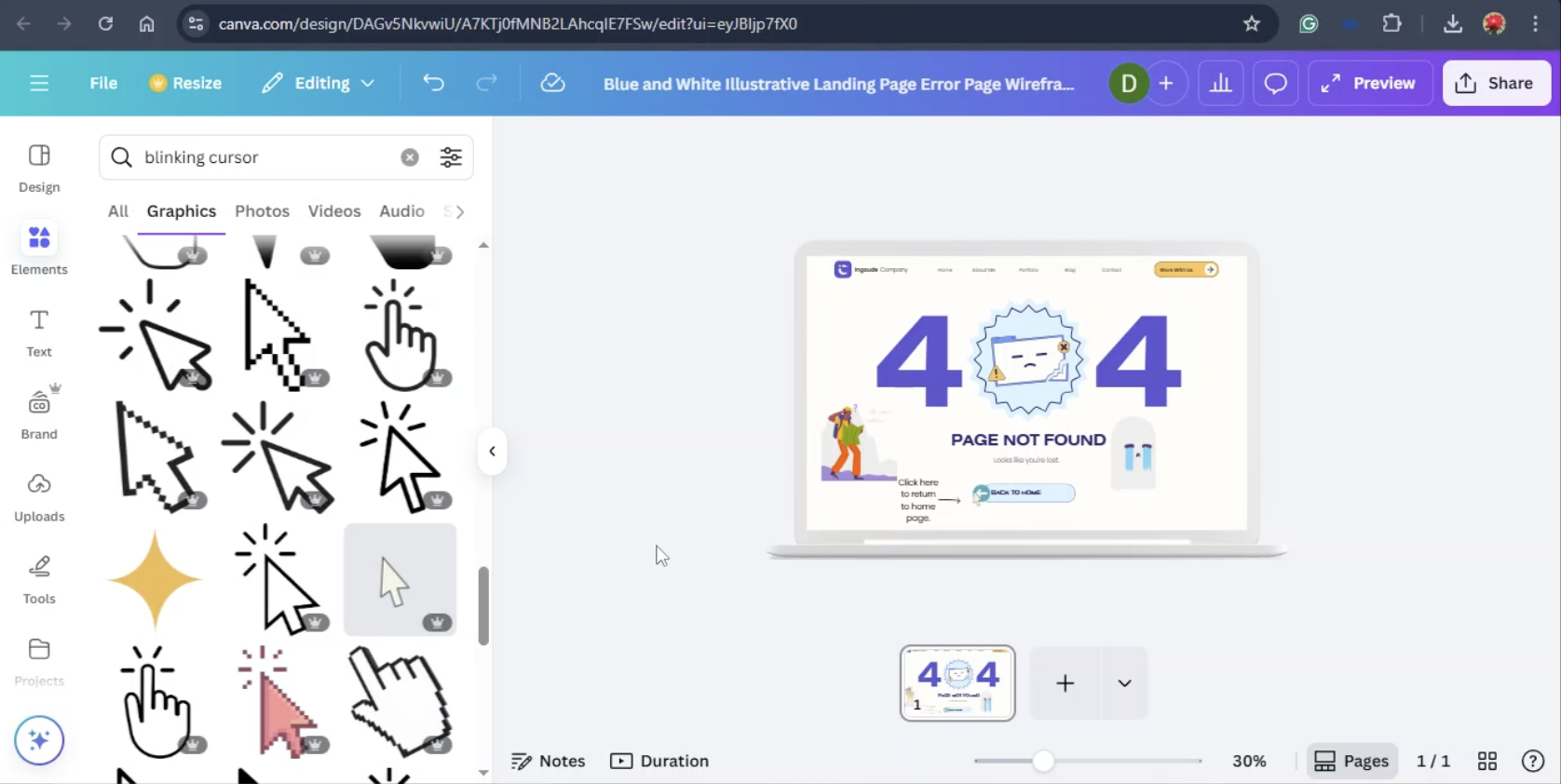}\\[3pt]
    \small \textit{``So I believe this is, this is great. I believe it's just simple.''}
\end{minipage}
&
\textbf{Software}: Canva

\textbf{Task}: Design a custom 404 error page with a visual and animated element.

\textbf{Help Need}:

A: help needed

\textbf{B: no help needed}
\\

\bottomrule
\end{tabularx}
\captionof{table}{Example instances for the (3) Help Prediction task. For the Help Need Detection task, the top three instances illustrate cases labeled as \textit{help needed}, while the last row shows an instance labeled as \textit{no help needed.}}
\end{center}

\clearpage

\section{Software Task Outcome Examples}
\begin{figure}[h]
    \centering
    \vspace{3em}

\begin{minipage}[t]{0.45\textwidth}
    \centering
    \begin{tabular}{cc}
        \includegraphics[width=0.48\linewidth]{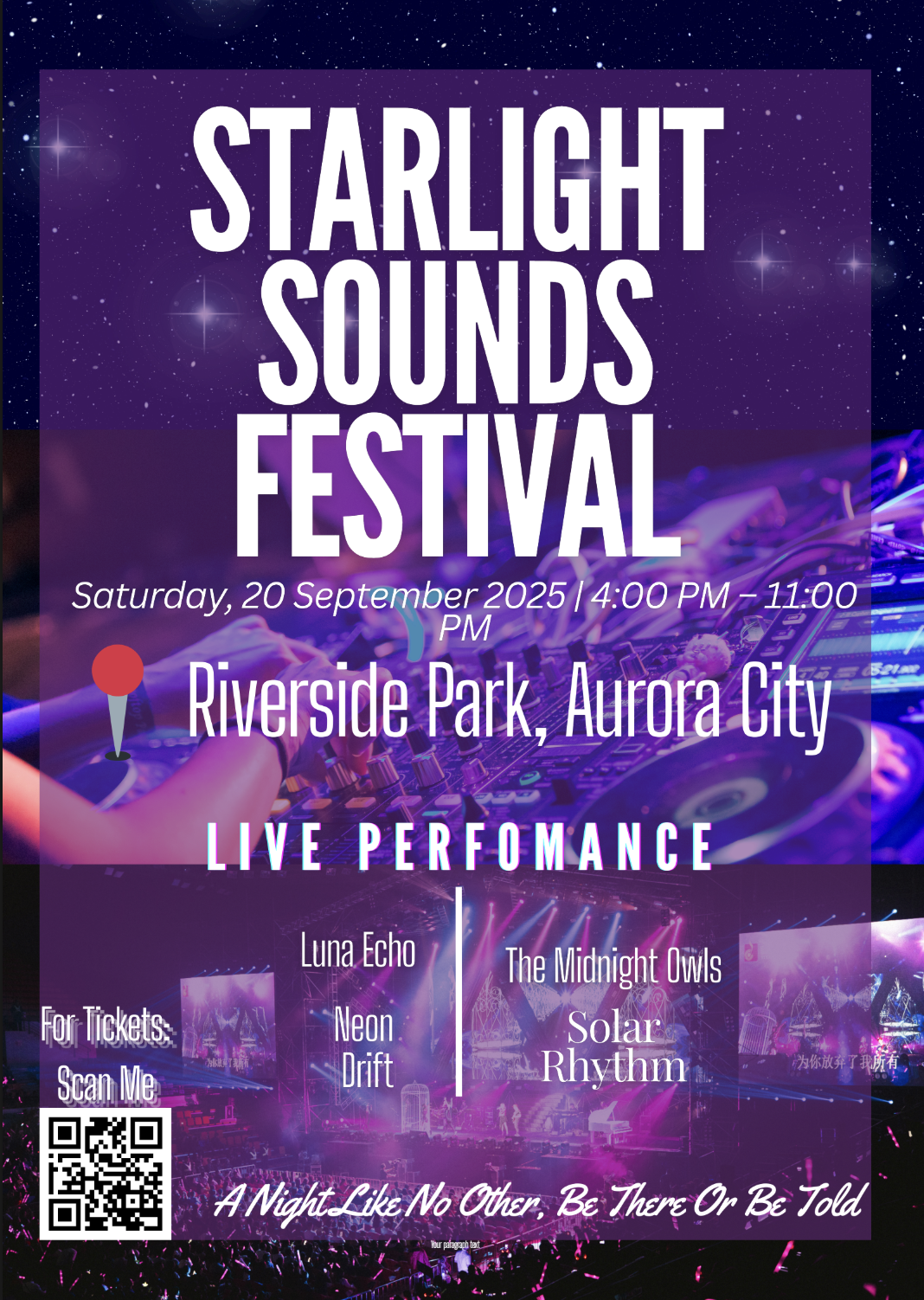} &
        \includegraphics[width=0.48\linewidth]{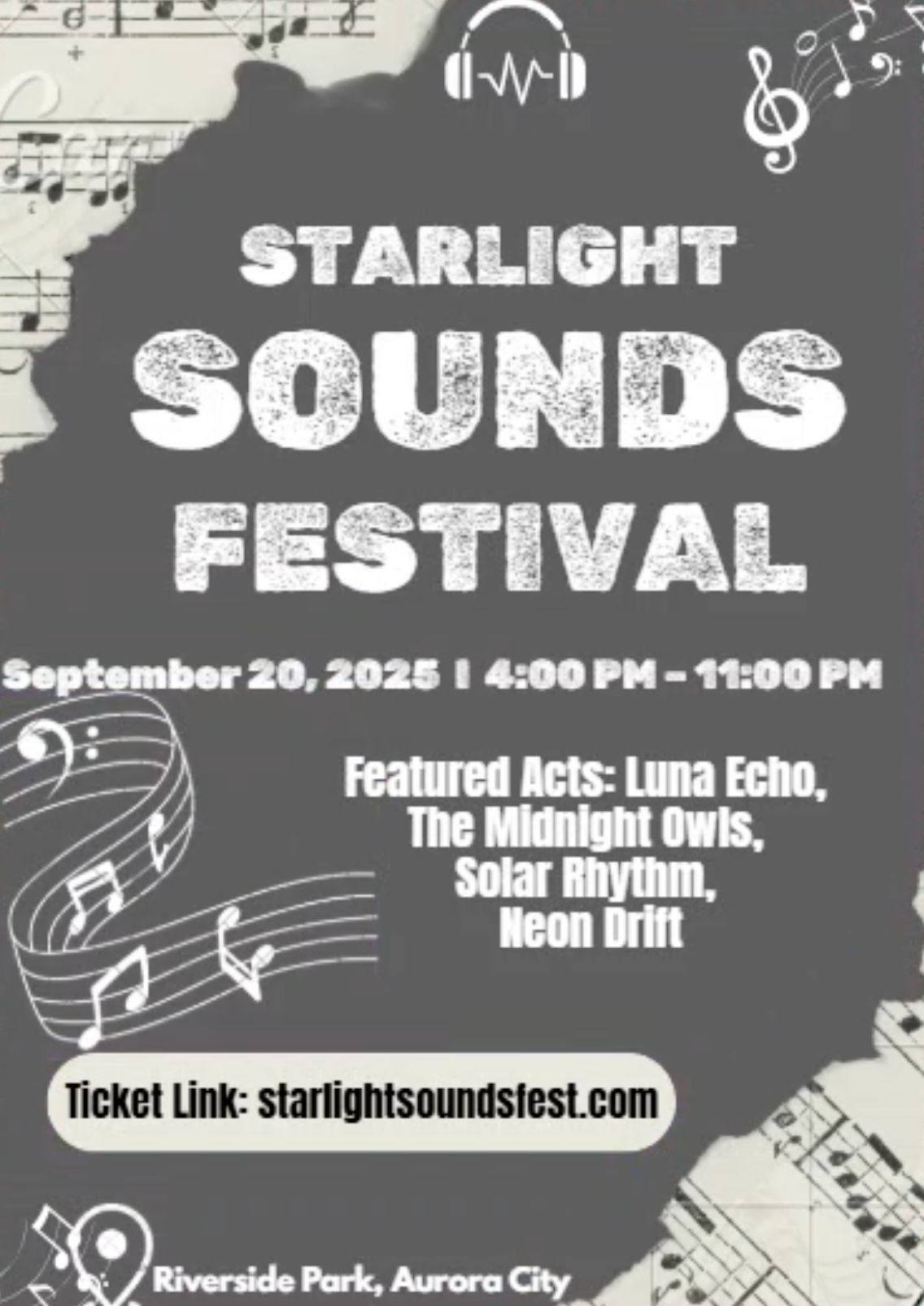} \\
        [6pt]
        \includegraphics[width=0.48\linewidth]{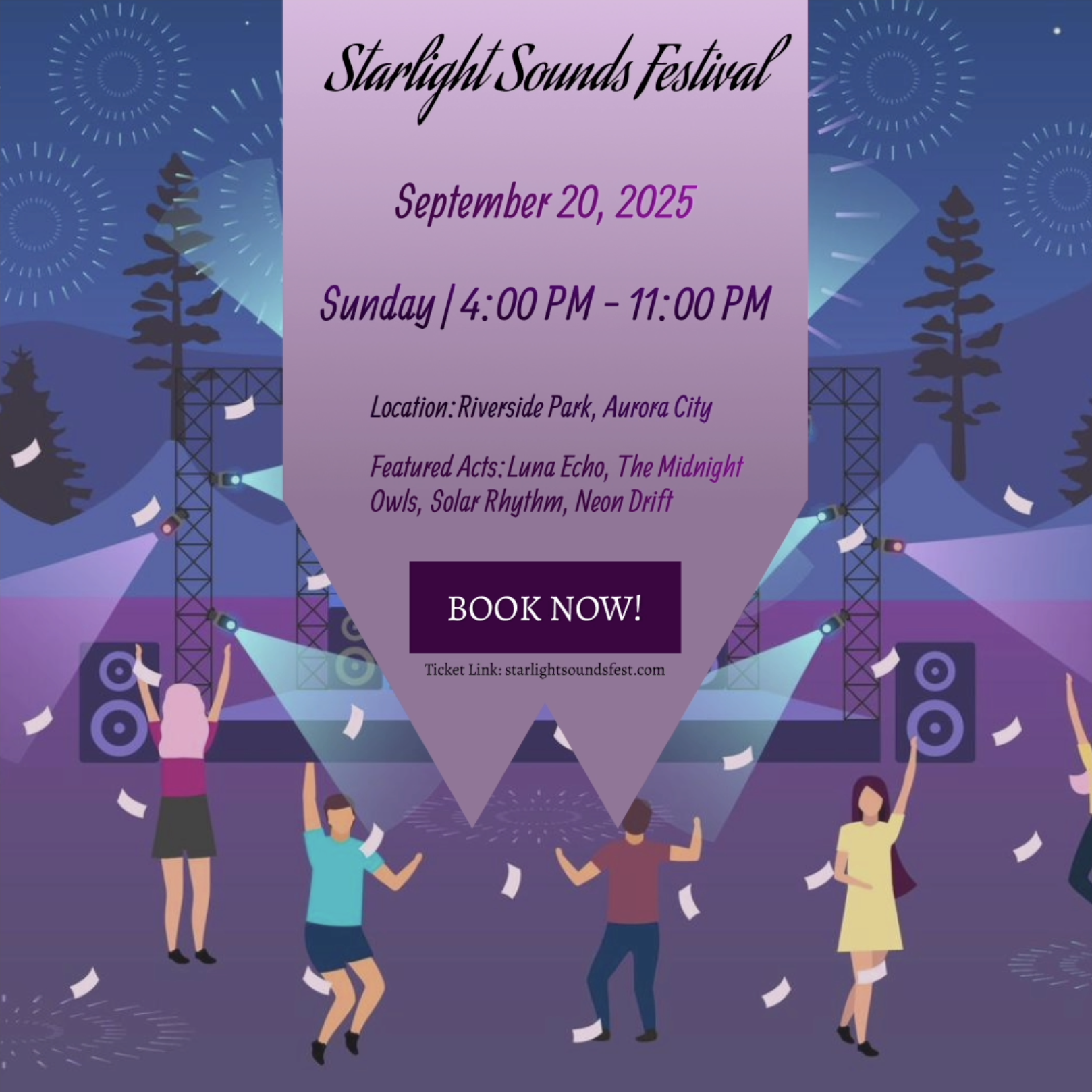} &
        \includegraphics[width=0.48\linewidth]{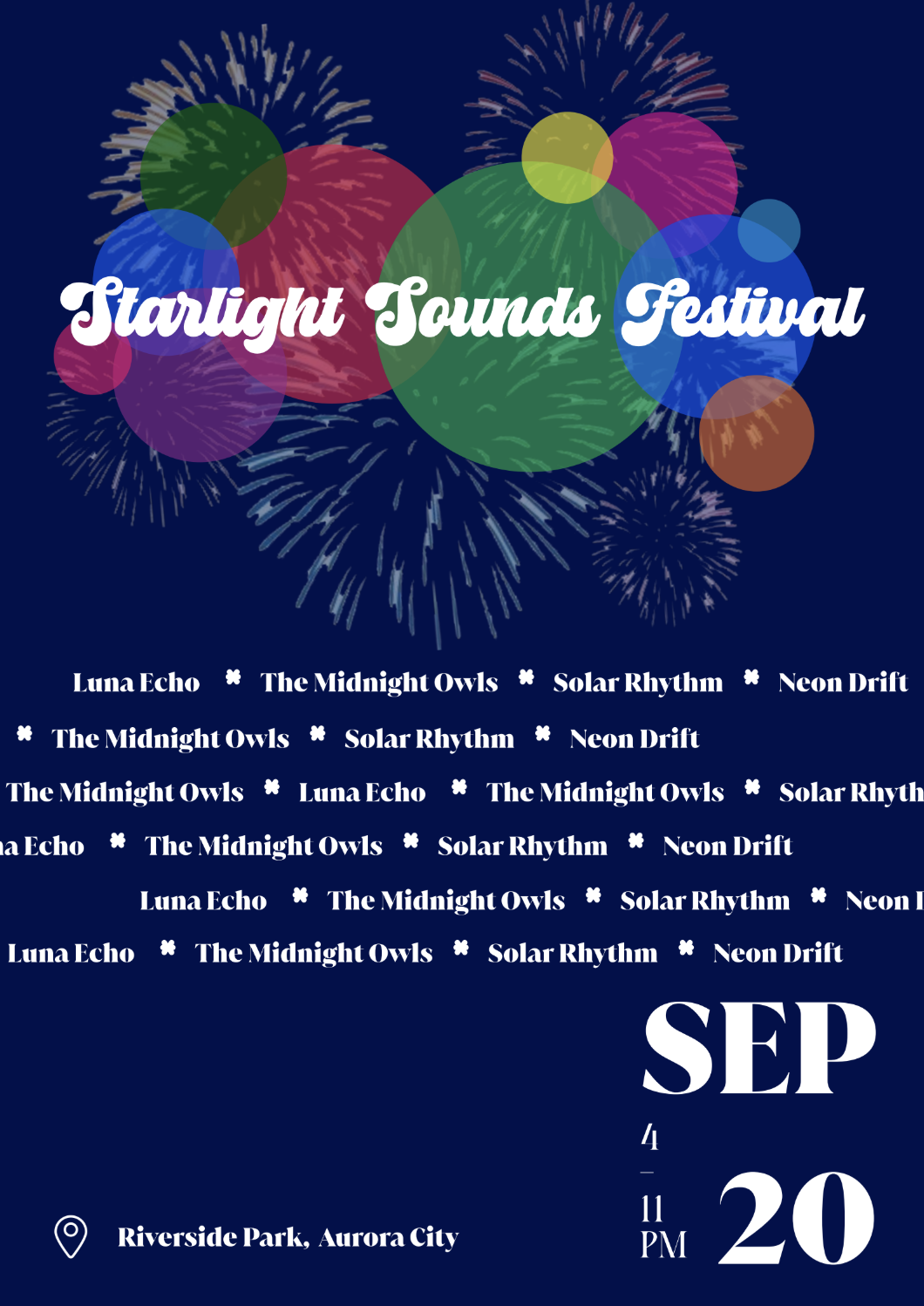} \\
    \end{tabular}
    \caption*{(a) Music event poster design in \textbf{Canva} (top) and \textbf{Figma} (bottom).}
\end{minipage}
\hfill
\begin{minipage}[t]{0.45\textwidth}
    \centering
    \begin{tabular}{cc}
        \includegraphics[width=0.48\linewidth]{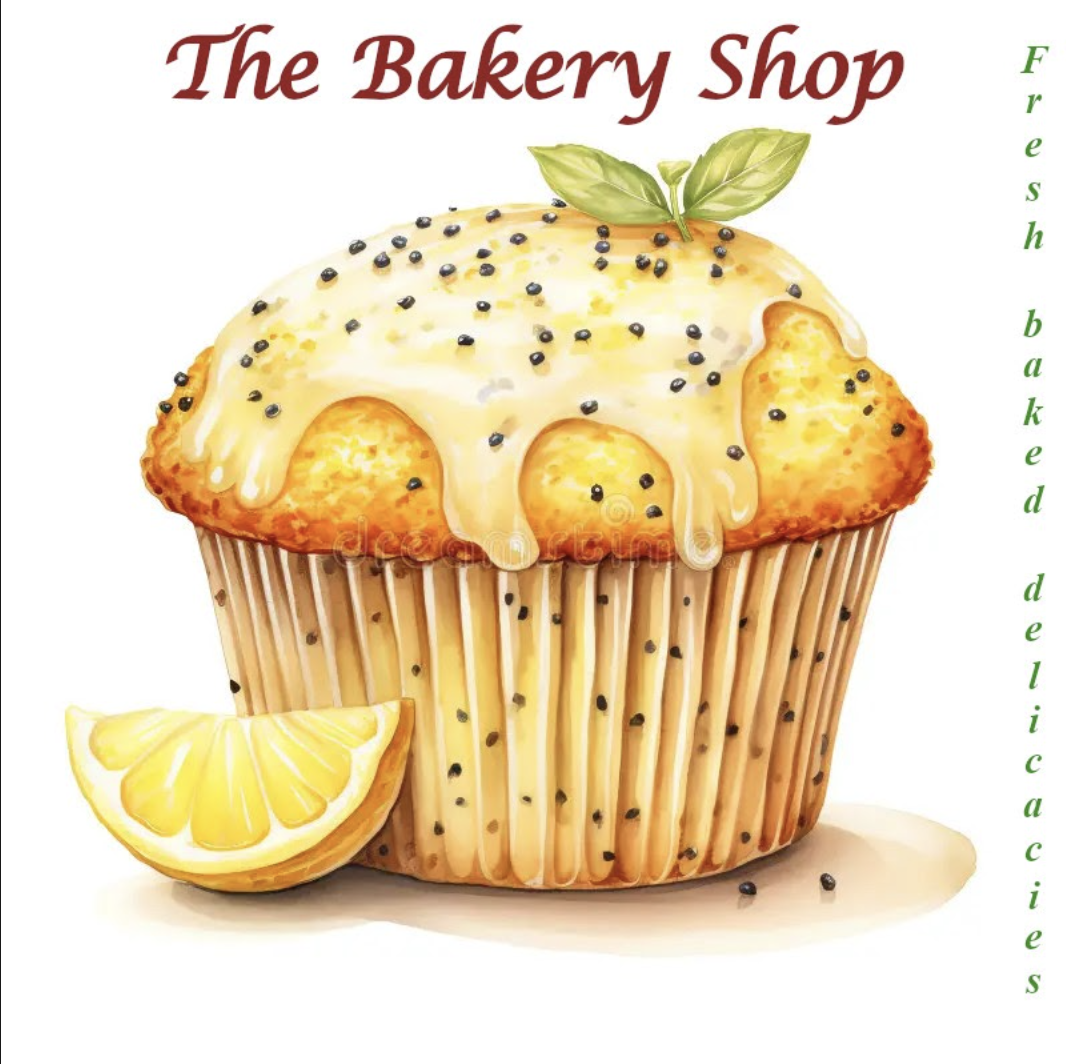} &
        \includegraphics[width=0.48\linewidth]{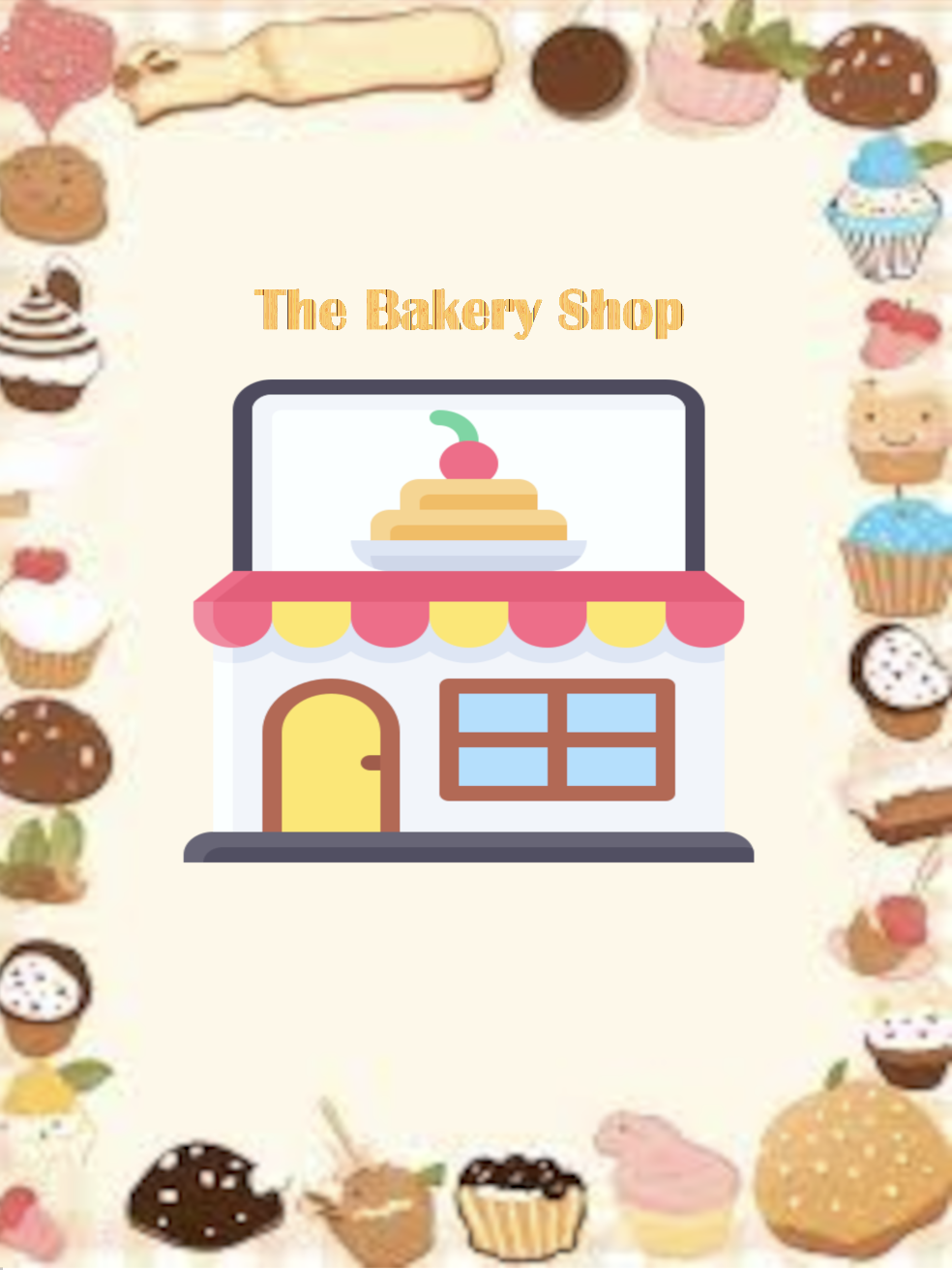} \\[6pt]
        \includegraphics[width=0.48\linewidth]{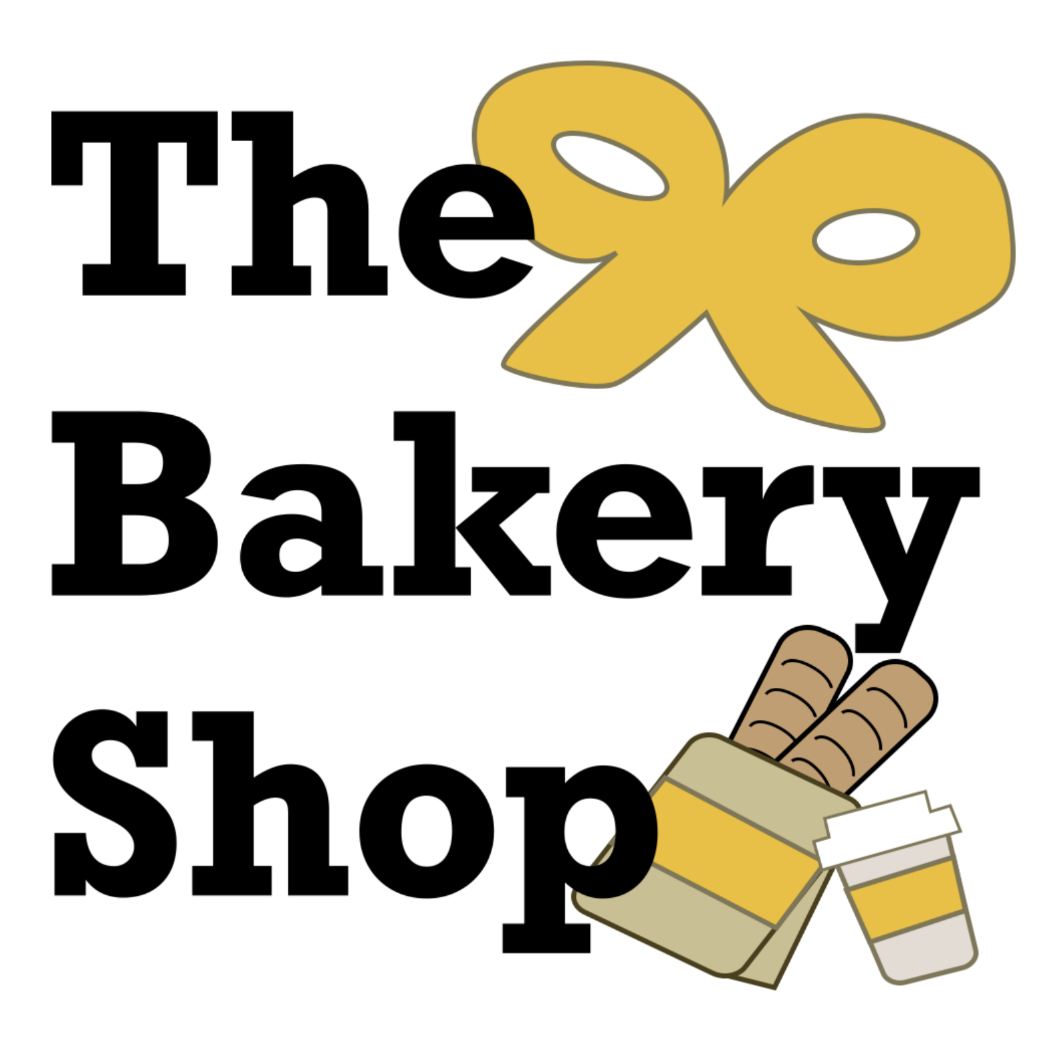} &
        \includegraphics[width=0.48\linewidth]{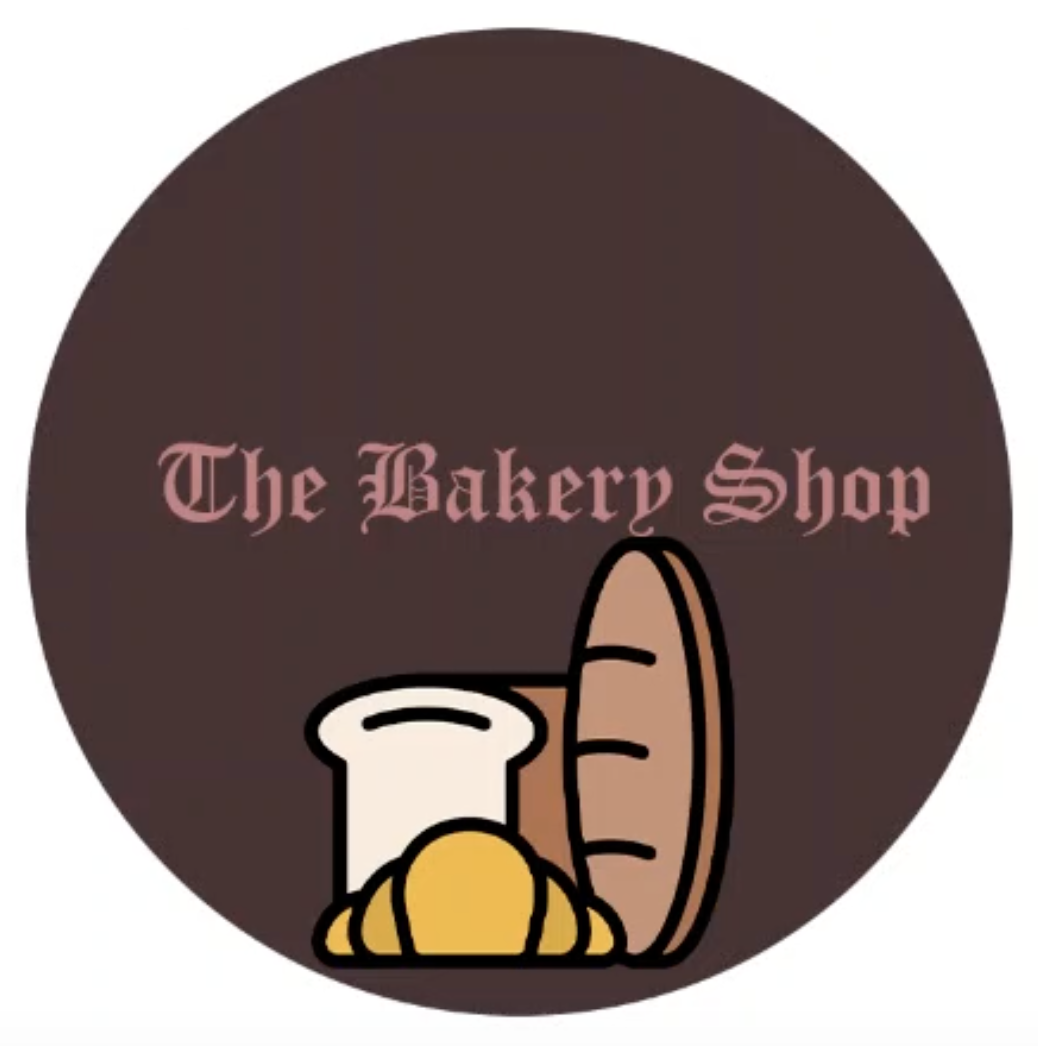} \\
    \end{tabular}
    \caption*{(b) Bakery logo design in \textbf{GIMP} (top) and \textbf{Photoshop} (bottom).}
\end{minipage}
\caption{Example outcomes of the assigned tasks.
The diversity across outputs reflects the open-ended nature of the tasks.}
\label{fig:outcomes}
\end{figure}

Figure~\ref{fig:outcomes} presents final artifacts produced by participants from the study. These examples highlight the open-ended nature of the assigned tasks. Despite receiving identical high-level instructions---such as ``Design a poster for a music festival" or ``Create a friendly bakery logo"---users produced markedly different results in terms of layout, aesthetic style, and complexity.
This diversity confirms that the study elicited non-linear, creative workflows rather than fixed execution. 

\clearpage

\section{Human Verification Interface}

\begin{figure*}[h]
  \centering
  \includegraphics[width=0.99\linewidth]{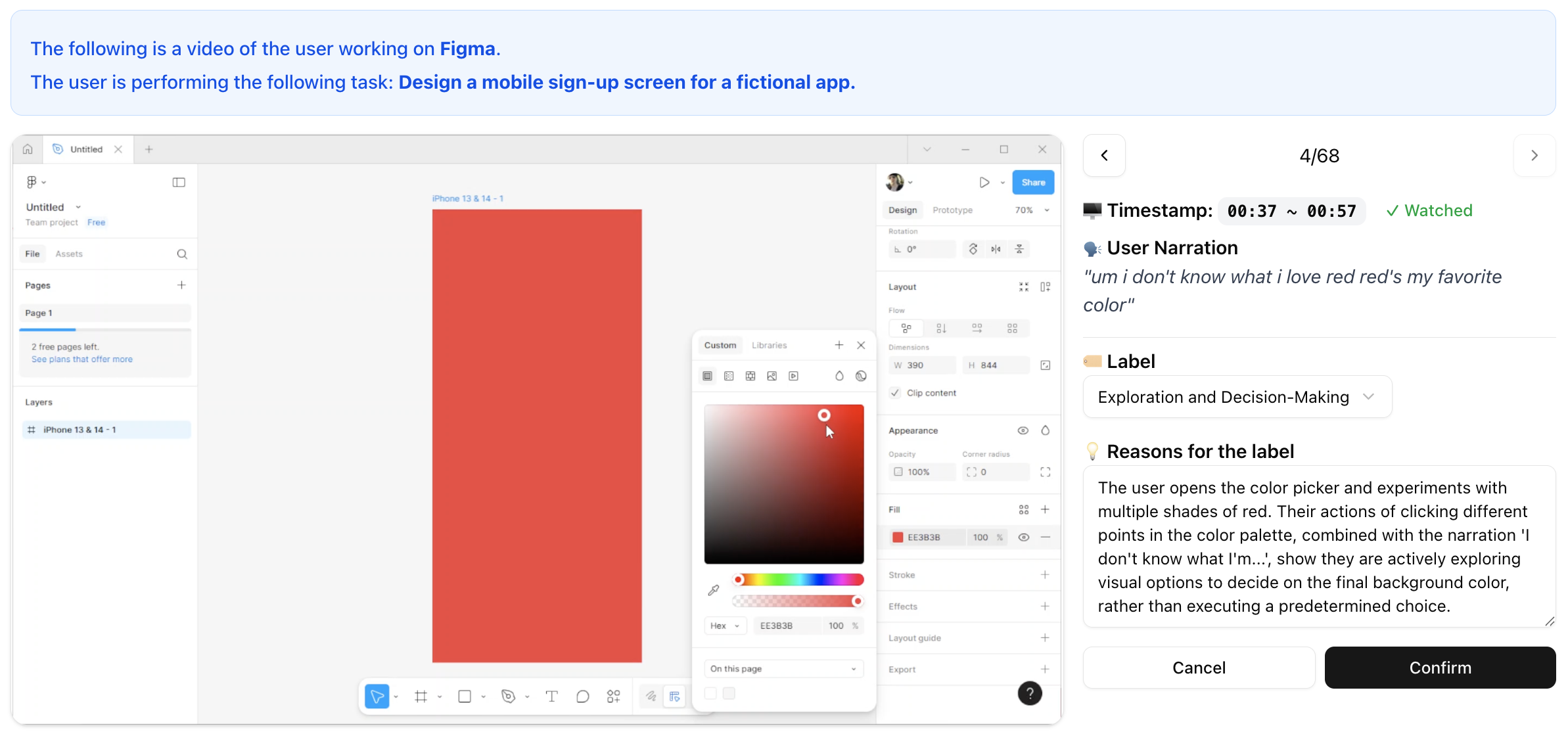}
  \caption{Annotation interface for validating and refining LLM-generated behavior-state labels. Annotators reviewed the predicted labels and the associated reasoning, correcting them if inaccurate. Each video segment was independently verified by two external annotators.}
  \label{fig:annotation_interface}
\end{figure*}

\begin{figure*}[h]
  \centering
  \includegraphics[width=0.99\linewidth]{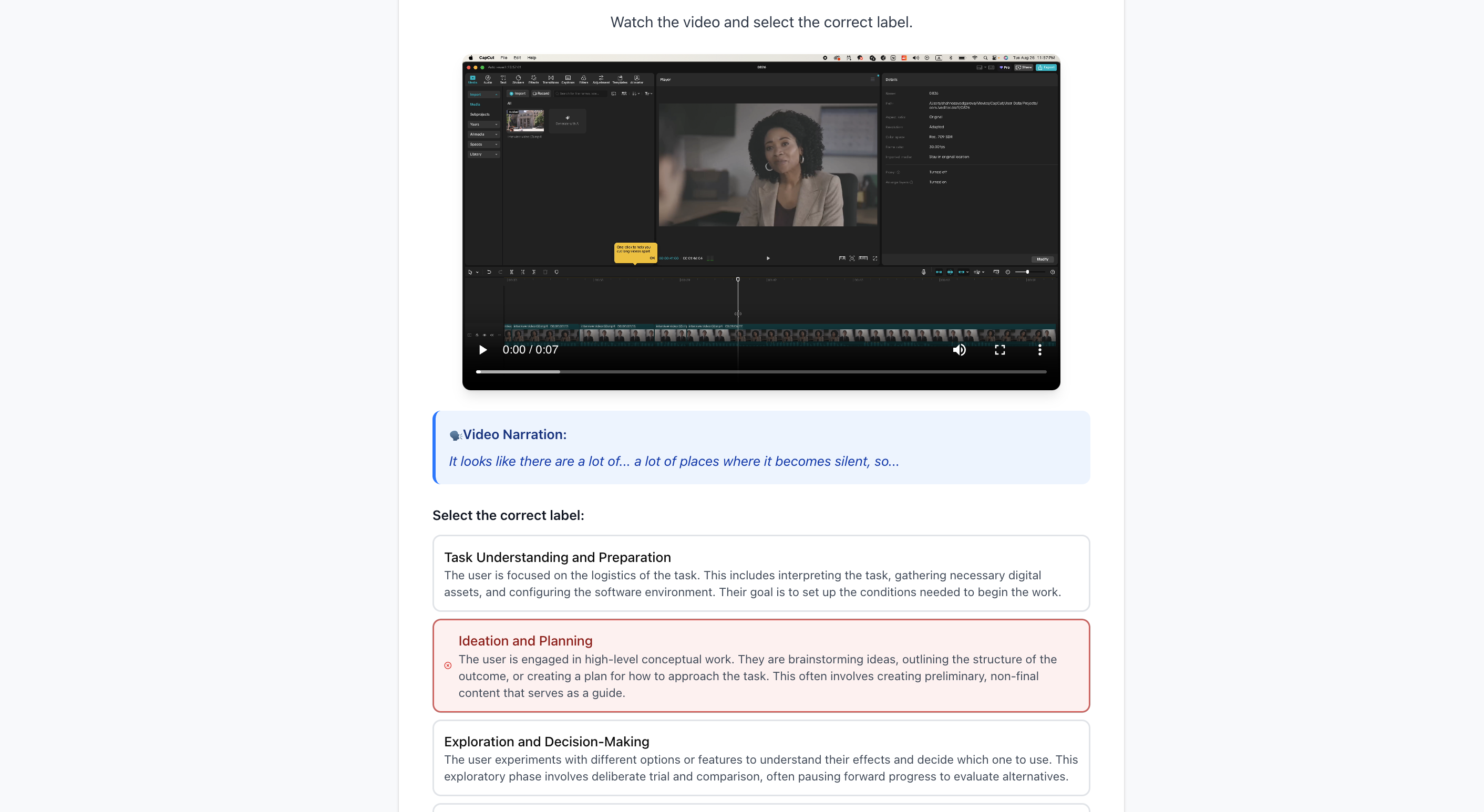}
  \caption{Before participating in the annotation, annotators completed a quiz phase where they had to correctly classify example video segments. This process ensured that all annotators possessed a solid understanding of the behavior taxonomy and definitions.}
  \label{fig:annotation_interface}
\end{figure*}

\clearpage
\section{Prompts}\label{sec:prompts}
\subsection{Taxonomy of User Behavior State Generation}

\begin{figure*}[h]
\begin{tcolorbox}[width=\linewidth, fontupper=\scriptsize]

\textbf{Taxonomy of User Behavior State Generation} \\

\hrule

\begin{Verbatim}[breaklines]
# Goal: Create a comprehensive taxonomy of user mental and behavior states by analyzing the video recording and transcript. Integrate visual observations (screen interactions, UI changes, cursor behavior) with audio/verbal cues (tone, hesitations, verbal expressions) and transcript content (exact quotes, semantic meaning).

# Analysis Guidelines:
- VISUAL EVIDENCE: Describe what you see on screen (tool selections, menu interactions, visual feedback, cursor patterns)
- AUDIO EVIDENCE: Note tone of voice, hesitations, exclamations, and vocal expressions
- TRANSCRIPT EVIDENCE: Extract precise quotes that reveal mental states and intentions
- CROSS-REFERENCE: Connect visual actions with verbal expressions to understand user intent and mental state

# Output Format (return as JSON):
{{
    "taxonomy": [
        {{
            "label": "…",
            "definition": "…",
            "evidence": [
                {{
                    "timestamp": "00:01:32",
                    "modality": "visual",
                    "description": "User clicks on the brush tool and immediately switches to eraser",
                    "significance": "Indicates uncertainty or trial-and-error behavior"
                }},
                {{
                    "timestamp": "00:01:35", 
                    "modality": "audio",
                    "description": "User says 'hmm, that's not right' with a frustrated tone",
                    "significance": "Verbal confirmation of confusion and frustration"
                }},
                {{
                    "timestamp": "00:01:35",
                    "modality": "transcript",
                    "quote": "hmm, that's not right, let me try something else",
                    "significance": "Shows problem-solving mindset and willingness to iterate"
                }}
            ]
        }},
        ...
    ]
}}

# Context:
Software: {SOFTWARE}
Task: {TASK_NAME}

# Transcript:
{TRANSCRIPT_JSON}

# Video Content:
SEE THE ATTACHED FILE.

\end{Verbatim}
\end{tcolorbox}
\caption{Prompt to generate a taxonomy of user behavior states given demonstration videos.}
\end{figure*}

\clearpage
\subsection{Data Annotation}

\begin{figure*}[h]
\begin{tcolorbox}[width=\linewidth, fontupper=\scriptsize]

\textbf{Behavior State Annotation} \\

\hrule

\begin{Verbatim}[breaklines]
# Goal
You are given a screen recording video snippet of a user using the software {SOFTWARE}. Annotate the video with the provided taxonomy of user mental and behavior states. Include the taxonomy label and reasoning that explains why the label is appropriate for the video. 

# Instructions
1. Your annotation label must be based on the user's current, on-screen behavior shown in the video.
2. Annotate video based on what you see, but if the label is not clear, use the think-aloud narration as auxiliary data to understand the user's intent or thought process.
3. Be aware that the user's narration may refer to past actions or future plans. Always align your annotation label with the user's current behavior at that specific moment in the video.

# Output Format (JSON)
{{
    "label": "…" // one of the labels in the taxonomy,
    "reasoning": "…" // Explanation on why the label is appropriate for the video,
}}

# Full Video Context
Software: {SOFTWARE}
Task the user is performing in the full screen recording video: {TASK_NAME}
The full transcript of the user: {TRANSCRIPT_JSON}

# Target Video Snippet Context
- Video snippet time range: 
{TRANSCRIPT_JSON[narration_index]["start"]} - {TRANSCRIPT_JSON[narration_index]["end"]}
Narration sentences of the video snippet: 
"{TRANSCRIPT_JSON[narration_index]["sentence"]}"

# Taxonomy
{TAXONOMY}

# Video Content
SEE THE ATTACHED FILE.

\end{Verbatim}
\end{tcolorbox}
\caption{Prompt to annotate a given video segment based on the taxonomy of user behavior states.}
\end{figure*}

\begin{figure*}[h]
\begin{tcolorbox}[width=\linewidth, fontupper=\scriptsize]

\textbf{Intent Annotation} \\

\hrule

\begin{Verbatim}[breaklines]
You are analyzing a user's screen recording video and think-aloud narration while they use software (e.g., video editing, design, or spreadsheet tools).

# Goal
For the given video snippet and its corresponding narration segment, infer what the user was aiming to complete or achieve by the end of this segment - their short-term goal or intention.

- The goal should represent **an outcome or result** that the user was either finishing or actively working on as the segment ends.
- Focus on **tangible, result-oriented goals** (e.g., "finish trimming the clip," "adjust the image color," "complete text alignment").
- Ignore interface-level or purely procedural descriptions (e.g., "click this," "open that," "drag the layer").
- If no clear goal or outcome is expressed or shown, return "no tangible goal".

# Output Format (JSON)
{{
    "original_narration": "<verbatim narration text>",
    "goal": "<concise description of what the user aimed to complete or was completing by the end of this segment, or 'no tangible goal'>",
    "evidence_narration_snippet": "<the exact portion of the narration text that supports this inferred goal>",
    "reasoning": "<brief explanation of how this goal was inferred based on the narration and video>"
}}

# Full Video Context
Software: {SOFTWARE}
Task the user is performing in the full screen recording video: {TASK_NAME}
The full transcript of the user: {TRANSCRIPT_JSON}

# Target Video Snippet Context
- Video snippet time range: {TRANSCRIPT_JSON[narration_index]["start"]} - {TRANSCRIPT_JSON[narration_index]["end"]} seconds
- Narration during this snippet: "{TRANSCRIPT_JSON[narration_index]["sentence"]}"

# Video Content
SEE THE ATTACHED FILE.
\end{Verbatim}
\end{tcolorbox}
\caption{Prompt used to annotate a given video segment with the user's intent.}
\end{figure*}

\begin{figure*}[h]
\begin{tcolorbox}[width=\linewidth, fontupper=\scriptsize]

\textbf{Help Annotation} \\

\hrule

\begin{Verbatim}[breaklines]

You are analyzing a user's screen recording video and think-aloud narration while they use software (e.g., video editing, design, or spreadsheet tools).

# Goal
For the given video snippet and its corresponding narration segment, infer **what kind of help or guidance the user is looking for**.

- Focus on identifying explicit requests for help or moments where the user seeks information, clarification, or suggestions.
- Help-seeking can appear in two main forms:
  1) **On-Screen Help Behavior:** The user performs on-screen actions to seek help (e.g., opening a web browser, typing a query into a search engine or LLM, viewing online documentation or tutorials).
  2) **Narration-Based Help Expression:** The user verbally expresses confusion, uncertainty, or asks questions (e.g., "I don’t know how to fix this," "Why is this not showing up?", "How do I do this?").
- Ignore casual statements or comments unrelated to problem-solving.
- If there is no clear indication that the user is seeking help, or if the type of help they need is implicit or ambiguous, return "no help needed".

# Output Format (JSON)
{{
    "help_needed": "<concise but meaningful description of the help sought - focus on the underlying need, such as 'explain masking feature', 'suggest alternative filter', 'debug export error', or 'clarify timeline snapping'",
    "help_source": "<'screen', 'narration', 'both', or 'none'>",
    "evidence_narration_snippet": "<exact portion of narration that indicates help-seeking (if any)>",
    "evidence_screen_behavior": "<description of what was seen on screen that indicates help-seeking (if any)>",
    "reasoning": "<brief explanation of how the need for help was inferred based on the narration or/and on-screen behavior>"
}}

# Full Video Context
Software: {SOFTWARE}
Task the user is performing in the full screen recording video: {TASK_NAME}
The full transcript of the user: {TRANSCRIPT_JSON}

# Target Video Snippet Context
- Video snippet time range: {TRANSCRIPT_JSON[narration_index]["start"]} – {TRANSCRIPT_JSON[narration_index]["end"]} seconds
- Narration during this snippet: "{TRANSCRIPT_JSON[narration_index]["sentence"]}"

# Video Content
SEE THE ATTACHED FILE.
\end{Verbatim}
\end{tcolorbox}
\caption{Prompt used to annotate a given video segment with whether help is needed and, if so, what specific help is required.}
\end{figure*}

\begin{figure*}[h]
\begin{tcolorbox}[width=\linewidth, fontupper=\scriptsize]

\textbf{Filtering On-Screen Help-Seeking Behavior} \\

\hrule

\begin{Verbatim}[breaklines]

You are analyzing a user's sequence of help-seeking actions while they use {SOFTWARE} for the task "{TASK_NAME}". 

You are given HELP_DATA, a list of chronological records where each record includes:
- index: original index of the record
- help: user's inferred help need (e.g., "find a graphic for a password field")
- screen_behavior: observed on-screen action
- narration: user's think-aloud narration

# Goal
- Keep only the help entries where the screen_behavior is about using external applications (e.g., Google Search, ChatGPT, Gemini, YouTube,etc.). Note that it doesn't include referring to the task instructions page.
- Return the list of segment ids that should be kept as a JSON object.

# Output Format (JSON)
Return only valid JSON:
{{
    "kept_segment_ids": [<int>, <int>, ...]
}}


# Data
HELP_DATA:
{HELP_JSON}
\end{Verbatim}
\end{tcolorbox}
\caption{Prompt used to filter on-screen help-seeking behavior from segments previously marked as help needed.}
\end{figure*}

\begin{figure*}[h]
\begin{tcolorbox}[width=\linewidth, fontupper=\scriptsize]

\textbf{Filtering Narration-Based Help-Seeking Behavior} \\

\hrule

\begin{Verbatim}[breaklines]

You are analyzing a user's sequence of actions while they use {SOFTWARE} for the task "{TASK_NAME}".

You are given HELP_DATA, a list of chronological records where each record includes:
- index: original index of the record
- help: user's inferred help need (e.g., "find a graphic for a password field")
- screen_behavior: observed on-screen action
- narration: user's think-aloud narration

# Goal
Keep only segments where the narration explicitly asks for help or guidance.

# Definition of explicit help
The narration contains a direct request for help or instruction, for example:
- "help me", "i need help", "i want help", "can you help", "please help"
- How or where questions about operating the software, such as:
  "how do i ...", "how can i ...", "how to ...", "where is ...", "which option should i ...",
  "what should i click", "what does this do".
- Requests for instructions or explanation:
  "show me how to ...", "tell me how to ...", "could someone explain ...", "is there a way to ..."

# Exclude the following
- Exploration or intent without a help request: "i'm going to try", "let me see", "i will search"
- Trial and error or self-correction without a request: "no, not that", "okay now i got it", "ah ok", "finally"
- Uncertainty alone: "maybe", "i think", "not sure" unless followed by a direct question that asks for guidance
- Statements addressed to self that do not ask for help: "i need to add text", "i'm looking for an icon"
- Generic questions not tied to getting guidance on what to do next

# Output Format
Return only valid JSON:
{{
    "kept_segment_ids": [<int>, <int>, ...]
}}

# Data
HELP_DATA:
{HELP_JSON}
\end{Verbatim}
\end{tcolorbox}
\caption{Prompt used to filter narration-based help-seeking behavior from segments previously marked as help needed.}
\end{figure*}

\begin{figure*}[h]
\begin{tcolorbox}[width=\linewidth, fontupper=\scriptsize]

\textbf{Filtering No Help Needed} \\

\hrule

\begin{Verbatim}[breaklines]

You are analyzing a user's sequence of actions while they use {SOFTWARE} for the task "{TASK_NAME}". 

You are given NO_HELP_DATA, a chronological list of records where each record includes:
- index: original segment index
- start_time, end_time: time range of the segment
- no_help_reasoning: explanation for why the user does not need help
- narration: the user's think-aloud narration

# Goal
Identify up to 5 segments where it is **explicitly clear** that the user does not need help.
These are moments when the user:
- Performs actions confidently, smoothly, and intentionally.
- Demonstrates clear understanding of what to do next without hesitation or correction.
- Speaks in a calm, matter-of-fact tone (e.g., "Now I'll add text here," "Perfect," "That looks good.").

# Exclude the following:
- **Trial and error:** any sign of experimentation, correction, or rapid alternation (e.g., "No, no, no," "Let me try again," "Okay, that worked.")
- **Self-resolution after confusion:** phrases like "now I got it," 'finally," "oh, that’s how," "ah okay," "I see," or any narration showing realization after failure or surprise.
- **Frustration or emotional reactions:** (e.g., "oh shit," "ugh," "why," "come on") even if followed by success.
- **Uncertain or exploratory speech:** "I think," "maybe," "let’s see," "try," "not sure."
- **Segments that merely lack confusion** but do not clearly express confidence or mastery.

# Rules
- Select at most 5 segments that best reflect calm, deliberate, fluent progress.
- Preserve the original order of the selected segments.
- Return only the segment indices of the kept entries.

# Output Format (JSON)
Return only valid JSON:
{{
    "kept_segment_ids": [<int>, <int>, ...]
}}

# Data
NO_HELP_DATA:
{NO_HELP_JSON}
\end{Verbatim}
\end{tcolorbox}
\caption{Prompt used to filter clear no-help-needed segments.}
\end{figure*}
\clearpage
\subsection{Model Evaluation}

\begin{figure*}[h]
\begin{tcolorbox}[width=\linewidth, fontupper=\scriptsize]

\textbf{Behavior State Detection} \\

\hrule

\begin{Verbatim}[breaklines]


# Goal
You are given a screen recording video snippet of a user working in {SOFTWARE}. 
Classify the video into one of the labels from the provided taxonomy of user mental and behavioral states.
Include both the taxonomy label and a reasoning that explains why this label best fits the observed segment.

# Instructions
1. Base your classification on the user's on-screen behavior shown in the video.
2. Provide:
    - label: one of the taxonomy labels
    - reasoning: a concise explanation of why this label fits the segment
3. Return the output strictly in valid JSON, with keys "label" and "reasoning". Do NOT wrap the JSON in markdown code blocks. Return only the raw JSON object.

# Output Format (JSON)     
{{
    "label": "…",
    "reasoning": "…"
}}


# Video Context
Software: {SOFTWARE}
Task performed in the full recording: {TASK_NAME}
Start and end times of the snippet (relative to the full recording): 
{start_time} - {end_time} seconds

# Taxonomy Descriptions
{TAXONOMY}

# Previous Segment Context (*** Optional based on the condition)
The user behavior in the immediately preceding segment was {previous_label}: {label_definition}

# Video Content
SEE THE ATTACHED FILE.
\end{Verbatim}
\end{tcolorbox}
\caption{Prompt used to evaluate the model on the (1) Behavior State Detection task.}
\end{figure*}

\begin{figure*}[h]
\begin{tcolorbox}[width=\linewidth, fontupper=\scriptsize]

\textbf{Intent Prediction} \\

\hrule

\begin{Verbatim}[breaklines]

# Goal
You are given a screen recording snippet of a user working in {SOFTWARE}.
Predict which option best describes the user's intention during this segment.

# Instructions
1. Watch the video and analyze on-screen actions.
2. Select the option (A-D) that best matches the goal of the user trying to achieve.
3. Use the provided behavior context to interpret the goal. (*** Optional based on the condition)
4. Return output in JSON:
   - label: one of A-D
   - reasoning: short explanation for your choice
5. Output only a valid JSON object (no Markdown).

# Output Format (JSON)
{"label": "A", "reasoning": "..."}

# Video Context
Software: {SOFTWARE}
Task performed in the full screen recording: {TASK_NAME}
Start and end times of the snippet (relative to the full recording): 
{start_time} - {end_time} seconds

# User Behavior Context (*** Optional based on the condition)
The following user behavior is identified: {label}: {label_definition}.
Consider this context when predicting the user's intention and selecting the most appropriate option.

# Options
{options_text}

# Video Content
SEE THE ATTACHED FILE.
\end{Verbatim}
\end{tcolorbox}
\caption{Prompt used to evaluate the model on the (2) Intent Prediction task.}
\end{figure*}

\begin{figure*}[h]
\begin{tcolorbox}[width=\linewidth, fontupper=\scriptsize]

\textbf{Help Need Detection} \\

\hrule

\begin{Verbatim}[breaklines]
# Goal
You are given a screen recording video snippet of a user working in {SOFTWARE}. 
Based on the video content, the user's intention, and the user behavior context (*** Optional based on the condition), determine if the user needs help or not in this segment.

# Instructions
1. Watch the video and observe the user's behavior and actions.
2. Consider the user's intention and behavior context provided to better understand what they are trying to accomplish and their current state.  (*** Optional based on the condition)
3. Determine if the user needs help or not.
4. Provide:
    - label: "yes" if the user needs help, "no" if they do not need help
    - reasoning: a concise explanation of why the user does or does not need help based on the observed behavior.
5. Output only a valid JSON object (no Markdown).

# Output Format (JSON)     
{{
    "label": "yes" | "no",
    "reasoning": "..."
}}

# Video Context
Software: {SOFTWARE}
Task performed in the full screen recording: {TASK_NAME}
Start and end times of the snippet (relative to the full recording): 
{start_time} - {end_time} seconds

# User Behavior Context (*** Optional based on the condition)
The following user behavior is identified in order: {label}: {label_definition}.
Consider this behavior context when determining if the user needs help.

# User Intention (*** Optional based on the condition)
The user's intention or goal during this segment is: {intention}
Consider this intention when determining if the user needs help to achieve this goal.

# Video Content
SEE THE ATTACHED FILE.
\end{Verbatim}
\end{tcolorbox}
\caption{Prompt used to evaluate the model on the (3-1) Help Need Detection task.}
\end{figure*}

\begin{figure*}[h]
\begin{tcolorbox}[width=\linewidth, fontupper=\scriptsize]

\textbf{Help Content Detection} \\

\hrule

\begin{Verbatim}[breaklines]

 # Goal
You are given a screen recording video snippet of a user working in {SOFTWARE}. 
Based on the video content, the user's intention, and the user behavior context (*** Optional based on the condition), predict which option best describes what kind of help or guidance the user is looking for during this segment.

# Instructions
1. Watch the video and predict what help the user might need at this segment.
2. Consider the user's intention and behavior context provided to better understand what they are trying to accomplish and their current state. (*** Optional based on the condition)
3. Select the option (A, B, C, or D) that best matches what help the user needs to accomplish their intention given their behavior context.
4. Provide:
    - label: one of the option letters (A, B, C, or D)
    - reasoning: a concise explanation of why this option best fits the observed segment
5. Output only a valid JSON object (no Markdown).

# Output Format (JSON)     
{{
    "label": "A",
    "reasoning": "..."
}}

# Video Context
Software: {SOFTWARE}
Task performed in the full screen recording: {TASK_NAME}
Start and end times of the snippet (relative to the full recording): 
{start_time} - {end_time} seconds

# User Behavior Context (*** Optional based on the condition)
The following user behavior is identified in order: {label}: {label_definition}.
Consider this behavior context when predicting what help the user might need.

# User Intention (*** Optional based on the condition)
The user's intention or goal during this segment is: {intention}

# Options
{options_text}

# Video Content
SEE THE ATTACHED FILE.
\end{Verbatim}
\end{tcolorbox}
\caption{Prompt used to evaluate the model on the (3-2) Help Content Prediction task.}
\end{figure*}


\end{document}